\documentclass[lettersize,journal]{IEEEtran}
\usepackage{amsmath,amsfonts}
\usepackage{algorithm}
\usepackage{array}
\usepackage{textcomp}
\usepackage{stfloats}
\usepackage{url}
\usepackage{verbatim}
\usepackage{graphicx}
\usepackage{cite}

\usepackage{algpseudocode}
\usepackage{amssymb}
\usepackage{mathtools}
\usepackage{bm}
\usepackage[acronym]{glossaries}
\usepackage{subcaption}
\usepackage{float}
\usepackage{hyperref}
\usepackage{cleveref}
\usepackage{lipsum}
\usepackage{tabularx}
\usepackage{tabularray}
\usepackage{makecell}
\usepackage{multirow}
\usepackage{MnSymbol}
\usepackage{arcs}
\usepackage{adjustbox}
\usepackage{threeparttable, tablefootnote}
\usepackage{multirow}
\usepackage{authblk}

\usepackage{tikz}
\usetikzlibrary{tikzmark}

\newcommand{\carla}{\texttt{CARLA}}
\newcommand{\osqp}{\texttt{OSQP}}
\newcommand{\ipopt}{\texttt{IPOPT}}
\newcommand{\acados}{\texttt{ACADOS}}
\newcommand{\hpipm}{\texttt{HPIPM}}

\algnewcommand\algorithmicnot{\textbf{not}}
\algdef{SE}[IF]{IfNot}{EndIf}[1]{\algorithmicif\ \algorithmicnot\ #1\ \algorithmicthen}{\algorithmicend\ \algorithmicif}%

\graphicspath{pic/}

\makeglossaries

\begin{document}

\title{Safe and Efficient Trajectory Optimization for Autonomous Vehicles using B-Spline with Incremental Path Flattening}

% \author{Jongseo Choi$^{1}$, Hyuntai Chin$^{1}$, Hyunwoo Park$^{1}$, Daehyeok Kwon$^{1,2}$,Sanghyun Lee$^{2}$, and Doosan Baek$^{1,2*}$% <-this % stops a space
% \thanks{$^{1}$ ThorDrive, Seoul, 07268, Republic of Korea
%         {\tt\small \{jschoi, htchin, hwpark, dhkwon, dsbaek\}@thordrive.ai}}%
% \thanks{$^{2}$ Seoul National University, Seoul, Republic of Korea\\
%         {\tt\small slee01@snu.ac.kr}}%
% \thanks{*Corresponding author}% <-this % stops a space
% }

\author{Jongseo Choi, Hyuntai Chin, Hyunwoo Park, Daehyeok Kwon, Doosan Baek, and Sang-Hyun Lee% <-this % stops a space
\thanks{This work was supported by Institute of Information \& communications Technology Planning \& Evaluation (IITP) grant funded by the Korea government(MSIT) (No.2021-0-01415, Development of multi-agent simulation \& test scenario generation SW for edge connected autonomous driving service verification) \textit{(Corresponding author: 
Doosan Baek, Sang-Hyun Lee.)}}%
\thanks{Jongseo Choi, Hyuntai Chin, Hyunwoo Park, and Daehyeok Kwon are with ThorDrive Co., Ltd, Seoul 07268, South Korea}%
\thanks{Daehyeok Kwon and Doosan Baek are with the Department of Electrical Engineering and Computer Science, Seoul National University, Seoul 08826, South Korea}%
\thanks{Sang-Hyun Lee are with the Department of Mobility Engineering, Ajou University, Suwon 16499, South Korea}%
}
%\thanks{This work was supported by ThorDrive Co., Ltd. \textit{(Corresponding author: Doosan Baek.)}}%



% \markboth{IEEE TRANSACTIONS ON INTELLIGENT TRANSPORTATION SYSTEMS}%


\newacronym{sota}{SOTA}{state-of-the-art}
\newacronym{sv}{SV}{swept volume}
\newacronym{D-sv}{D-SV}{Discrete-time swept volume}
\newacronym{C-sv}{C-SV}{Continuous-time swept volume}
\newacronym{d-sv}{D-SV}{discrete-time swept volume}
\newacronym{c-sv}{C-SV}{continuous-time swept volume}
\newacronym{Avs}{AVs}{Autonomous vehicles}
\newacronym{avs}{AVs}{autonomous vehicles}
\newacronym{Av}{AV}{Autonomous vehicle}
\newacronym{av}{AV}{autonomous vehicle}
\newacronym{uav}{UAV}{unmanned aerial vehicle}
\newacronym{uavs}{UAVs}{unmanned aerial vehicles}
\newacronym{s-t}{s-t}{spatio-temporal}
\newacronym{S-t}{s-t}{Spatio-temporal}
\newacronym{pvd}{PVD}{path-velocity Decoupled}
\newacronym{pvc}{PVC}{path-velocity coupled}
\newacronym{ocp}{OCP}{optimal-control problem}
\newacronym{nlp}{NLP}{nonlinear programming}
\newacronym{esdfs}{ESDFs}{Euclidean Signed Distance Fields}
\newacronym{ego-planner}{EGO-Planner}{ESDF-free gradient-based local planning framework}
\newacronym{ipf}{IPF}{incremental path flattening}
\newacronym{kw-rf}{KW-RF}{Knotwise Trajectory Refinement}
\newacronym{l-bfgs}{L-BFGS}{limited-memory BFGS}
\newacronym{mpc}{MPC}{model predictive control}
\maketitle

%\acrshort{ipf} is a new method to find a collision-free path by flattening the path and reducing the \acrshort{sv} using iteratively increasing curvature penalty around vehicle collision points.
%\acrshort{ipf} is a new method that iteratively increases the path curvature weight around vehicle collision points to find a collision-free path by reducing the \acrshort{sv}

\begin{abstract}
Gradient-based trajectory optimization with B-spline curves is widely used for \acrfull{uavs} due to its fast convergence and continuous trajectory generation. However, the application of B-spline curves for path-velocity coupled trajectory planning in \acrfull{avs} has been highly limited because it is challenging to reduce the over-approximation of the vehicle shape and to create a collision-free trajectory using B-spline curves while satisfying kinodynamic constraints. To address these challenges, this paper proposes novel disc-type \acrfull{sv}, \acrfull{ipf}, and kinodynamic feasibility penalty methods. The disc-type \acrshort{sv} estimation method is a new technique to reduce \acrshort{sv} over-approximation and is used to find collision points for \acrshort{ipf}. In \acrshort{ipf}, the collision points are used to push the trajectory away from obstacles and to iteratively increase the curvature weight, thereby reducing \acrshort{sv} and generating a collision-free trajectory. Additionally, to satisfy kinodynamic constraints for \acrshort{avs} using B-spline curves, we apply a clamped B-spline curvature penalty along with longitudinal and lateral velocity and acceleration penalties. Our experimental results demonstrate that our method outperforms state-of-the-art baselines in various simulated environments. We also conducted a real-world experiment using an \acrshort{av}, and our results validate the simulated tracking performance of the proposed approach.

\end{abstract}

%B-spline-based trajectory optimization has been widely used in the field of robot navigation, as the convex-hull property of the B-spline curve guarantees its dynamical feasibility with a small number of control variables. Several recent works demonstrated that quadrotor-like vehicles, which have simple dynamical feasibility constraints, fully utilize the B-spline property for trajectory optimization. Nevertheless, it is still challenging to leverage the B-spline-based optimization algorithm to generate a collision-free trajectory for autonomous vehicles because their complex vehicle kinematics make it difficult to use the convex-hull property. In this paper, we propose a novel incremental path flattening method with a new swept volume method that enables a B-spline-based trajectory optimization algorithm to incorporate vehicle kinematic collision avoidance constraints. Furthermore, a curvature constraint is added with other feasibility constraints (e.g., velocity and acceleration) for the vehicle kinodynamic constraints. Our experimental results demonstrate that our method outperforms state-of-the-art baselines in various simulated environments and verifies its valid tracking performance with an autonomous vehicle in a real-world scenario.

\begin{IEEEkeywords}
Autonomous vehicles, collision avoidance, motion planning, trajectory optimization.
\end{IEEEkeywords}

%%%%%%%%%%%%%%%%%%%%%%%%%%%%%%%%%%%%%%%%%%%%%%%%%%%%%%%%%%%%%%%%%%%%%%%%%%%%%%%%
\section{INTRODUCTION} 
\label{section:Introduction}
\begin{figure}[t]
    \centering
    \includegraphics[scale=1, width=\linewidth]{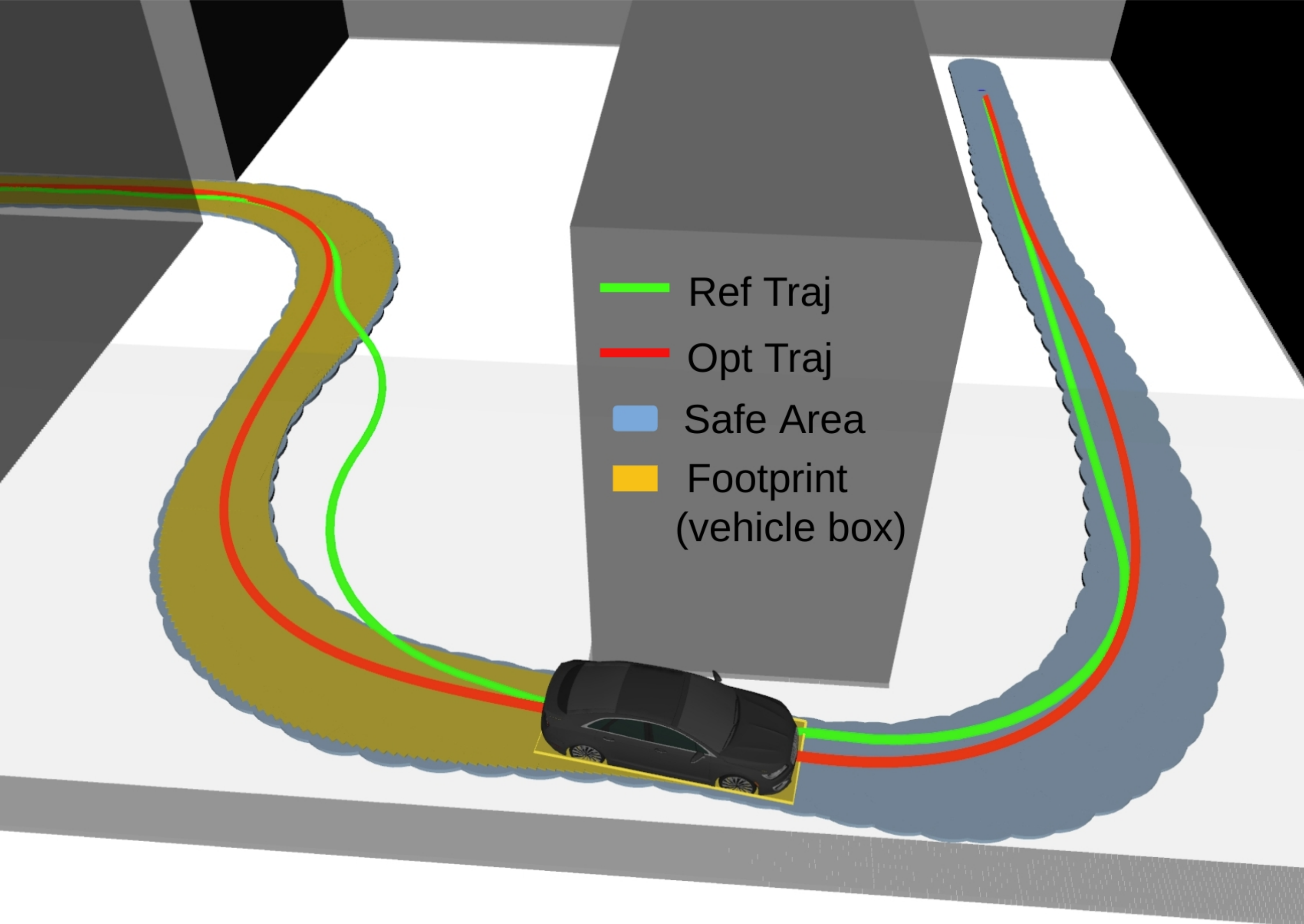}
    \caption{Proposed algorithm. The algorithm can generate an optimized trajectory (red) that allows the vehicle to pass through a narrow corridor based on a reference path (hybrid $\text{A}^*$, green). The footprint (yellow) of the vehicle box inside of the safe area (blue) generated by the proposed \acrshort{sv} estimation method shows that the proposed algorithm can generate a safe, tractable trajectory. The simulated video and more examples can be found at \url{https://youtu.be/iRCl1vtn5dk}.}
    \label{fig:Intro_figure}
\end{figure}

\IEEEPARstart{E}{xtensive} research has focused on trajectory optimization for \acrshort{avs} to generate safe and efficient trajectories. Trajectory optimization algorithms are primarily classified as path-velocity decoupled methods \cite{de-coupled_lim2018hierarchical, DL-IAPS_zhou2020autonomous, em-planner_fan2018baidu}, or coupled methods \cite{coupled_rosmann2017kinodynamic, ocp_li2017optimal, TDR-OBCA_he2021tdr}. Decoupled methods calculate the velocity constraint based on the curvature at positions over time, which is estimated using the given acceleration constraint and the initial speed profile. However, since the exact positions over time cannot be known in advance, it is challenging to accurately calculate the velocity constraint over time \cite{gaussain_cheng2022real}. Despite this drawback, such methods are computationally efficient because they benefit from a two-dimensional breakdown (x-y and s-t domains), which enables their widespread use for trajectory optimization. 
%Decoupled methods face challenges in fully utilizing lateral acceleration to reduce arrival time due to the difficulty in accurately determining the velocity constraint imposed by the lateral acceleration constraint of the initial speed profile \cite{gaussain_cheng2022real}.

On the other hand, coupled methods can consider path and velocity constraints simultaneously but are usually computationally expensive because of their higher-dimensional (x-y-t) optimization. A coupled trajectory optimization problem is commonly described as an \acrfull{ocp}, which entails finding discretized control variables of a dynamic system over a future time horizon. An \acrshort{ocp} has many advantages, including the modeling of complex dynamic systems and the flexible handling of optimal controls under their boundary conditions \cite{li2021optimization, curvy_li2022autonomous}. However, the modeling complexity of an \acrshort{ocp} (e.g., vehicle kinematics, and numerous control variables) can degrade computational efficiency. Additionally, the discretization of a trajectory requires high-precision discretization, which increases the number of constraints, to generate a smooth trajectory.

Gradient-based trajectory optimization has been widely used in numerous papers \cite{ratliff2009chomp, mukadam2016gaussian, esdf_bspline_usenko2017real, B-spline_zhou2019robust, gaussain_cheng2022real} due to its fast convergence, and the use of \acrfull{esdfs} has been crucial for calculating the gradient information. However, computing \acrshort{esdfs} is computationally expensive. A recent paper \cite{zhou2020ego-planner} introduced a method that calculates the magnitude and direction of the gradient using a collision-free guiding path without relying on \acrshort{esdfs}, significantly reducing computation time.

Additionally, B-spline curves have been employed in various gradient-based trajectory optimization methods \cite{esdf_bspline_usenko2017real, B-spline_zhou2019robust, zhou2020ego-planner} because of their advantages, such as continuous trajectory planning, which can be generated with several control points, and the convex-hull property \cite{convex_hull_de1978practical, B-spline_zhou2019robust, zhou2020ego-planner} that guarantees the feasibility of vehicle dynamics, including velocity, acceleration, and jerk. However, these papers are only applicable to \acrshort{uavs} with a circular shape, such as quadrotors, and applying them to rectangular-shaped \acrshort{avs} leads to an over-approximation problem. Furthermore, the application of B-spline curves for \acrshort{avs} faces challenges in reducing the over-approximation of the vehicle shape and in finding collision-free trajectories while satisfying kinodynamic constraints. As a result, B-spline curves are often used only in path-velocity decoupled problems or are used conservatively due to the over-approximation problem (for more details, see Section \ref{subsection:related_works_B}).

To address these challenges, this paper proposes the use of disc-type \acrshort{sv}, \acrshort{ipf}, and kinodynamic feasibility penalties to extend the computationally efficient \acrfull{ego-planner} \cite{zhou2020ego-planner} for quadrotors to be applicable to rectangular-shaped \acrshort{avs}. Through these methods, the proposed algorithm reduces the over-approximation of the vehicle shape and finds collision-free trajectories that satisfy kinodynamic constraints, allowing it to be used in narrow corridors as shown in Fig. \ref{fig:Intro_figure}.

The \acrshort{sv} should be considered for safe trajectory optimization to prevent the well-known \textit{corner-cutting} problem \cite{C_C_deits2015efficient}, which is normally handled using a large enough number of control points \cite{C_C_zhu2015convex, li2021optimization}. Our disc-type \acrshort{sv} estimation method can minimize the over-approximation of \acrshort{sv} by calculating the number of discs and the minimum disc radius needed to cover the actual vehicle \acrshort{sv} and find collision points for \acrshort{ipf}. \acrshort{ipf} is a new method that can find a collision-free path for \acrshort{avs} by using the collision points to push the path away from obstacles with close obstacle collision penalties and iteratively increasing the curvature to flatten the path.

Additionally, to satisfy the vehicle's kinodynamic constraints, we simplified and applied curvature penalties using the properties of clamped B-spline curves. Moreover, thanks to the convex-hull property of B-spline, we calculated and applied the smoothness penalty of the trajectory using only the control points. However, since the convex-hull property of B-spline curves is not directly applicable as constraints in nonholonomic systems with distinct longitudinal and lateral constraints, we applied the longitudinal and lateral velocity and acceleration penalties using integration in the kinodynamic feasibility penalty. Furthermore, we improved the existing refinement method, originally used for holonomic systems, to ensure that it satisfies the velocity and acceleration constraints for nonholonomic systems with distinct longitudinal and lateral components.

%It is necessary to prevent the well-known \textit{corner-cutting} problem \cite{C_C_deits2015efficient} which is normally handled by having large enough control points \cite{C_C_zhu2015convex, li2021optimization}.

%Compared to existing \acrfull{sota} works, our proposed approach is able to generate a safe, efficient, and robust trajectory. The main contributions of this paper can be summarized as follows:

The main contributions of this paper are summarized as follows:

\begin{itemize}
\item We propose a novel path-velocity coupled trajectory optimization approach that enables the use of the \acrshort{ego-planner}, which is a trajectory optimization framework for quadrotors, in \acrshort{avs} by utilizing disc-type \acrshort{sv} and \acrshort{ipf}.

\item We integrate a clamped B-spline curvature penalty and incorporate longitudinal and lateral velocity and acceleration penalties into the trajectory optimization process to generate a feasible trajectory for \acrshort{avs}.

\item We propose a revised solution that adapts the existing refinement method for nonholonomic systems, considering both longitudinal and lateral components.

\item Time efficiency and tracking performance of the proposed trajectory optimization approach are validated through numerous driving tasks in comparison with \acrlong{sota} algorithms.
\end{itemize}

\section{RELATED WORK}
\label{sectiion:related_work}
\subsection{Safe Trajectory Optimization for \texorpdfstring{\acrshort{avs}}{AVS}} \label{subsection:related_works_A}
The major concern in \acrshort{av} trajectory optimization is the generation of collision-free trajectories considering vehicle kinematics. References \cite{non-linear_li2015simultaneous, non-linear_zhang2020optimization} propose \acrfull{nlp} methods with direct obstacle-to-obstacle collision avoidance constraints; these methods include the triangle area criterion method \cite{triangle_li2015unified} and signed distance fields \cite{signed_dist_schulman2014motion}. However, these approaches suffer from inefficiencies because most constraints are redundant; not all obstacles are in collision paths with the vehicle simultaneously. The numerous redundant constraints can significantly slow the system down.

To overcome this issue, researchers \cite{C_C_zhu2015convex, corridor_andreasson2015fast, corridor_liu2018convex, corridor_li2020maneuver} generate safe convex corridors, which are substantially advantageous in terms of computational efficiency because the number of obstacles is irrelevant. Nevertheless, these algorithms suffer from the \textit{corner-cutting} problem \cite{C_C_deits2015efficient}. This problem can be prevented by using a large enough number of control points \cite{C_C_zhu2015convex, li2021optimization}, which can prolong computation, or using various methods to estimate the \acrshort{sv} \cite{swept_volume_proposed_scheuer1997continuous, swept_volume_over_approx_ghita2012trajectory, Bai_Li_footprints_li2023embodied} between the control points of a trajectory. However, they can lead to conservative \acrshort{sv}s because they use convex polygons, thus creating redundant regions at a curve. Therefore, a safe and efficient \acrshort{sv} with minimal over-approximation is vital in the generation of a feasible trajectory for \acrshort{avs} to increase the robustness of the optimization in complex environments.

\subsection{Trajectory Optimization with B-Spline Curves} \label{subsection:related_works_B}
Trajectory optimization algorithms with B-spline curves \cite{B-spline_ding2019efficient, B-spline_zhou2019robust} for quadrotors generate efficient, dynamically feasible trajectories with the computational efficiency caused by the convex-hull property of B-spline curves. \acrshort{ego-planner} \cite{zhou2020ego-planner} accelerates computation by generating gradient information using repulsive force from obstacles instead of \acrshort{esdfs}.

Because of the benefits of B-spline curves, numerous researchers have attempted to use them for \acrshort{avs}. References \cite{B-spline_path_maekawa2010curvature, B-spline_path_elbanhawi2015continuous} use B-spline curves to generate curvature-continuous paths for \acrshort{avs}, but velocity is not considered in path planning. Reference \cite{B-spline_zhang2020trajectory_strip} manipulates B-spline curves using their local-control property for collision avoidance in spatial-temporal path planning. Reference \cite{B-spline_SSC_ding2019safe} uses piecewise $\text{Bézier}$ curves, which have similar properties to B-spline curves, to optimize a trajectory in a spatio-temporal semantic corridor. Reference \cite{B-spline_van2021cooperative} uses B-spline curves for string stability in cooperative driving. However, these algorithms \cite{B-spline_path_maekawa2010curvature, B-spline_path_elbanhawi2015continuous, B-spline_zhang2020trajectory_strip, B-spline_SSC_ding2019safe, B-spline_van2021cooperative} mostly use B-spline curves for a path and velocity separately. Reference \cite{B-spline_mercy2017spline} uses B-spline curves for both path-velocity coupled trajectory planning by utilizing a hyperplane as the collision avoidance constraint, but this method imposes overly conservative constraints by simplifying obstacle shapes such as into circles. The computational inefficiency of this approach is the same as that of the obstacle-to-obstacle collision avoidance method due to redundant constraints and the limited problem solution area caused by the over-approximation of the obstacle volume. Therefore, there is a need for an optimization method that can generate safe and efficient trajectories for autonomous driving in complex environments.

Section \ref{section:B-spline_and_vehicle_kinematics} defines B-spline and \acrshort{sv} estimation method. Section \ref{section:Trajectory_Optimization} explains the overall trajectory optimization algorithm. The evaluation of this algorithm via simulation and real driving tests is explained in Sections \ref{section:evaluation} and \ref{section:experimental_result}. Finally, the conclusion is presented in Section \ref{section:conclusion}.

%By formulating the trajectory optimization with a new method in this paper, the properties of the B-spline curve are fully utilized for path-velocity coupled planning.

%%%%%%%%%%%%%%%%%%%%%%%%%%%%%%%%%%%%%%%%%%%%%%%%%%%%%%%%%%%%%%%%%%%%%%%%%%%%%%%%
% \section{B-spline and Vehicle Kinematic Collision Avoidance Constraint}
\section{B-spline and SV Estimation Method}
\label{section:B-spline_and_vehicle_kinematics}
% \begin{figure}[t]
%     \centering
%     \includegraphics[scale=1, width=\linewidth]{pic/3.bspline_curve.pdf}
%     \caption{B-spline curve with convex-hull property ($p_b=3$)}
%     \label{fig:B-spline_curve}
% \end{figure}

This section provides a detailed parameterization of the B-spline and \acrshort{sv} estimation methods utilized for trajectory optimization with B-spline curves (Section \ref{section:Trajectory_Optimization}). Initially, we present the vehicle model. Subsequently, we parameterize the B-spline primitives as described in \cite{B-spline_zhou2019robust} and \cite{zhou2020ego-planner}, followed by an introduction to the velocity, acceleration, and curvature properties of the B-spline curve. Regarding the \acrshort{sv} estimation method, we review the discrete-time \acrshort{sv} estimation techniques from \cite{curvy_li2022autonomous} and \cite{Bai_Li_footprints_li2023embodied}, and propose a continuous-time \acrshort{sv} estimation method.

%Instead of finding a collision-free path for each colliding segment in an iteration, a collision-free path $\Gamma$, which is generated by hybrid ${A}^*$ based path planning algorithm \cite{hybrid_a_star_dolgov2010path}, is used as an initial guess for the trajectory optimization.

\subsection{Vehicle Model} \label{subsection:Vehicle model}
We designed the module to generate trajectories that can be followed by a vehicle controllable in terms of longitudinal and lateral velocity and acceleration, and curvature. For instance, by utilizing our optimized B-spline and the transformation equations of the B-spline and \acrshort{ocp} used in the bicycle model \cite{B-spline_mercy2017spline}, vehicle states (e.g., steering angle and angular velocity) can be derived. The vehicle is assumed to move based on the rear wheel axle, and we consider non-slip conditions.

% In this paper, we employ a simplified kinematic bicycle model \cite{kong2015kinematic} for the B-spline-based vehicle model, which uses the vehicle's steering angle for control. However, instead of satisfying the constraint on the vehicle's rotation using the steering angle directly, we design the vehicle model to satisfy the constraint using the curvature of the B-spline curve. Specifically, since $\kappa = \tan(\delta) / \textsc{L}_\textsc{W}$, instead of $\delta_{min} \leq \delta \leq \delta_{max}$, we use $\kappa_{min} \leq \kappa \leq \kappa_{max}$, where $\kappa$ is the curvature of the B-spline curve, $\delta$ is the steering angle, and $\textsc{L}_\textsc{W}$ is the vehicle's wheelbase. The vehicle is assumed to move based on the rear-wheel axle, similar to a front-wheel driven vehicle, and no slipping is assumed. 

\subsection{B-Spline Primitives} \label{subsection:B-spline_Primitives}
An initial reference trajectory is parameterized to a \textbf{uniform} B-spline curve $\boldsymbol{\Phi}^{ref}$ with degree $p_b$, $N_c + 1$ \textbf{control points} $\{\mathbf{Q}_{0}, \mathbf{Q}_{1}, \ldots, \mathbf{Q}_{N_c}\}$, and a \textbf{knot vector} $[t_0, t_1, \ldots, t_M]$, where $\mathbf{Q}_i \in \mathbb{R}^2$, $M = N_c + p_b + 1$, $t_m \in \mathbb{R}$, and $t \in [t_m, t_{m+1}) \subset [t_{p_b}, t_{M-p_b}]$. Every \textbf{knot span} $\Delta t = t_{m+1} - t_m$ is the same for a uniform B-spline curve and $t$ is normalized as $u_t = (t-t_m)/\Delta t$. Since the derivatives of B-spline curves of degree $p_b$ are B-spline curves of degree $p_b - 1$, they retain the convex-hull property, ensuring the dynamic feasibility of the entire trajectory. Therefore, the derivatives of a B-spline curve can be formulated with the control points of velocity $\mathbf{V}_i$, acceleration $\mathbf{A}_i$, and jerk $\mathbf{J}_i$, which are obtained as
\begin{equation}\label{eq:control_points_of_V_A_J}
\begin{aligned}
\mathbf{V}_i= \frac{\mathbf{Q}_{i+1}-\mathbf{Q}_i}{\Delta t}, \mathbf{A}_i= \frac{\mathbf{V}_{i+1}-\mathbf{V}_i}{\Delta t}, \mathbf{J}_i= \frac{\mathbf{A}_{i+1}-\mathbf{A}_i}{\Delta t}.
\end{aligned}
\end{equation}

The position of a B-spline at time $t$ (normalized as $u$) can be evaluated using a matrix representation \cite{B-spline_matrix_qin1998general}:

\begin{equation}\label{eq:B-spline_equation}
\begin{aligned}
&\bm{p}_t = \bm{u}^{\top}_t\mathbf{M}_{p_b+1} \mathbf{Q}_{m}, \\
&\bm{u}_t = [1 \quad u_t \quad u^2_t \quad \ldots \quad u^{p_b}_t]^{\top}, \\
&\mathbf{Q}_m = [\mathbf{Q}_{m-p_b} \quad \mathbf{Q}_{m-p_b+1} \quad \mathbf{Q}_{m-p_b+2} \quad \ldots \mathbf{Q}_m]^{\top},
\end{aligned}
\end{equation}

where $\mathbf{M}_{p_b+1}$ is a constant matrix determined by $p_b$. In this work, $\bm{p}_t = (x_t, y_t)$ denotes the midpoint of the rear axle of the ego vehicle at time $t$, and the derivatives at time $t$ are velocity $\bm{v}_t = (v^x_t, v^y_t)$ and acceleration $\bm{a}_t = (a^x_t, a^y_t)$.

%In addition, the derivatives can be evaluated by $\frac{d^k\bm{p}(u)}{dt^k}$, where $\bm{p}^{(k)}_t = (p^{x,(k)}_{t}, p^{y,(k)}_{t})$ denotes the $k$-th-order derivative at time $t$ as well as the position $\bm{p}_t = \bm{p}^{(0)}_t$, the velocity $\bm{v}_t = \bm{p}^{(1)}_t$, the acceleration $\bm{a}_t = \bm{p}^{(2)}_t$, and the jerk $\bm{j}_t = \bm{p}^{(3)}_t$. %The vehicle heading angle at time $t$ is obtained by $\theta_t = \tan^{-1}(v^{y}_t/v^{x}_t)$.

\subsection{Longitudinal and Lateral Velocity and Acceleration}
Unlike quadrotors, which are free to move each axis (x, y, and z), \acrshort{avs} are constrained in both the longitudinal direction $s$ and the lateral direction $d$. Therefore, the B-spline derivatives in each direction should also be formulated. Because $\bm{p}_t$ is the midpoint of the rear axle of the ego vehicle at time $t$, we assume that the velocity $\bm{v}_t$ is always tangent to the vehicle path which means the lateral velocity is equal to zero. Acceleration in both directions $a^{s}_t$, $a^{d}_t$ are calculated as

\begin{equation}\label{eq:longitudinal_v_a_j}
\begin{aligned}
a^{s}_t = \|\bm{a}_t\|\cos(\psi_t - \theta_t),
a^{d}_t = \|\bm{a}_t\|\sin(\psi_t - \theta_t),
\end{aligned}
\end{equation}

where $\theta_t = \tan^{-1}(v^y_t/v^x_t)$, $\psi_t = \tan^{-1}(a^y_t/a^x_t)$, and $\| \cdot \|$ is L2 norm. $\theta_t$ is the vehicle's heading angle at time $t$. %Since the point $\bm{p}_t$ is the midpoint of the rear axle of the ego vehicle at time $t$ as mentioned, we assume the velocity $\bm{v}_t$ is always tangent to the vehicle path. Therefore, the longitudinal velocity $v_t^s$ is equal to $\|\bm{v}_t\|$, and the lateral velocity $v_t^d$ is equal to zero.

% \begin{equation}\label{eq:longitudinal_v_a_j}
% \begin{aligned}
% v_i^s = \|\mathbf{V}_i\|, a_i^s=\frac{v_{i+1}^s - v_i^s}{\Delta t}, %j_i^s=\frac{a_{i+1}^s-a_i^s}{\Delta t},
% \end{aligned}
% \end{equation}

% \begin{equation}\label{eq:lateral_v_a_j}
% \begin{aligned}
% a_i^l = \sqrt{\|\mathbf{A}_i\|^2 - {a_i^s}^2}, j_i^l = \sqrt{\|\mathbf{J}_i\|^2 - {j_i^s}^2}.
% \end{aligned}
% \end{equation}

% \subsection{Flattening Angle}
% A flattening angle $F_i$ is defined as three consecutive control points shown Fig. \ref{fig:B-spline_curve}. It will be used to avoid vehicle collision by flattening the path near vehicle collision points in the \acrshort{ipf} method. It is calculated by
% \begin{equation}\label{eq:flattening_angle}
% \begin{aligned}
% F_i = \cos^{-1}(\frac{(\mathbf{Q}_{i} - \mathbf{Q}_{i-1}) \cdot (\mathbf{Q}_{i+1} - \mathbf{Q}_{i})}{\|\mathbf{Q}_{i} - \mathbf{Q}_{i-1}\|\|\mathbf{Q}_{i+1} - \mathbf{Q}_{i}\|}).
% \end{aligned}
% \end{equation}

\begin{figure}[t]
    \centering
    \includegraphics[scale=1, width=\linewidth]{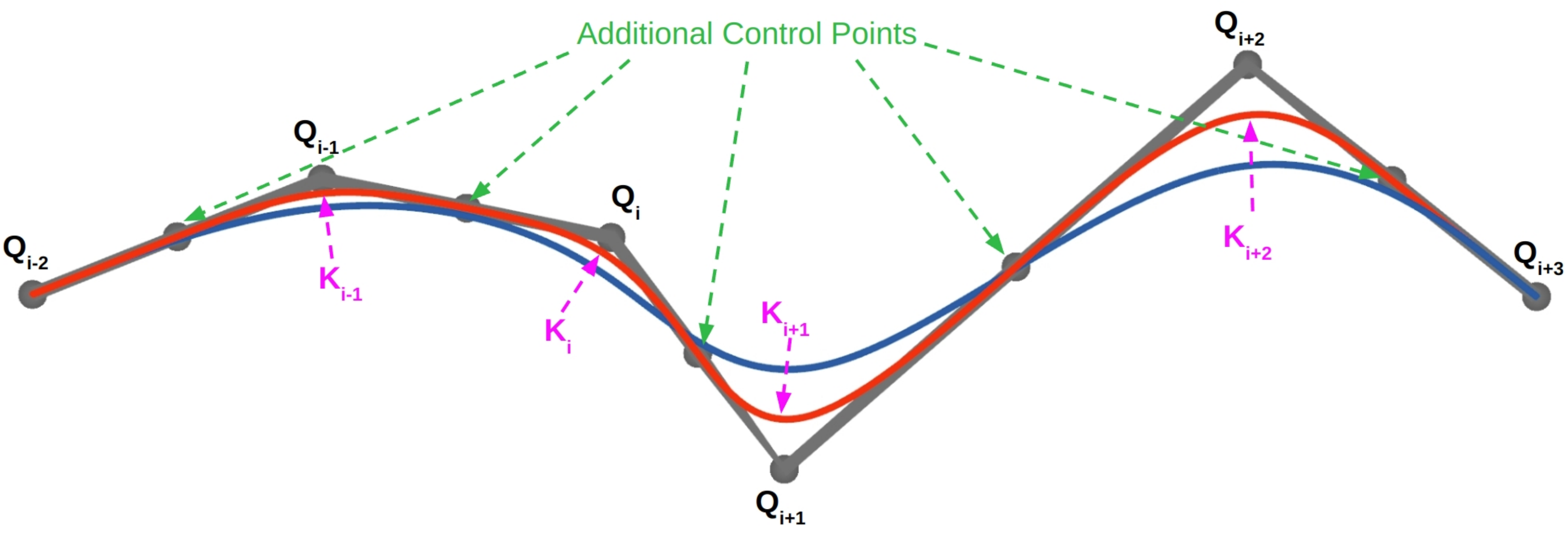}
    \caption{Clamped B-spline curve. This curve is made by adding control points between successive control points. The curvature $K_i$ ($u = 0.5$) of the clamped B-spline curve (red) is bigger than the curvature $\kappa_t$ of the B-spline curve (blue) at time $t$.}
    \label{fig:clamped_bspline_compare}
\end{figure}

\subsection{Curvature} \label{subsection:Curvature}
The curvature of the B-spline is important for creating a tractable path for \acrshort{avs}, but the complicated nature of the B-spline curvature function increases the time needed to impose the curvature constraint on an entire B-spline curvature in real-time \cite{B-spline_extrema_zhao2013real}. Therefore, instead of a B-spline curve, a clamped B-spline curve \cite{clamped_B-spline_curvature_elbanhawi2015continuous}, which always has a bigger curvature than a B-spline curve $|\kappa(t)| \leq max\{|K_{i+1}|, |K_{i+2}|\}$ where $t \in [t_{i+p_b}, t_{i+p_b+1})$, as shown in Fig. \ref{fig:clamped_bspline_compare}, is used to simplify the B-spline curvatures. The curvature of the clamped B-spline curve is obtained as 

\begin{equation}\label{eq:clamped_B-spline_curvature}
\begin{aligned}
K_i &= \frac{1}{6}\frac{\sin{\alpha_i}}{l_i}\left(\frac{1-\cos{\alpha_i}}{8}\right)^{-3/2},\\
\alpha_i &= \cos^{-1}(\frac{(\mathbf{Q}_{i-1} - \mathbf{Q}_{i}) \cdot (\mathbf{Q}_{i+1} - \mathbf{Q}_{i})}{\|\mathbf{Q}_{i-1} - \mathbf{Q}_{i}\|\|\mathbf{Q}_{i+1} - \mathbf{Q}_{i}\|}),
\end{aligned}
\end{equation}

where the length $l_i = min\{\|\mathbf{Q}_{i} - \mathbf{Q}_{i-1}\|, \|\mathbf{Q}_{i+1} - \mathbf{Q}_{i}\|\}$ of the segment, and $K_i$ denotes the maximum curvature of the clamped B-spline curve at $u=0.5$.

According to \cite{clamped_B-spline_curvature_elbanhawi2015continuous}, a small value of $l_i$ results in a large constraint on the angle of the segment to satisfy the maximum curvature. We use a minimum length $\textsc{L}_\text{min}$ for numerical stability and clip any segment length $l_i$ that is less than $\textsc{L}_\text{min}$ to $\textsc{L}_\text{min}$. This can increase the curvature (mostly in the initial and final segments) in a short distance, but this increased short-distance curvature has little effect on the tracking performance, as shown in Section \ref{section:evaluation}.

% The position of the B-spline at time $t$ (normalized as $u$) can be evaluated using the matrix representation \cite{B-spline_matrix_qin1998general}.

% \begin{equation}\label{eq:B-spline_equation}
% \begin{aligned}
% &\mathbf{p}(u) = u^{\top}\mathbf{M}_{p_b+1}\mathbf{q}_m \\
% &u = [1 \quad u \quad u^2 \quad \ldots \quad u^{p_b}]^{\top} \\
% &\mathbf{q}_m = [\mathbf{Q}_{m-p_b} \quad \mathbf{Q}_{m-p_b+1} \quad \mathbf{Q}_{m-p_b+2} \quad \ldots \mathbf{Q}_m]^{\top}.
% \end{aligned}
% \end{equation}

% In addition to that, since the derivative of a B-spline is also a B-spline, the derivatives can be evaluated by $\frac{d^k\mathbf{p}(u)}{dt^k}$, where $\mathbf{p}^{(k)}_t = (p^{x,(k)}_{t}, p^{y,(k)}_{t})$ denotes the $k$-th-order derivative at time $t$ as well as the position $\bm{p}_t = \mathbf{p}^{(0)}_t$, the velocity $\mathbf{v}_t = \mathbf{p}^{(1)}_t$, the acceleration $\mathbf{a}_t = \mathbf{p}^{(2)}_t$, and the jerk $\mathbf{j}_t = \mathbf{p}^{(3)}_t$. The vehicle heading angle at time $t$ is $\theta_t = \tan^{-1}(v^{y}_t/v^{x}_t)$.

\begin{figure}[t]
    \centering
    \includegraphics[scale=1, height=5cm]{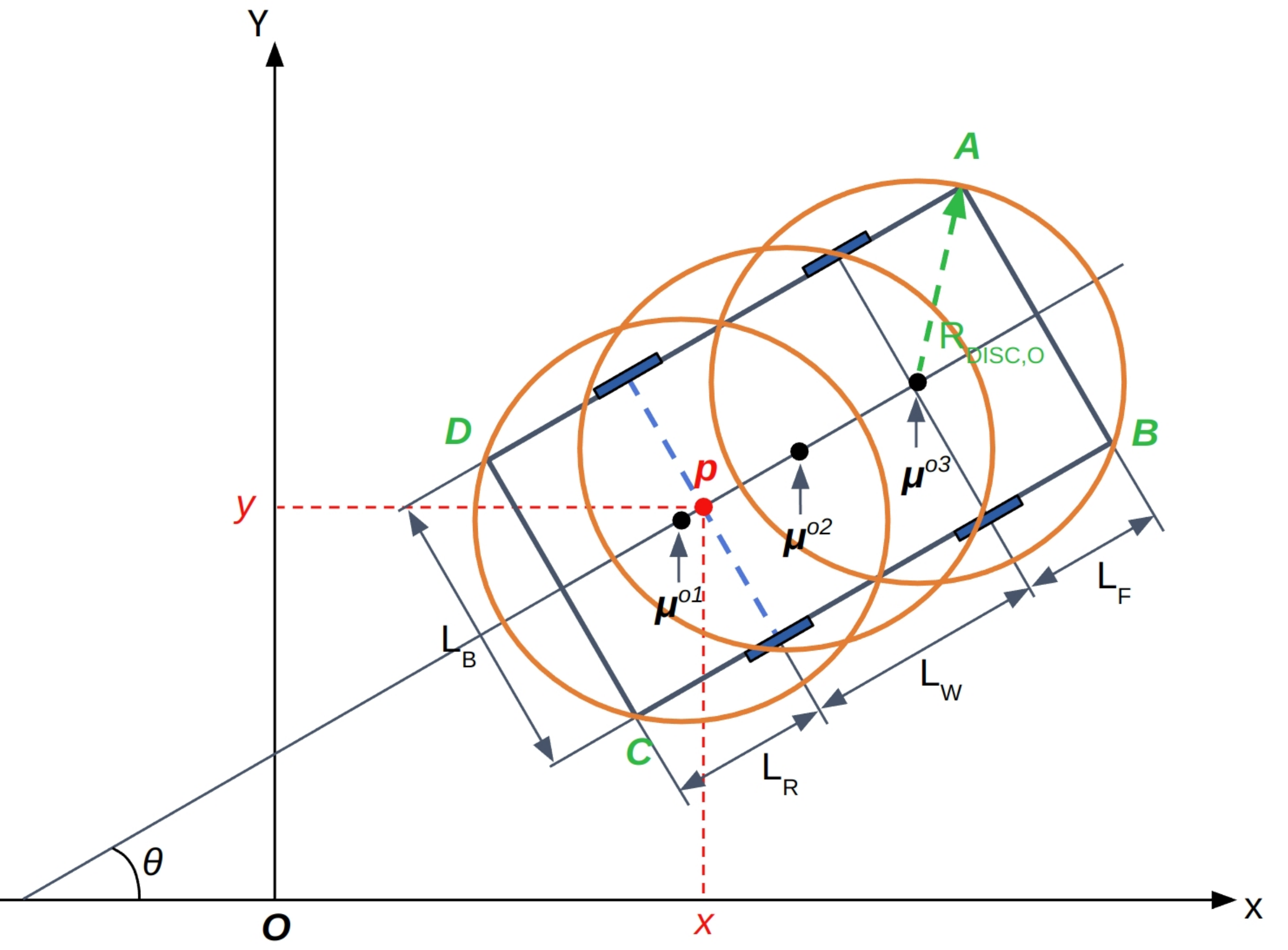}
    \caption{Example of discrete-time \acrshort{sv} estimation with three discs.}
    \label{fig:vehicle_discs}
\end{figure}

\subsection{\texorpdfstring{\acrshort{sv}}{SV} Estimation Method} \label{subsection:vehicle_kinematics_with_coll_constraints}
To cover the discrete-time \acrshort{sv} of a rectangular-shaped \acrshort{av}, various methods such as circles, ellipses, capsules, and convex-polygons can be employed. In continuous-time coverage of \acrshort{sv} to address the \textit{corner-cutting} problem, convex-polygon methods are predominantly used \cite{swept_volume_proposed_scheuer1997continuous, swept_volume_over_approx_ghita2012trajectory, Bai_Li_footprints_li2023embodied}. However, these methods have a structural drawback of creating redundant regions in curves. Therefore, we summarize the \acrfull{d-sv} method \cite{curvy_li2022autonomous, Bai_Li_footprints_li2023embodied}, which models rectangular-shaped \acrshort{av}s as discs in discrete-time. Based on this, we propose the \acrfull{c-sv} method, which aims to reduce over-estimation by using minimum-radius discs while maintaining a similar margin level as the \acrshort{d-sv} method.

\subsubsection{\acrshort{D-sv} Estimation Method}
The \texorpdfstring{\acrshort{d-sv}}{D-SV} of a rectangular-shaped vehicle can be covered with $\textsc{N}_\textsc{DISC,O}$ discs, and the discs have the same radius $\textsc{R}_\textsc{DISC,O}$, as shown in Fig. \ref{fig:vehicle_discs}. The point $(x_t, y_t)$ is the same as point $\bm{p}_t$ of the B-spline curve at time $t$. The four vertices of the ego vehicle are defined as $\bm{A} = (x_{A}, y_{A})$, $\bm{B} = (x_{B}, y_{B})$, $\bm{C} = (x_{C}, y_{C})$, and $\bm{D} = (x_{D}, y_{D})$. The center points of the \acrshort{d-sv} are defined as $\bm{\mu}^{ok} = (x^{\mu,ok}, y^{\mu,ok})$, and these values can be calculated as described in \cite{curvy_li2022autonomous}. The disc radius $\textsc{R}_\textsc{DISC,O}$ can also be calculated as

\begin{equation}\label{eq:radius_of_discs}
\begin{aligned}
\textsc{R}_\textsc{DISC,O} = \sqrt{(\frac{\textsc{L}_\textsc{R} + \textsc{L}_\textsc{W} + \textsc{L}_\textsc{F}}{2 \textsc{N}_\textsc{DISC,O}})^2 + (\frac{\textsc{L}_\textsc{B}}{2})^2},
\end{aligned}
\end{equation}

where $\textsc{L}_\textsc{W}$ represents the wheelbase as shown in Fig. \ref{fig:vehicle_discs}. Additional parameters related to the vehicle geometry, such as $\textsc{L}_\textsc{F}$, $\textsc{L}_\textsc{R}$, and $\textsc{L}_\textsc{B}$, are also illustrated in Fig. \ref{fig:vehicle_discs}.

\begin{figure} [t]
    \centering
    \begin{subfigure}[b]{0.9\linewidth}        %% or \columnwidth
        \centering
        \includegraphics[height=4.5cm, width=0.74\linewidth]{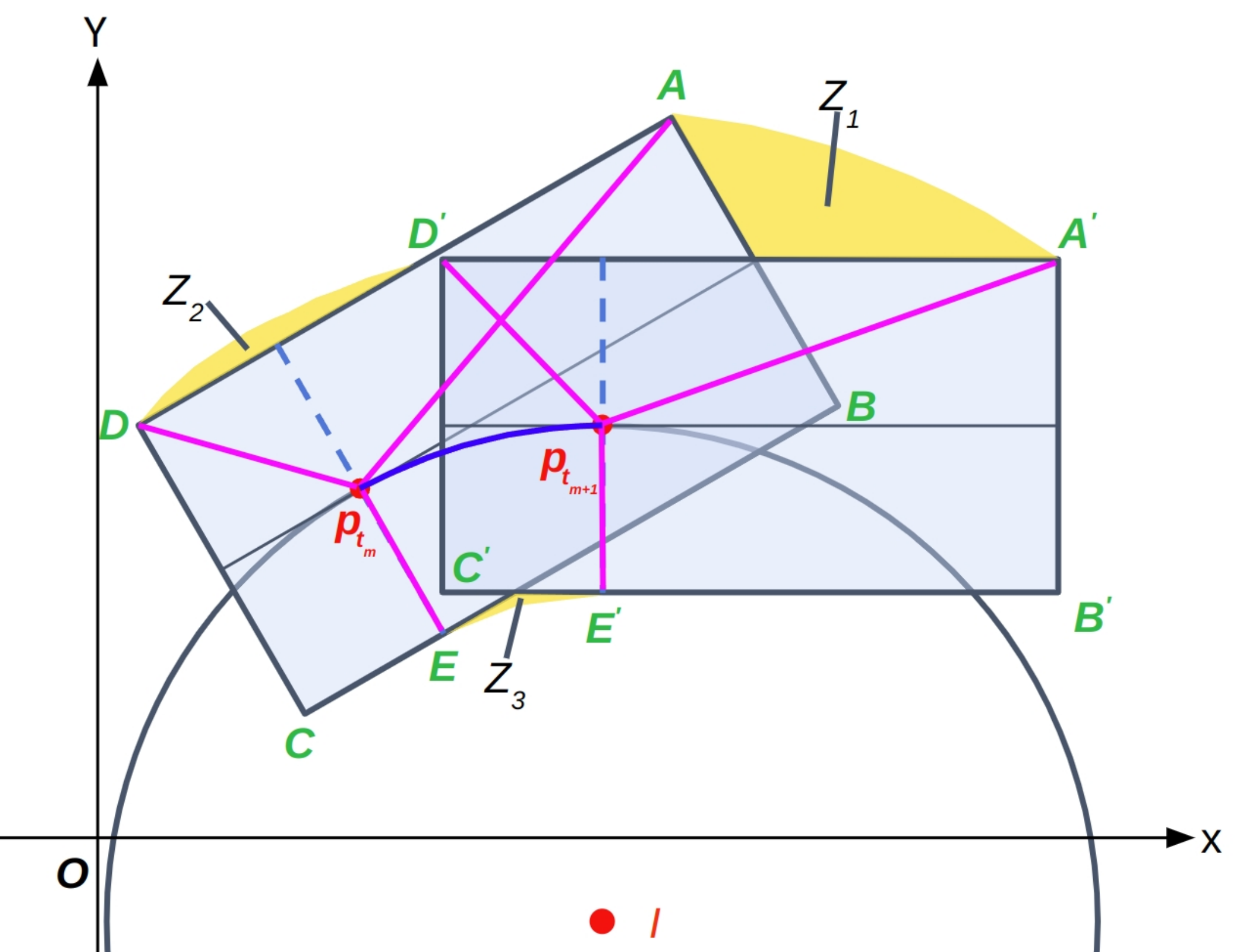}
        \caption{\acrshort{sv}s between two consecutive time knots $(t_m, t_{m+1})$}
        \label{fig:SV_method1}
    \end{subfigure}
    \\[1ex]
    \begin{subfigure}[b]{0.9\linewidth}        %% or \columnwidth
        \centering
        \includegraphics[height=4.5cm, width=0.74\linewidth]{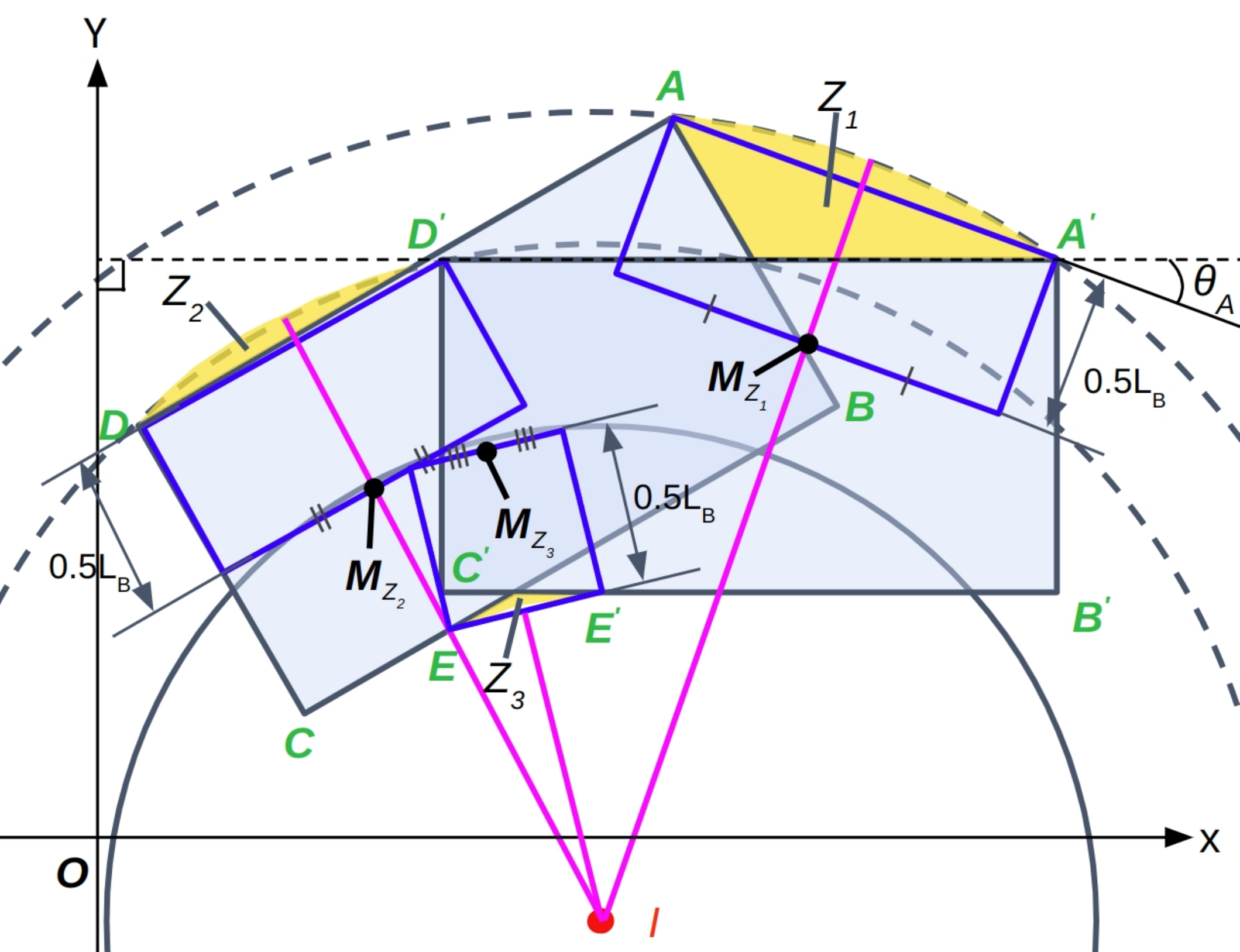}
        \caption{Three zones of \acrshort{c-sv}s}
        \label{fig:SV_method2}
    \end{subfigure}
    \\[1ex]
    \begin{subfigure}[b]{0.9\linewidth}        %% or \columnwidth
        \centering
        \includegraphics[height=4.5cm, width=0.8\linewidth]{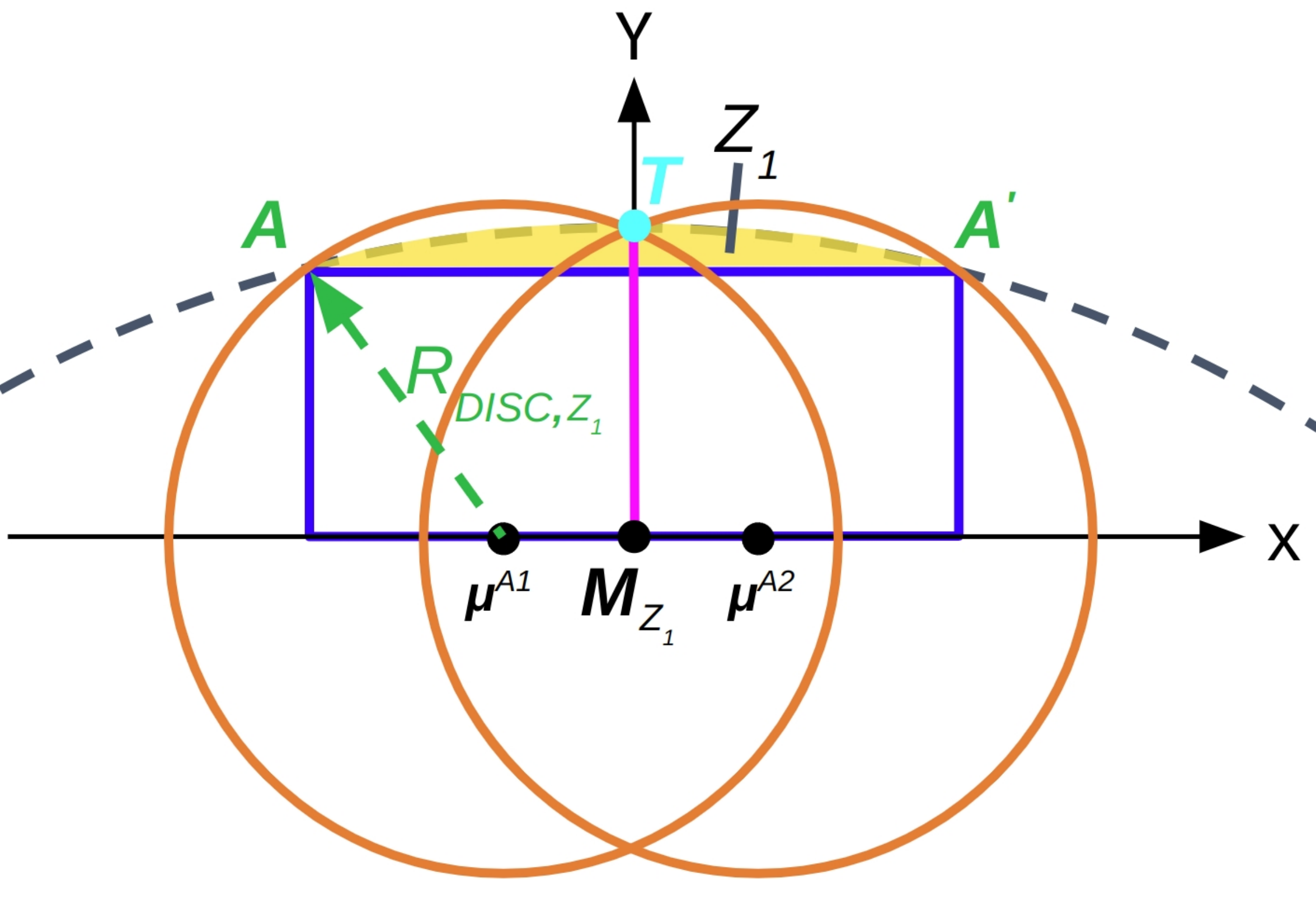}
        \caption{An aligned view of angle $\theta_{A}$ of (b)}
        \label{fig:SV_method3}
    \end{subfigure}
    \caption{Example of continuous-time \acrshort{sv} estimation ($\textsc{CW}$)}
    \label{fig:SV_method}
\end{figure}
% Example of continuous-time \acrshort{sv} estimation ($\textsc{CW}$)
% A normalized angle of review of (c) to facilitate theoretical analysis

\subsubsection{\acrshort{C-sv} Estimation Method} \label{subsubsection:C-sv estimation method}
% The \acrshort{d-sv} estimation method is formulated for a discretely sampled trajectory. However, the continuous execution of a trajectory may involve a collision between time steps. Therefore, an estimation method of \acrshort{c-sv} \cite{swept_volume_proposed_scheuer1997continuous} is introduced. Our \acrshort{c-sv} estimation method is a disc-type estimation approach that can minimize \acrshort{c-sv} over-approximation using minimum-radius discs rather than convex polygons \cite{swept_volume_proposed_scheuer1997continuous, swept_volume_over_approx_ghita2012trajectory, Bai_Li_footprints_li2023embodied}, which lead to redundant regions at curves.

% 먼저 \acrshort{sv}을 설명하기 앞서 Section \ref{subsection:Curvature}에서 언급했던것처럼 B-spline curve의 연속된 두 time knots points 사이의 곡률은 clamped B-spline으로 구한 곡률 $K$를 넘을 수 없기 때문에 이 두 points 사이의 곡률은 최대곡률 $K$라고 가정을 한다. 또한, 이 두 points 사이의 길이는 $\textsc{L}_\textsc{W}$보다 크지 않다고 가정을 하였다.

Before explaining the \texorpdfstring{\acrshort{c-sv}}{C-SV} estimation method, as mentioned in Section \ref{subsection:Curvature}, the curvature between two consecutive time knots of the B-spline curve cannot exceed the curvature $K$ obtained by the clamped B-spline. For \acrshort{c-sv}, since the \acrshort{sv} increases with higher curvature between two points, we assume that the curvature between these two points is this maximum curvature $K$. Additionally, we assume that the distance between two points does not exceed the \textit{vehicle-length} (i.e., $\textsc{L}_\textsc{R} + \textsc{L}_\textsc{W} + \textsc{L}_\textsc{F}$).

An example of \acrshort{c-sv} estimation with a clockwise ($\textsc{CW}$) turning vehicle is shown in Fig. \ref{fig:SV_method}. The \acrshort{c-sv}s, which are outside of the vehicle volume at $t_m$ and $t_{m+1}$, can be divided into three zones illustrated in Fig. \ref{fig:SV_method1}. For the first and second zones $Z_1$, $Z_2$ of the \acrshort{c-sv}s, it is generated based on the vertices $\bm{A}$ and $\bm{D}$, which are the farthest from the instantaneous center of rotation $\bm{I}$ of the ego vehicle, as the outer edge $\overline{AD}$ of the vehicle sweeps through. For the third zone $Z_3$ of the \acrshort{c-sv}s, it is generated based on the vertex $\bm{E}$, which is the nearest to $\bm{I}$ on the inner rear axis of the vehicle, as the inner edge $\overline{BC}$ sweeps through.

% 첫 번째 파트의 경우 비교적 작은 영역이기 때문에 Fig.\ref{fig:SV_method2}와 같이 disc 하나로 \acrshort{sv}을 모델링하였다. $M_{\circled{1}}$는 disc의 중점으로 $\bm{p}_{t_{m}}$ 과 $\bm{p}_{t_{m+1}}$의 중간점이고 이 원의 반지름 $R_{\circled{1}}$ 은 점 $M_{\circled{1}}$ 와 점 $E^{'}$ 사이의 거리로 계산된다.

% For the first zone, as it is a relatively small area, the SV is modeled as a single disc, as shown in Fig. \ref{fig:SV_method2}. $M_{Z_1}$ is the center point of the disc, which is the midpoint between $\bm{p}_{t_{m}}$ and $\bm{p}_{t_{m+1}}$, and the radius of this circle $R_{Z_1}$ is calculated as the distance between the points $M_{Z_1}$ and $E^{'}$.

To reduce over-estimation of the \acrshort{c-sv}s, we modeled them using multiple discs. First, as shown in Fig. \ref{fig:SV_method2}, we draw a rectangle with ($\bm{A}$, $\bm{A}^{'}$), ($\bm{D}$, $\bm{D}^{'}$), and ($\bm{E}$, $\bm{E}^{'}$) as one side and $0.5\textsc{L}_\textsc{B}$ length as the other side, and calculate the middle points $\bm{M}_{Z_1}$, $\bm{M}_{Z_2}$ and $\bm{M}_{Z_3}$ of the one side accordingly. Since these zones have the same mechanism, without loss of generality, we explain with an aligned view of angle $\theta_A$ of Fig. \ref{fig:SV_method2} based on the rectangle of the first zone shown in Fig. \ref{fig:SV_method3} for the convenience of analysis. The key point in this section is that the outermost edge of the first zone should be covered by the newly modeled discs. First, the number of discs to be modeled, $N_{DISC,A}$, is calculated using the formula $\lceil \overline{AA^{'}} / \textsc{R}_\textsc{DISC,O} \rceil$. Here, setting one side of the rectangle to $0.5\textsc{L}_\textsc{B}$ and using $\textsc{R}_\textsc{DISC,O}$ to determine $N_{DISC,A}$ ensures that we maintain the same extent of margin established in the \acrshort{d-sv}. The center points of the \acrshort{c-sv} discs are defined as $\bm{\mu}^{Ak}$ and can be calculated by the following equation:

\begin{equation}\label{eq:swept_volume_center_of_disc}
\begin{aligned}
x^{\mu,Ak}_t &= x_{M_{Z_1},t} + \Bigg{(}\frac{2k - 1}{2 N_{DISC,A}} \cdot \overline{AA^{'}} - \frac{\overline{AA^{'}}}{2}\Bigg{)} \cdot \cos{\theta}_{A,t}, \\
y^{\mu,Ak}_t &= y_{M_{Z_1},t} + \Bigg{(}\frac{2k - 1}{2 N_{DISC,A}} \cdot \overline{AA^{'}} - \frac{\overline{AA^{'}}}{2}\Bigg{)} \cdot \sin{\theta}_{A,t}, \\
&k \in \{1, \ldots, N_{DISC,A}\}.
\end{aligned}
\end{equation}

% Next, the minimum radius of the \acrshort{c-sv} discs can be determined by the larger radius between the minimum radius $\textsc{R}^{a}_{Z_1}$, which can include $\bm{A}$ and $\bm{A}^{'}$, and the minimum radius $\textsc{R}^{b}_{Z_1}$, which can include point $\bm{T}$, the outermost edge of the first zone. Point $\bm{T}$ represents the furthest point from the center within the first zone, and is critical in determining the minimum radius required to encompass the entire zone.
% Next, the minimum radius of the \acrshort{c-sv} discs can be determined by the larger radius between the minimum radius $\textsc{R}^{a}_{Z_1}$, which can include $\bm{A}$ and $\bm{A}^{'}$, and the minimum radius $\textsc{R}^{b}_{Z_1}$, which can include point $\bm{T}$, the outermost edge of the first zone.
Next, the minimum radius of the \acrshort{c-sv} discs can be determined by the larger radius between the minimum radius $\textsc{R}^{a}_{Z_1}$ and $\textsc{R}^{b}_{Z_1}$. $\textsc{R}^{a}_{Z_1}$ represents the smaller radius of the disc centered at $\bm{\mu}^{Ak}$ that can include either point $\bm{A}$ or $\bm{A}^{'}$, while $\textsc{R}^{b}_{Z_1}$ represents the minimum radius of the disc centered at $\bm{\mu}^{Ak}$ that can include point $\bm{T}$, which is the furthest point of the first zone from the instantaneous center of rotation $\bm{I}$.

The minimum radius $\textsc{R}^{a}_{Z_1}$ can be calculated using Eq. \ref{eq:radius_of_discs} as $\sqrt{(\frac{\overline{AA^{'}}}{2 \textsc{N}_{DISC,A}})^2 + (\frac{\textsc{L}_\textsc{B}}{2})^2}$, and the minimum radius $\textsc{R}^{b}_{Z_1}$ can be obtained using the distance from the point $\bm{\mu}^{Ak}$ to $T$ where $k$ is an index closest to the point $\bm{M}_{Z_1}$ ($k=1$ in Fig. \ref{fig:SV_method3}). $\textsc{R}^{b}_{Z_1}$ can be calculated using the Pythagorean theorem as $\sqrt{(\overline{M_{Z_1}T})^2 + (\overline{M_{Z_1}\mu^{Ak}})^2}$, where the length of $\overline{M_{Z_1}T}$ is $\overline{IA} - \overline{IM_{Z_1}}$ (the magenta line in Fig. \ref{fig:SV_method3}). For the third zone, since the \acrshort{c-sv} is concave, unlike the first and second zones, it can be substituted with $0.5\textsc{L}_\textsc{B}$ instead of the length of $\overline{M_{Z_1}T}$. Therefore, $\textsc{R}_{DISC,Z_1}$ is calculated as $max\{{\textsc{R}^{a}_{Z_1}, \textsc{R}^{b}_{Z_1}}\}$. Note that, from a practical perspective, the second and third zones can be excluded if the number of $\textsc{N}_\textsc{DISC,O}$ is not large, as they can typically be encompassed by the radius $\textsc{R}_\textsc{DISC,O}$ in most cases.

The \acrshort{c-sv} estimation is formulated using multiple discs with the minimum-radius $\textsc{R}_{DISC,v}$ where $v \in \{Z_1, Z_2, Z_3 \}$ that can cover the first, second, and third zones of the \acrshort{c-sv}s. Thus, it can be used for the optimization algorithm of safe and efficient trajectories.

\section{B-spline Trajectory Optimization}
\label{section:Trajectory_Optimization}

%Unlike a holonomic system that is free to move each axis like quadrotors, \acrshort{avs} are a nonholonomic system where both motions to longitudinal direction $s$ and lateral direction $d$ are constrained. Therefore, the derivatives of the B-spline for each direction are also necessary to formulate. Velocity $\mathbf{v}^{\sigma}_t$, acceleration $\mathbf{a}^{\sigma}_t$, and jerk $\mathbf{j}^{\sigma}_t$, where ${\sigma}=\{s, d\}$ are calculated by

%\begin{equation}\label{eq:longitudinal_v_a_j}
%\begin{split}
%\mathbf{v}^{s}_t &= \|\mathbf{p}^{(1)}_t\|, \\
%\mathbf{a}^{s}_t &=\|\mathbf{p}^{(2)}_t\|\cos(\theta^{(2)}_t - \theta^{(1)}_t), \\
%\mathbf{j}^{s}_t &=\|\mathbf{p}^{(3)}_t\|\cos(\theta^{(3)}_t - \theta^{(1)}_t),
%\end{split}
%\;
%\begin{split}
%\mathbf{v}^{d}_t &= 0, \\
%\mathbf{a}^{d}_t &= \|\mathbf{p}^{(2)}_t\|\sin(\theta^{(2)}_t - \theta^{(1)}_t), \\
%\mathbf{j}^{d}_t &= \|\mathbf{p}^{(3)}_t\|\sin(\theta^{(3)}_t - \theta^{(1)}_t),
%\end{split}
%\end{equation}

%where $\theta^{(k)}_t = \tan^{-1}(p^{y,(k)}_t/p^{x,(k)}_t)$ and $\theta^{(1)}_t$ is the vehicle heading angle at time $t$.

%\begin{equation}\label{eq:longitudinal_v_a_j}
%\begin{aligned}
%v_i^s = \|\mathbf{V}_i\|, a_i^s=\frac{v_{i+1}^s - v_i^s}{\Delta t}, %j_i^s=\frac{a_{i+1}^s-a_i^s}{\Delta t},
%\end{aligned}
%\end{equation}

%\begin{equation}\label{eq:lateral_v_a_j}
%\begin{aligned}
%a_i^l = \sqrt{\|\mathbf{A}_i\|^2 - {a_i^s}^2}, j_i^l = \sqrt{\|\mathbf{J}_i\|^2 - {j_i^s}^2}.
%\end{aligned}
%\end{equation}

The proposed trajectory optimization approach utilizing B-spline curves consists of two optimization stages based on the \acrshort{ego-planner} \cite{zhou2020ego-planner}. The first stage, known as rebound optimization, iteratively seeks a smooth and collision-free trajectory. This stage employs smoothness, collision, and flattening penalties for optimization. After each iteration, circle and \acrshort{sv} collision checking is performed. In the event of a collision, a collision vector is added in the case of a circle collision, while for a \acrshort{sv} collision, the flattening weight is increased by identifying the flattening index set. In the second stage, referred to as trajectory refinement, the planning horizon of the trajectory is relaxed to the extent that it exceeds the kinodynamic constraints (velocity, acceleration, and curvature). Utilizing a fitness penalty to preserve the existing collision-free path and a kinodynamic feasibility penalty, this stage generates a trajectory that is both collision-free and kinodynamic feasible.

% The proposed trajectory optimization with B-spline curves approach has two optimization stages based on \acrshort{ego-planner} \cite{zhou2020ego-planner}. The first stage called rebound optimization, iteratively finds a smooth and collision-free trajectory. It optimizes the trajectory with some penalties such as smoothness, collision, and flattening  and  two collision avoidance constraints: circle and kinematics collision avoidance constraints. The circle collision avoidance constraint is collision checking with one disc, which has radius $\textsc{R}_\textsc{CIRCLE}$ as shown in Fig. \ref{fig:IPF1}, and the kinematics collision avoidance constraints are collision checking with the discrete- and continuous-time \acrshort{sv} estimation methods in Section \ref{subsection:vehicle_kinematics_with_coll_constraints} on the optimized trajectory. The second optimization stage, called trajectory refinement, ensures that the optimized trajectory from the first stage is bounded by the kinodynamic feasibility constraints (velocity, acceleration, and curvature) if it violates the kinodynamic feasibility constraints.

\subsection{Problem Formulation}
Initially, the control variables are the B-spline curve's control points $\mathbf{Q}$. Each control point $\mathbf{Q}$ independently holds its own environmental information. A collision-free trajectory $\bm{\Gamma}$ satisfying the terminal constraints is provided. The collision-free trajectory $\bm{\Gamma}$ is parameterized to the reference uniform B-spline curve $\boldsymbol{\Phi}^{ref}$ with the same time interval $\Delta t$. Since both the initial and final state boundary conditions for the B-spline curve $\{\mathbf{p}_{t_{p_b}}, \mathbf{v}_{t_{p_b}}, \mathbf{a}_{t_{p_b}}, \theta_{t_{p_b}}, \mathbf{p}_{t_{M-p_b}}, \mathbf{v}_{t_{M-p_b}}, \mathbf{a}_{t_{M-p_b}}, \theta_{t_{M-p_b}}\}$ are determined, control points $\{\mathbf{Q}_{p_b}, \mathbf{Q}_{p_b+1}, \ldots, \mathbf{Q}_{N_c-p_b}\}$ without the initial and final $p_b$ control points are used as the control variables for optimization. %Due to the fixed initial and final control points, the initial and final angles $\{ \alpha_1, \ldots, \alpha_{p_b - 2}, \alpha_{N_c-p_b+2}, \ldots, \alpha_{N_c-1}\}$ satisfy the constraint $\cos^{-1}(\frac{a^2-32\sqrt{a^2+64}+256}{a^2}) \leq \alpha_i \leq \pi$ obtained by Fig. \ref{eq:clamped_B-spline_curvature} where $a=6 \kappa_\text{max} \textsc{L}_i$, and $\kappa_\text{max}$ denotes a maximum curvature.

The rebound optimization cost function $J$ is formulated as

\begin{equation}\label{eq:rebound_cost_function}
\begin{aligned}
\min_{\mathbf{Q}}J = \lambda_{sm} J_{sm} + \lambda_{cl} J_{cl} + \lambda_{fl} J_{fl},
\end{aligned}
\end{equation}

where $J_{sm}$, $J_{cl}$, and $J_{fl}$ are smoothness, collision, and flattening penalties respectively, and $\lambda_{sm}$, $\lambda_{cl}$, and $\lambda_{fl}$ are their corresponding weights.

%\subsubsection{Init/Goal-Boundary Penalty}
%To satisfy the initial and goal heading constraint, a proposed method using an anisotropic displacement method in the fitness penalty from \cite{zhou2020ego-planner} is utilized. Init/goal-heading points are pulled to the reference trajectory $\boldsymbol{\Phi}^{ref}$ with a large enough weight $\lambda_p$. The initial heading points index set $\mathbf{\Omega}^{init}$ is obtained by $\mathbf{\Omega}^{init} = \{\forall i \mid 
%\sum_{k=0}^{i} \|\mathbf{Q}_{k+1} - \mathbf{Q}_k \| < L_\textsc{HEAD}, i \in \{1, \ldots, N_c\} \}$ and the goal heading points index set $\mathbf{\Omega}^{goal}$ is obtained by $\mathbf{\Omega}^{goal} = \{\forall i \mid \sum_{k=0}^i \|\mathbf{Q}_{N_c - k} - \mathbf{Q}_{N_c - k - 1} \| < L_\textsc{HEAD}, i \in \{1, \ldots, N_c\} \}$, where $L_\textsc{HEAD}$ is a minimum length to ensure the heading point. The init/goal-heading points boundary penalty function is 
%\begin{equation}\label{eq:two-point_boundary_penalty}
%\begin{aligned}
%J_p = j_p(\mathbf{Q}_i^{ref}, \mathbf{Q}_i),
%\end{aligned}
%\end{equation}
%where $i \in \{\mathbf{\Omega}^{init}, \mathbf{\Omega}^{goal}\}$, a function $j_p$ indicates a fitting penalty function from %\cite{zhou2020ego-planner}, $\mathbf{Q}_i^{ref}$ and $\mathbf{Q}_i$ are control points of a safe B-spline curve %$\mathbf{\Phi}_s$ and a re-generated B-spline curve $\mathbf{\Phi}_f$ as input parameters of the function respectively.

%The initial penalty is at $t=t_m, m=p_b$ and the goal penalty is at $t=t_{M-p_b}, m=N_c$.

\subsubsection{Smoothness Penalty}
The smoothness penalty cost, which minimizes the high-order derivatives and curvature of the whole trajectory, is defined as

\begin{equation}\label{eq:smooth_penalty}
\begin{aligned}
J_{sm} = \sum_{i=0}^{N_c-2} \Bigl(\frac{{\|\mathbf{A}_i\|}}{s_\text{A}}\Bigl)^2
+ \sum_{i=0}^{N_c-3} \Bigl(\frac{{\|\mathbf{J}_i\|}}{s_\text{J}}\Bigl)^2 + \sum_{i=1}^{N_c - 1} \Bigl(\frac{{K_i}}{\kappa_\text{max}}\Bigl)^2,
\end{aligned}
\end{equation}

where $s_\text{A}$ and $s_\text{J}$ are scale parameters for the acceleration and jerk, respectively, and $\kappa_\text{max}$ is the maximum parameter of curvature.

\begin{figure}[t]
    \centering
    \begin{subfigure}[b]{0.48\linewidth}        %% or \columnwidth
        \centering
        \includegraphics[width=\linewidth]{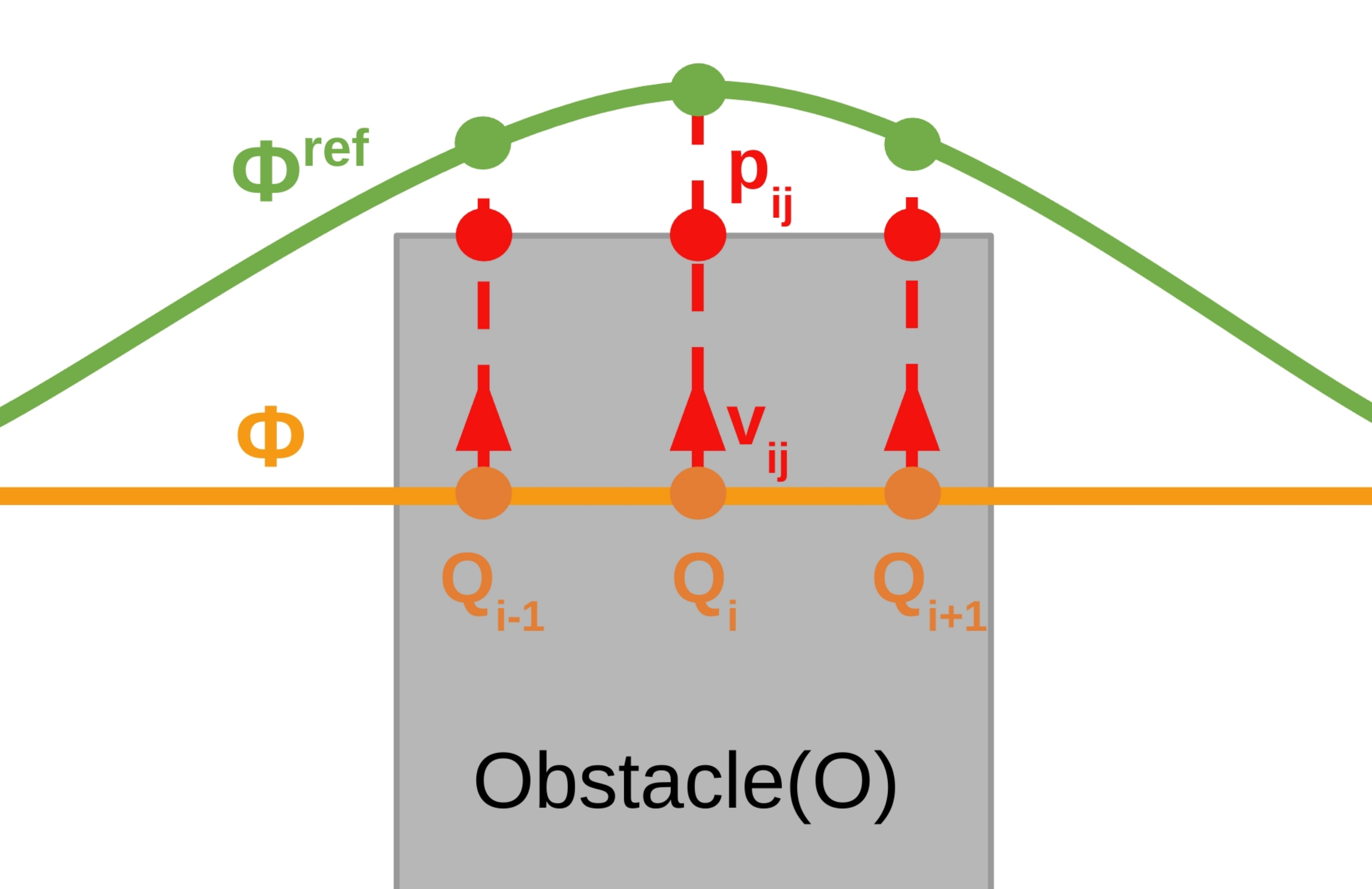}
        \caption{Collision Obstacle}
        \label{fig:coll_penalty_inside}
    \end{subfigure}
    %\hfil
    \begin{subfigure}[b]{0.48\linewidth}        %% or \columnwidth
        \centering
        \includegraphics[width=\linewidth]{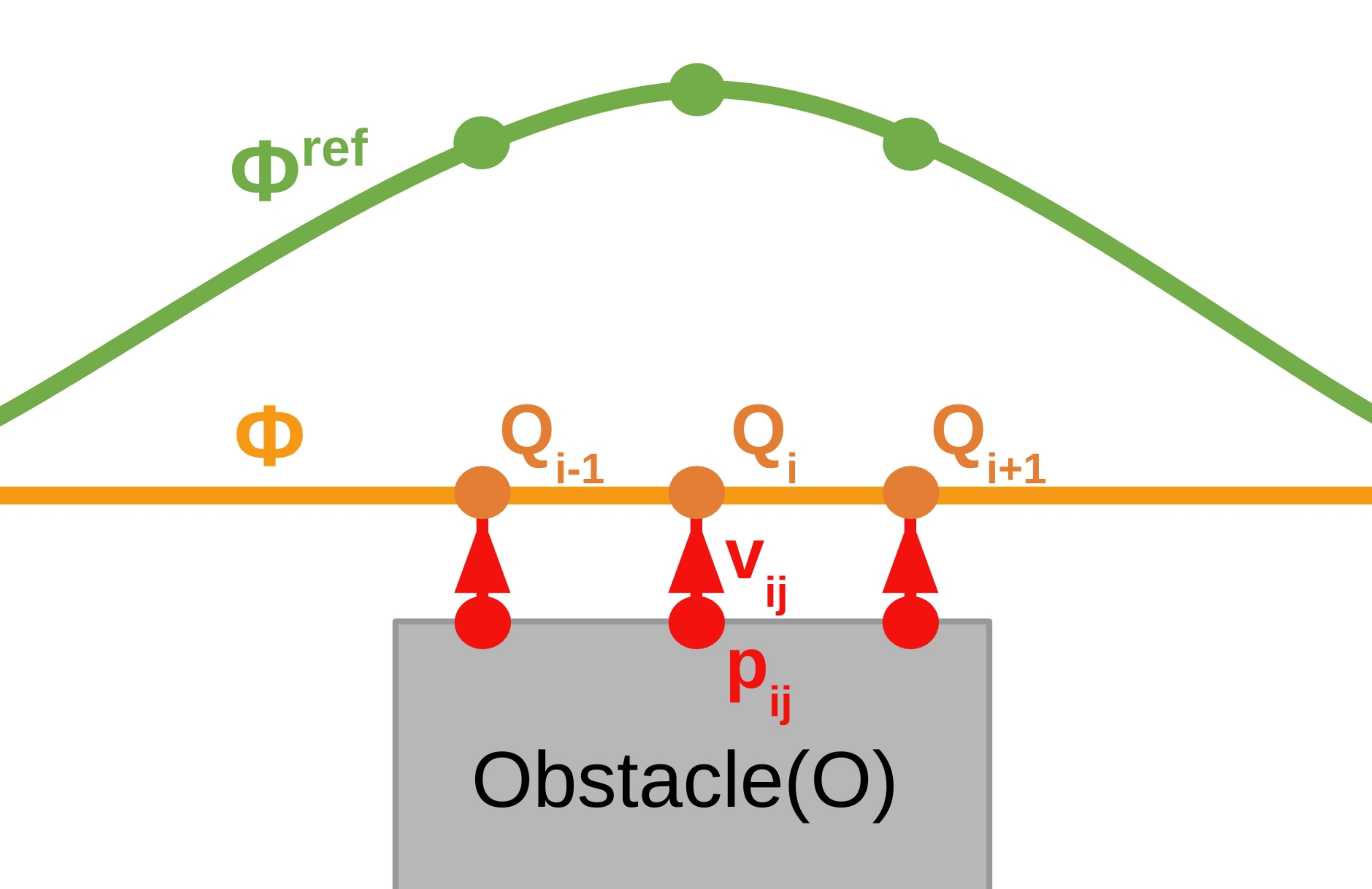}
        \caption{Close Obstacle}
        \label{fig:coll_penalty_outside}
    \end{subfigure}
    \caption{Collision penalty: (a) collision obstacle, (b) close obstacle. For the collision obstacle trajectory $\mathbf{\Phi}$, the anchor point $\mathbf{p}_{ij}$ is the closest point at the obstacle surface from the line between the corresponding control points $\mathbf{Q}_i$ and $\mathbf{Q}^{ref}_i$, and the direction vector $\mathbf{v}_{ij}$ is a unit vector pointing toward $\mathbf{p}_{ij}$ from $\mathbf{Q}_i$. For the close-obstacle trajectory $\mathbf{\Phi}$, which is within a safe clearance $s_f$, the anchor point $\mathbf{p}_{ij}$ is the closest point from the corresponding control point $\mathbf{Q}_i$, and the direction vector $\mathbf{v}_{ij}$ is pointing toward $\mathbf{Q}_i$ from $\mathbf{p}_{ij}$.}
    \label{fig:coll_penalty}
\end{figure}

\subsubsection{Collision Penalty} \label{subsubsection: collision penalty}
\begin{figure*}[!ht]
        \begin{subfigure}[b]{0.22\textwidth}
                \includegraphics[width=\linewidth]{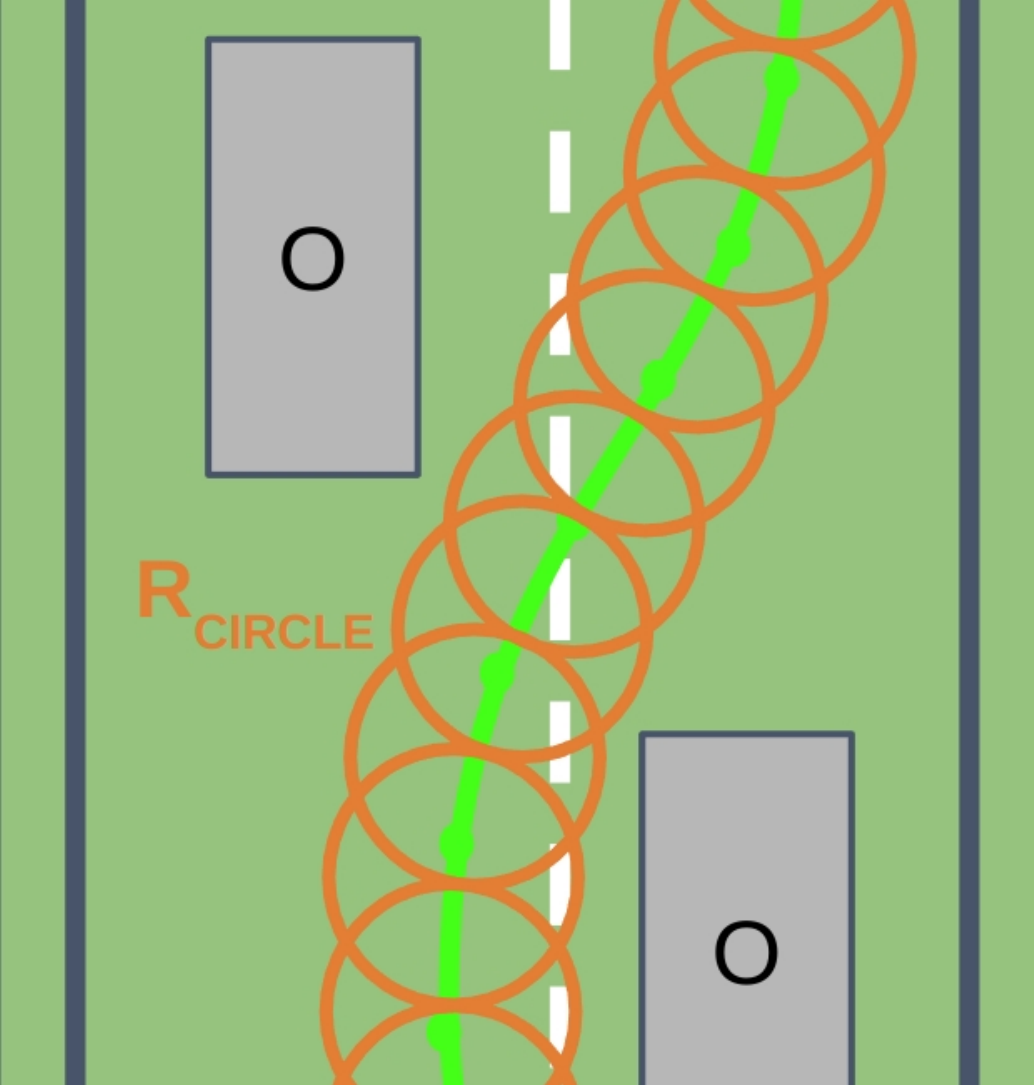}
                \caption{}
                \label{fig:IPF1}
        \end{subfigure}%
        \hfill
        \begin{subfigure}[b]{0.22\textwidth}
                \includegraphics[width=\linewidth]{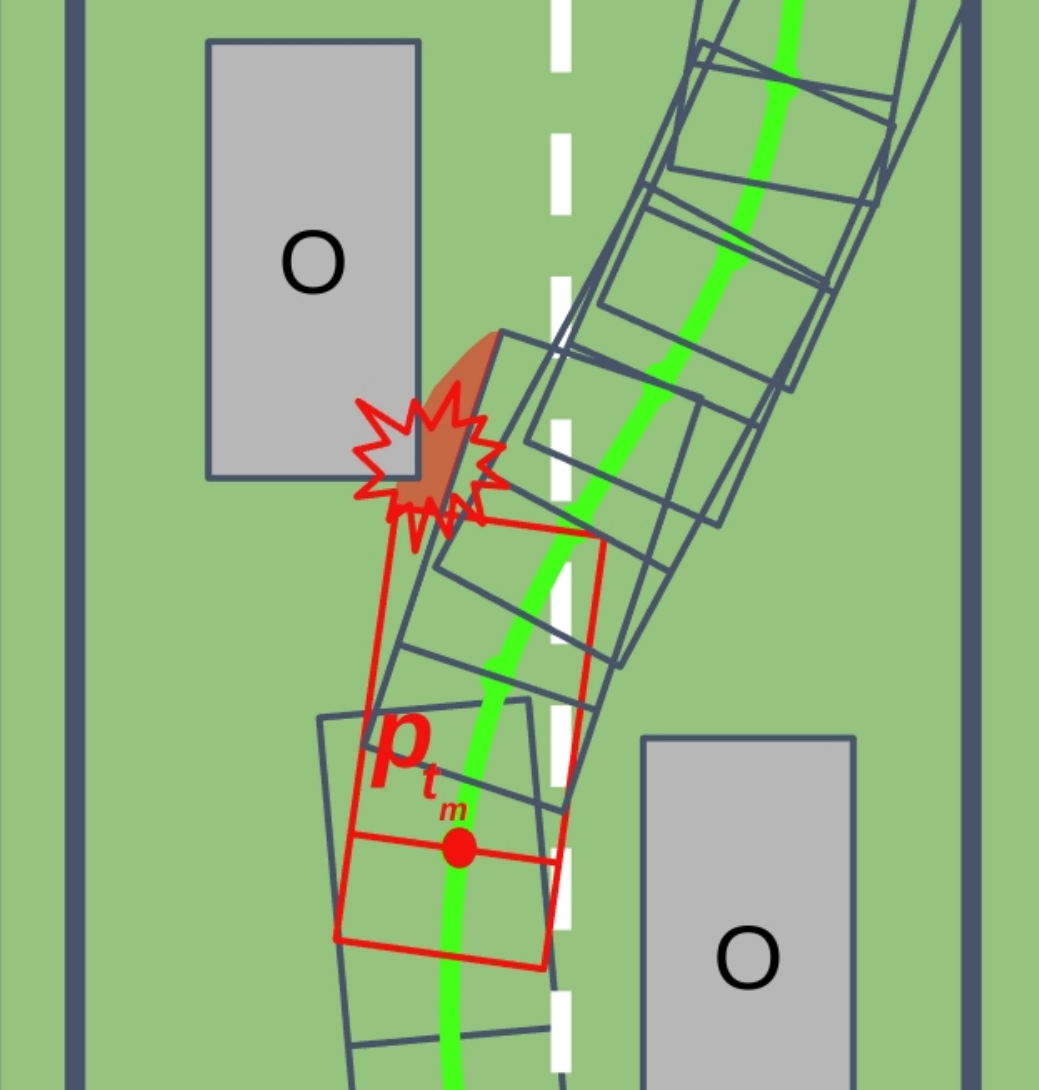}
                \caption{}
                \label{fig:IPF2}
        \end{subfigure}%
        \hfill
        \begin{subfigure}[b]{0.22\textwidth}
                \includegraphics[width=\linewidth]{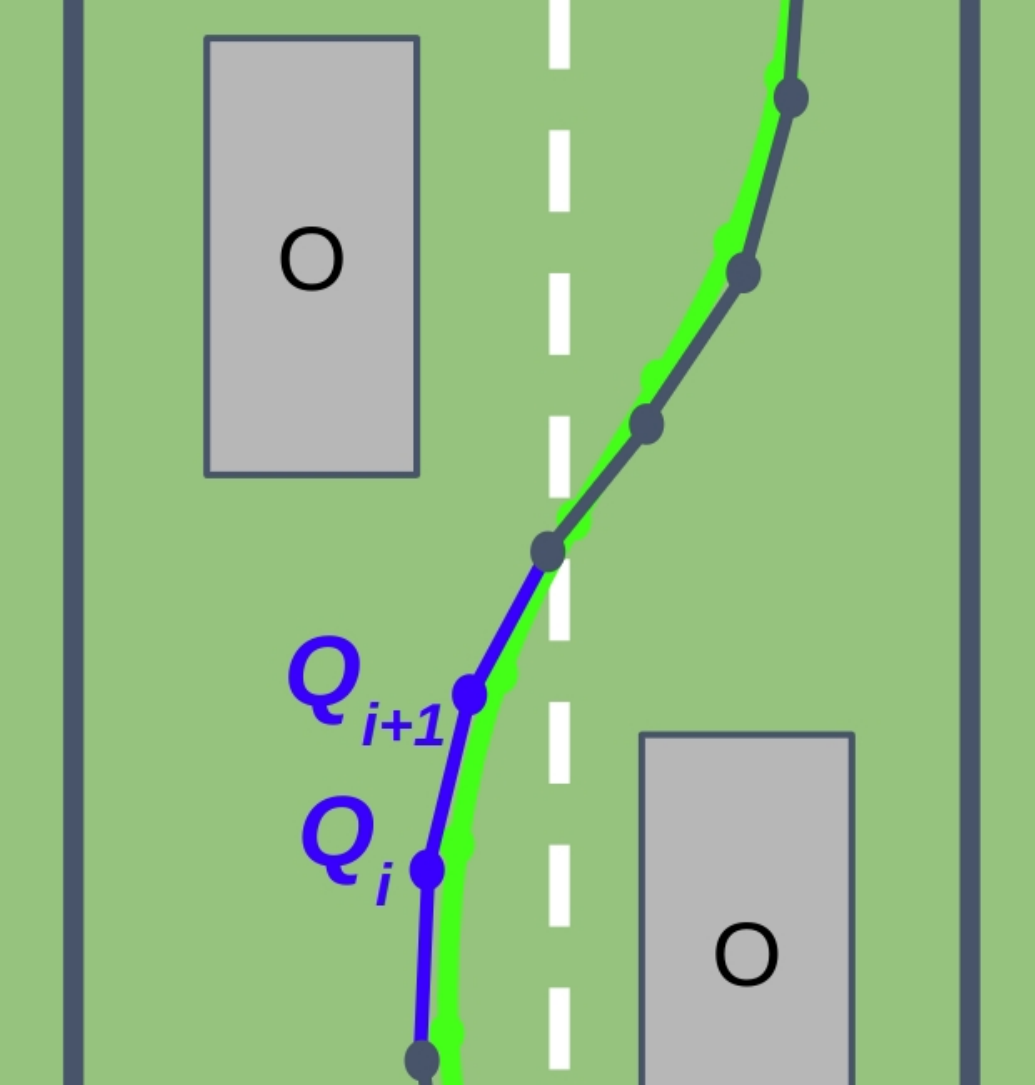}
                \caption{}
                \label{fig:IPF3}
        \end{subfigure}%
        \hfill
        \begin{subfigure}[b]{0.22\textwidth}
                \includegraphics[width=\linewidth]{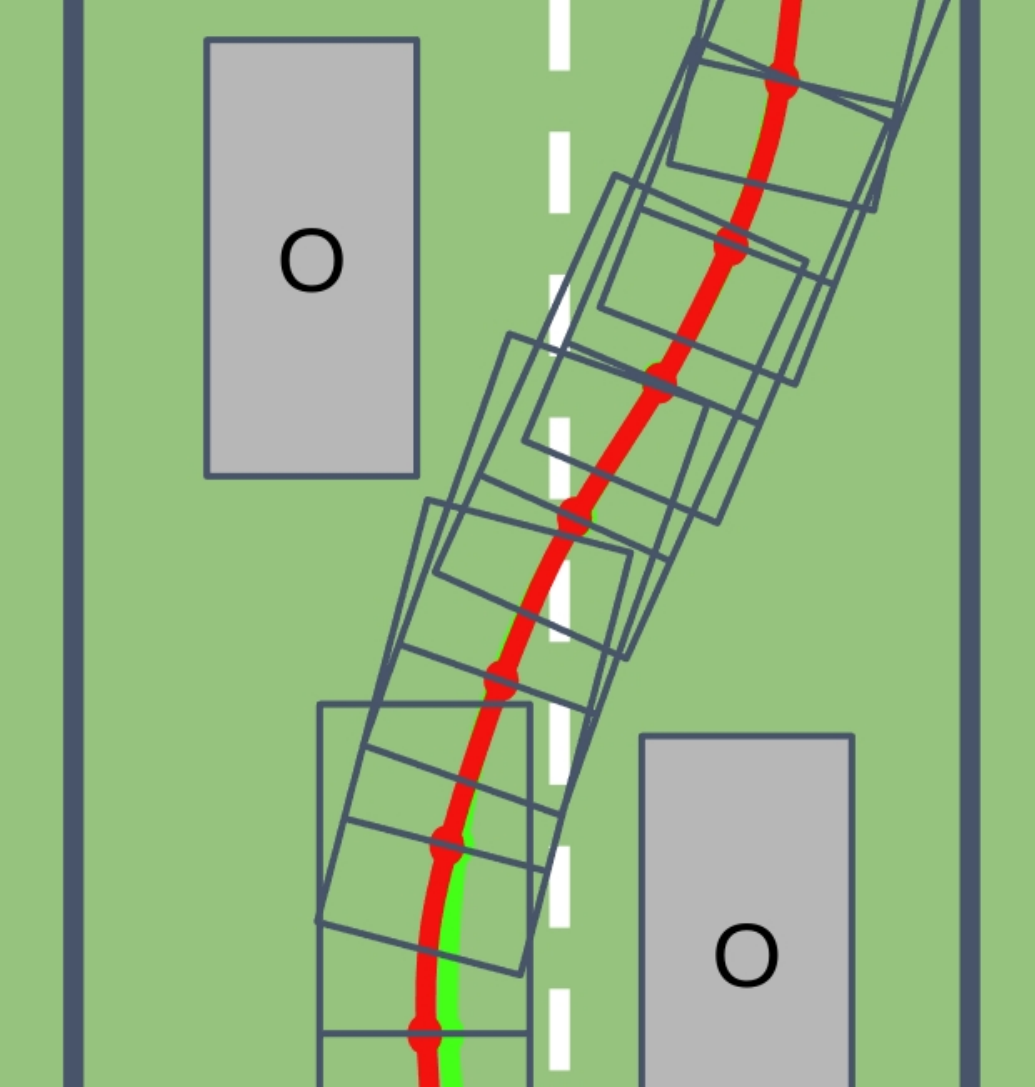}
                \caption{}
                \label{fig:IPF4}
        \end{subfigure}%
        % \hfill
        % \begin{subfigure}[b]{0.18\textwidth}
        %         \includegraphics[width=\linewidth]{pic/4./4.IPF5.pdf}
        %         \caption{Non-Collision}
        %         \label{fig:IPF_Non-Collision}
        % \end{subfigure}
        \caption{\acrshort{ipf} process. (a) No collision occurs along the optimized trajectory, which has a disc radius $\textsc{R}_\textsc{CIRCLE}$. (b) The optimized trajectory has a collision, as determined via \acrshort{sv} collision checking, at a time knot $t_m$. (c) The flattening control points $\mathbf{Q}_i, i \in \mathbf{\Omega}^{fl}$ are found. (d) A collision-free path is found via \acrshort{ipf}.}\label{fig:IPF_process}
\end{figure*}

As shown in Fig. \ref{fig:coll_penalty_inside}, \acrshort{ego-planner} generates $\{\mathbf{p}_{ij}, \mathbf{v}_{ij}\}$, where $\mathbf{p}_{ij}$ is an anchor point and $\mathbf{v}_{ij}$ is a repulsive direction vector, for a collision segment of a trajectory in an iteration. In the original \acrshort{ego-planner}, the quadrotor was modeled as a single circle. Therefore, as shown in Fig. \ref{fig:IPF1}, only obstacles that collided with the optimized trajectory's $\textsc{R}_\textsc{CIRCLE}$ were considered for penalty calculation. However, since our approach models the rectangular-shaped \acrshort{av} as multiple small discs, there can be collisions even if there is no collision with $\textsc{R}_\textsc{CIRCLE}$, as illustrated in Fig. \ref{fig:IPF2}. Consequently, we have introduced a close obstacle collision penalty that pushes the \acrshort{av} away from the close obstacles, as depicted in Fig. \ref{fig:coll_penalty_outside}. In addition, it works with the \acrshort{ipf} method to find a collision-free path. The collision penalty is 

\begin{equation}\label{eq:collision_penalty}
\begin{aligned}
J_{cl} = \sum_{i=p_b}^{N_c-p_b}\sum_{j=1}^{N_p} j_{cl}(i,j),
\end{aligned}
\end{equation}

where $j_{cl}(i,j)$ is a twice continuously differentiable penalty function in collision penalty from \cite{zhou2020ego-planner} and $N_p$ is the number of $\{\mathbf{p}_{ij}, \mathbf{v}_{ij}\}$ pairs.

\subsubsection{Flattening Penalty}
The flattening penalty can locally flatten a path around vehicle collision points to find a collision-free path by reducing the \acrshort{c-sv}. It is formulated as

\begin{equation}\label{eq:flattening_penalty}
\begin{aligned}
J_{fl} = \sum_{i \in \mathbf{\Omega}^{fl}} w_i^{fl} \Bigl(\frac{{K_i}}{\kappa_\text{max}}\Bigl)^2,
\end{aligned}
\end{equation}

where $\mathbf{\Omega}^{fl}$ is a flattening index set (Section \ref{section:ipf}) and $w_i^{fl}$ is a local flattening weight at the control point $\mathbf{Q}_i$. It locally reduces the curvature $K_i$ by increasing the flattening weight.

% \begin{figure}[t]
%     \centering
%     \includegraphics[scale=1, height=5cm]{pic/4./4.IPF_collision.pdf}
%     \caption{Circle collision avoidance constraint with a disc radius $\textsc{R}_\textsc{CIRCLE}$. This constraint (orange arc) cannot fully cover the kinematics collision avoidance constraints (green arc). An obstacle (gray) is placed between the circle and the kinematics collision avoidance constraints.}
%     \label{fig:IPF_collision}
% \end{figure}

\begin{algorithm}[t]
\caption{Rebound Optimization with IPF}\label{alg:Rebound_optimization_with_IPF}
\textbf{Input: } Collision-free trajectory $\bm{\Gamma}$ \\
\textbf{Output: } Optimized trajectory $\bm{\Phi}^{opt}$
\begin{algorithmic}[1]

\State $\mathbf{Q} \gets \Call{P\textnormal{arameterize}T\textnormal{o}B\textnormal{spline}}{\bm{\Gamma}}$
\State $\bm{C} \gets \Call{G\textnormal{en}C\textnormal{ollision}V\textnormal{ector}}{\mathbf{Q}}$
%\State $\mathbf{\Omega}^{fl} \gets \Call{F\textnormal{ind}F\textnormal{lat}P\textnormal{oints}I\textnormal{ndex}S\textnormal{et}}{\bm{\Phi}^{ref}}$
%\State $w_{\kappa, i} \gets \gamma \cdot w_{\kappa, i} , i \in \mathbf{\Omega}^{fl}$
\While {$iter \leq \text{iter}_\text{max}$}
    %\If {$\Call{\textnormal{is}C\textnormal{onverged}}{ }$}
    \State $\bm{\Phi}^{opt} \gets \Call{U\textnormal{niform}B\textnormal{spline}}{\mathbf{Q}}$
    \If {$\Call{\textnormal{is}C\textnormal{ircle}C\textnormal{ollision}}{\bm{\Phi}^{opt}}$}
        \State $\bm{C} \gets \Call{G\textnormal{en}C\textnormal{ollision}V\textnormal{ector}}{\mathbf{Q}}$
         
    \ElsIf {$\Call{\textnormal{is}S\textnormal{v}C\textnormal{ollision}}{\bm{\Phi}^{opt}}$}
        \State $\mathbf{\Omega}^{fl} \gets \Call{F\textnormal{ind}F\textnormal{lattening}I\textnormal{ndex}S\textnormal{et}}{\bm{\Phi}^{opt}}$
        \State $w_i^{fl} \gets \gamma^{fl} \cdot w_i^{fl} , i \in \bm{\Omega}^{fl}$
    \Else 
        \State \Return true
    \EndIf
    %\EndIf
    \State $(J, \bm{G}) \gets \Call{E\textnormal{valuate}P\textnormal{enalty}}{\mathbf{Q}}$
    \State $\mathbf{Q} \gets \Call{T\textnormal{rajectory}O\textnormal{ptimize}}{J, \bm{G}}$
    \State $iter \gets iter + 1$
\EndWhile

\end{algorithmic}
\end{algorithm}
    
\subsection{\acrshort{ipf}} \label{section:ipf}
As mentioned in collision penalty (Section \ref{subsubsection: collision penalty}), to reduce the \acrshort{sv} of the rectangular-shaped \acrshort{av}, we model the \acrshort{av} using multiple small discs. Consequently, a close obstacle collision penalty is required to mitigate the collision risk of the \acrshort{av}. Additionally, to further reduce the collision risk on curves, we identify the control points around the collision points (Fig. \ref{fig:IPF3}) through \acrshort{sv} collision checking. By reducing the curvature in these segments, the path is flattened, making the \acrshort{sv}s smaller and allowing the vehicle to navigate through narrower paths (Fig. \ref{fig:IPF4}).

A collision time lies in a knot span $[t_m, t_{m+1})$. Therefore, a collision index set of the knot vector that can be found by the \acrshort{sv} collision checking is defined as 

\begin{equation}\label{eq:collision_checking_function}
\begin{aligned}
\bm{\Omega}^{coll} &= \{m \in \bm{H} \mid dist(f_{disc,j}(\mathbf{\Phi}(t_m))) \leq R_{disc,j} \}, \\
\bm{H} &= \{p_b, p_b + 1, \ldots, M - p_b\},
\end{aligned}
\end{equation}
% $R_{DISC,\smallcircled{k}}$ ($k \in \{\circled{1}, \circled{2}, \circled{3} \}$
where $j \in \{O, \bm{Z}\}$; $\bm{Z}$ is the first, second and third zones from the \acrshort{c-sv}, $f_{disc,O}$ and $f_{disc,v}$ are the center point functions of the \acrshort{d-sv} and \acrshort{c-sv} discs, respectively, and $dist(\bm{p}_t)$ is a function that determines the distance from $\bm{p}_t$ to the closest obstacle. \textit{KD-tree} \cite{kd_tree_rusu20113d} is used for the $dist(\bm{p}_t)$ function in this work. The curvature of the B-spline curve in the knot span $[t_m, t_{m+1})$ is always smaller than the curvature $max\{|K_{m-p_b + 1}|, |K_{m-p_b+2}|\}$ of the clamped B-spline curve. Therefore, the collision time index set is extended to a flattening index set as

\begin{equation}\label{eq:collision_time_span_index_set}
\begin{aligned}
\bm{\Omega}^{fl} = \{m-p_b+1, m-p_b+2 \mid m \in \bm{\Omega}^{coll} \}. %\{ d_Q(\mathbf{p}_{t_m}) \mid t \leq t_m \leq t^{'}, \int_{t \in \mathbf{\Omega}^{coll}}^{t^{'}} \mathbf{\Phi}(t) dt = \textsc{L}_\textsc{FLAT} \},
\end{aligned}
\end{equation}

The \acrshort{ipf} process iteratively increases the flattening weights, allowing it to find a collision-free trajectory. The \acrshort{ipf} algorithm is presented as Algorithm \ref{alg:Rebound_optimization_with_IPF}. Here, we set three terminal conditions for TrajectoryOptimize(): when the optimization gradient is lower than $\epsilon_g$, when the amount of the cost change is lower than $\epsilon_c$, or at every $\alpha$ iteration of the trajectory optimization for early exit. If isCircleCollision(), which is the circle collision checking function with $\textsc{R}_\textsc{CIRCLE}$ on the optimized trajectory, is true, then collision vectors are generated. If isSvCollision(), which is the \acrshort{sv} collision checking function with \acrshort{d-sv} and \acrshort{c-sv} discs, is true, then the flattening weight $w^{fl}_{i}$ is iteratively increased such that the local flattening weight ratio $\gamma^{fl} > 1$ and $i \in \bm{\Omega}^{fl}$.

\begin{algorithm}[t]
\caption{Refinement Optimization}\label{alg:Refinement_optimization}
\textbf{Input: } Re-allocated control points $\mathbf{Q}^{'}$ \\
\textbf{Output: } Optimized trajectory $\bm{\Phi}^{opt}$
\begin{algorithmic}[1]

%\State $\mathbf{Q} \gets \Call{G\textnormal{et}C\textnormal{ontrol}P\textnormal{oints}}{\bm{\Phi}}$
%\State $\bm{C} \gets \Call{G\textnormal{en}C\textnormal{ollision}V\textnormal{ector}}{\mathbf{Q}}$
%\State $\mathbf{\Omega}^{fl} \gets \Call{F\textnormal{ind}F\textnormal{lat}P\textnormal{oints}I\textnormal{ndex}S\textnormal{et}}{\bm{\Phi}^{ref}}$
%\State $w_{\kappa, i} \gets \gamma \cdot w_{\kappa, i} , i \in \mathbf{\Omega}^{fl}$
\While {$iter \leq \text{iter}_\text{max}$}
    %\If {$\Call{\textnormal{is}C\textnormal{onverged}}{ }$}
    \State $\bm{\Phi}^{opt} \gets \Call{U\textnormal{niform}B\textnormal{spline}}{\mathbf{Q}^{'}}$
    \If {$\Call{\textnormal{is}C\textnormal{ollision}}{\bm{\Phi}^{opt}}$}
        \State $\lambda_{ft} \gets \gamma^{ft} \cdot \lambda_{ft}$
         
    \ElsIf {$\Call{\textnormal{is}F\textnormal{easible}}{\bm{\Phi}^{opt}}$}
        \State \Return true
    \EndIf
    %\EndIf
    \State $(J, \bm{G}) \gets \Call{E\textnormal{valuate}P\textnormal{enalty}}{\mathbf{Q}^{'}}$
    \State $\mathbf{Q}^{'} \gets \Call{T\textnormal{rajectory}O\textnormal{ptimize}}{J, \bm{G}}$
    \State $iter \gets iter + 1$
\EndWhile

\end{algorithmic}
\end{algorithm}

\begin{algorithm}[t]
\caption{B-spline Trajectory Optimization}\label{alg:B-spline_TO}
\textbf{Input: } Collision-free trajectory $\bm{\Gamma}$ \\
\textbf{Output: } Optimized trajectory $\bm{\Phi}^{opt}$
\begin{algorithmic}[1]

\State $\bm{\Phi}^{opt} \gets \Call{R\textnormal{ebound}O\textnormal{ptimization}W\textnormal{ith}IPF}{\bm{\Gamma}}$
\IfNot {$\Call{\textnormal{is}F\textnormal{easible}}{\bm{\Phi}^{opt}}$}
    \State $\bm{\Phi}^{'} \gets \Call{R\textnormal{e}A\textnormal{llocation}T\textnormal{ime}}{\bm{\Phi}^{opt}}$
    \State $\mathbf{Q}^{'} \gets \Call{G\textnormal{et}C\textnormal{ontrol}P\textnormal{oints}}{\bm{\Phi}^{'}}$
    %\State $\mathbf{Q} \gets \Call{P\textnormal{arameterize}T\textnormal{o}B\textnormal{spline}}{\bm{\Phi}}$
    %\State $\bm{\Phi}^{'} \gets \Call{U\textnormal{niform}B\textnormal{spline}}{\mathbf{Q}}$
    \State $\bm{\Phi}^{opt} \gets \Call{R\textnormal{efinement}O\textnormal{ptimization}}{\mathbf{Q}^{'}}$
\EndIf\\
\Return{$\bm{\Phi}^{opt}$}

\end{algorithmic}
\end{algorithm}

\subsection{Trajectory Refinement}
Since the trajectory is modified after the first stage, the initial time allocation may not be accurate, necessitating the trajectory refinement stage to generate a feasible trajectory. Therefore, if the collision-free trajectory obtained from the first stage violates kinodynamic feasibility constraints such as velocity, acceleration, or curvature, the limits exceeding ratio is first calculated using the maximum velocity and acceleration. Time reallocation is then performed to ensure that the velocity and acceleration are within the constraints by relaxing the time. Optimization is carried out using the kinodynamic feasibility penalty to satisfy these constraints. However, since a collision-free trajectory is already available from the first stage, the collision penalty is not used. Instead, if a collision occurs during optimization, the weight of the fitness penalty, which pulls the trajectory towards the collision-free trajectory, is incrementally increased to ensure a collision-free result.

% The initial time allocation of the initial trajectory is inaccurate because the trajectory can be largely modified by obstacles. This modification can make the trajectory violate the feasibility constraints, such as velocity and acceleration. Therefore,
Researches \cite{zhou2020ego-planner, sfc_liu2017planning} relax the violation criterion through time reallocation by lengthening the knot spans with the maximum exceeding ratio of the derivative limits to the axes of all knot spans. The limits exceeding ratio is $r_e = max\{1, |\mathbf{V}_r / v_\text{max}|, \sqrt[2]{|\mathbf{A}_r / a_\text{max}|}, \sqrt[3]{|\mathbf{J}_r / j_\text{max}|} \}$, where $r \in \{x,y,z\}$ axis. Therefore, the limits exceeding ratio relaxes the velocity as $\mathbf{V}_r^{'} = \mathbf{V}_r / r_e$ and the acceleration as $\mathbf{A}_r^{'} = \mathbf{A}_r / r_e$. Since the L2 norm is $\|\mathbf{x}\|=\sqrt{x_1^2 + x_2^2}$, the velocity and acceleration norms can be also relaxed as $\|\mathbf{V}^{'}\| = \|\mathbf{V}\| / r_e$ and $\|\mathbf{A}^{'}\| = \|\mathbf{A}\| / r_e$, respectively. Therefore, the limits exceeding ratio from the longitudinal and lateral velocity and acceleration is

%However, applying the same ratio to all knot spans leads to an inefficient slow-down to other knot spans that is unnecessary. Therefore, we apply different exceeding ratios to each knot span to increase the trajectory efficiency. The exceeding ratios $\mu$ are calculated by
% The limitations of feasibility constraints denotes $c_\text{m} = min\{|c_\text{min}|, |c_\text{max}|\}$ where $c \in \{v, a, j\}$.

\begin{equation}\label{eq:exceeding_ratio}
\begin{aligned}
r_e = max\{1, v_t^s/v_m^s, \sqrt{a_t^s/a_m^s}, \sqrt{a_t^d/a_m^d} \},
\end{aligned}
\end{equation}

where $t \in [t_{p_b}, t_{M-p_b}]$, and the subscript $m$ denotes the maximum if the derivative is greater than or equal to zero and the minimum if the derivative is less than zero. Therefore, starting from the optimized trajectory $\bm{\Phi}^{opt}$, a new trajectory $\bm{\Phi}^{'}$ is obtained through time reallocation ($\Delta t^{'} = r_e \Delta t$) \cite{zhou2020ego-planner}. This new trajectory $\bm{\Phi}^{'}$ maintains the same shape and the same number of control points as $\bm{\Phi}^{opt}$.

% 그러므로 optimized trajectory인 $\bm{\Phi}^{opt}$에서 부터 time reallocation ($\Delta t^{'} = r_e \Delta t$)을 통하여 새로운 $\bm{\Phi}$를 얻게 되고 이는 $\bm{\Phi}^{opt}$와 같은 모양과 같은 수의 control points를 유지하게 됩니다.

% \begin{equation}\label{eq:refinement_cost_function}
% \begin{aligned}
% \Delta t^{'} = r_e \Delta t,
% \end{aligned}
% \end{equation}

The optimization cost function $J^{'}$ for refinement is reformulated as

\begin{equation}\label{eq:refinement_cost_function}
\begin{aligned}
\min_{\mathbf{Q}}J^{'} = \lambda_{sm} J_{sm} + \lambda_{ft} J_{ft} + \lambda_{fs} J_{fs},
\end{aligned}
\end{equation}

where $J_{ft}$ and $J_{fs}$ are fitness and feasibility penalties, respectively, and $\lambda_{ft}$ and $\lambda_{fs}$ are their corresponding weights.

\subsubsection{Fitness Penalty}
% In this trajectory refinement stage, instead of the collision penalty, the fitness penalty ensures that the optimized trajectory is free of collisions. This is because the trajectory refinement does not make significant modifications to the trajectory shape after rebound optimization. It pulls the optimized trajectory toward the collision-free trajectory obtained from the rebound optimization stage, using anisotropic displacements between the two trajectories. We use the fitness penalty function $j_{ft}$ \cite{zhou2020ego-planner}, which is calculated as

In this trajectory refinement stage, instead of the collision penalty, the fitness penalty ensures that the optimized trajectory is free of collisions. The fitness penalty pulls the control points of the optimizing trajectory towards the control points of the existing collision-free trajectory using anisotropic displacements between the two trajectories, as proposed in \cite{zhou2020ego-planner}. Therefore, as shown in Algorithm \ref{alg:Refinement_optimization}, we increased the fitness weight whenever a collision occurred to maintain a collision-free trajectory. We use the fitness penalty function $j_{ft}$ from \cite{zhou2020ego-planner}, where the fitness penalty cost is calculated as

\begin{equation}\label{eq:fitness_penalty}
\begin{aligned}
J_{ft} = \sum_{i=p_b}^{N_c-p_b} j_{ft}(i).
\end{aligned}
\end{equation}

The fitness weight $\lambda_{ft}$ increases by the fitness weight ratio $\gamma^{ft}$ if the path has a collision.

\subsubsection{Kinodynamic Feasibility Penalty}
A kinodynamic feasibility penalty enables \acrshort{av}s to meet the kinodynamic feasibility constraints by confining the derivatives and curvature of the trajectory within their limitations. However, since the kinodynamic feasibility penalty is divided into longitudinal and lateral velocity and acceleration penalties, the convex-hull property cannot be used. Therefore, the feasibility penalty cost function is formulated using an integral expression as

% \begin{equation}\label{eq:feasibility_penalty}
% \begin{aligned}
% J_{fs} = &\sum_{i=1}^{N_c-1} j_{fs}(\frac{\|\mathbf{V}_i\|}{v_\text{max}})
% + \sum_{i=1}^{N_c-2} j_{fs}(\frac{\|\mathbf{A}_i\|}{a_\text{max}})\\
% &+ \sum_{i=1}^{N_c-3} j_{fs}(\frac{\|\mathbf{J}_i\|}{j_\text{max}})
% + \sum_{i={p_b - 1}}^{N_c - p_b + 1} j_{fs}(\frac{K_i}{\kappa_\text{max}}),
% \end{aligned}
% \end{equation}

\begin{equation}\label{eq:feasibility_penalty}
\begin{aligned}
J_{fs} = &\int_{t_{p_b}}^{t_{M-p_b}} j_{fs}(v_t^s) dt
+ \int_{t_{p_b}}^{t_{M-p_b}} j_{fs}(a_t^s) dt\\
& + \int_{t_{p_b}}^{t_{M-p_b}} j_{fs}(a_t^d) dt
+ \sum_{i={1}}^{N_c - 1} j_{fs}(K_i),
\end{aligned}
\end{equation}

where $j_{fs}$ is a twice continuously differentiable function similar to the feasibility penalty of \cite{zhou2020ego-planner}. We use the Gauss\textendash Legendre n-point quadrature formula \cite{gauss-legendre_press1988numerical} to calculate the integral in (\ref{eq:feasibility_penalty}).

\begin{equation}\label{eq:feasibility_cost_function}
\begin{aligned}
j_{fs}(c) = \left\{ \begin{array}{lr} a_1(c/c_{m})^2+b_1(c/c_{m})+c_1 & (c/c_{m} \geq 1) 
\\(c/c_{m}-\lambda)^3 & (\lambda < c/c_{m} < 1) 
\\ 0 & (-\lambda \leq c/c_{m} \leq \lambda)
\\(\lambda-c/c_{m})^3 & (-1 < c/c_{m} < -\lambda)
\\a_2(c/c_{m})^2+b_2(c/c_{m})+c_2 & (c/c_{m} \leq -1) \end{array}\right. ,
\end{aligned}
\end{equation}

% \begin{equation}\label{eq:feasibility_cost_function}
% \begin{aligned}
% j_{fs}(c_i) = \left\{ \begin{array}{lr} ac_i^2+bc_i+c & (c_i \geq c_\text{max}) 
% \\(c_i-\lambda c_\text{max})^3 & (\lambda c_\text{max} < c_i < c_\text{max}) 
% \\ 0 & (0 \leq c_i \leq \lambda c_\text{max}) \end{array}\right. ,
% \end{aligned}
% \end{equation}

% \begin{equation}\label{eq:feasibility_cost_function}
% \begin{aligned}
% j_{fs}(c_i) = \left\{ \begin{array}{lr} a_1c_i^2+b_1c_i+c_1 & (c_i \geq c_\text{max}) 
% \\(c_i-\lambda c_\text{max})^3 & (\lambda c_\text{max} < c_i < c_\text{max}) 
% \\ 0 & (\lambda c_\text{min} \leq c_i \leq \lambda c_\text{max}) 
% \\ (\lambda c_\text{min} - c_i)^3 & ( c_\text{min}< c_i < \lambda c_\text{min}) 
% \\ a_2c_i^2+b_2c_i+c_2 & (c_i \leq c_\text{min}) \end{array}\right. ,
% \end{aligned}
% \end{equation}

where $a_1,b_1,c_1,a_2,b_2$, and $c_2$ are selected for second-order continuity; $0 < \lambda \leq 1$ is an elastic coefficient; and the subscript $m$ denotes the maximum if $c \geq 0$ and the minimum if $c < 0$. The refinement optimization algorithm, shown as in Algorithm \ref{alg:Refinement_optimization}, starts with the reallocated control points from the trajectory optimized during rebound optimization using \acrshort{ipf} if the isFeasible() function is false. This function checks whether the kinodynamic feasibility constraints (velocity, acceleration, and curvature) are violated. If isCollision() is true, which is the case if either isCircleCollision() or isSvCollision() is true, then the fitness weight is iteratively increased by $\gamma^{ft}$ to create a collision-free trajectory using the fitness penalty. The terminal conditions for TrajectoryOptimize() are derived from Algorithm \ref{alg:Rebound_optimization_with_IPF}. The entire B-spline trajectory optimization algorithm is presented as Algorithm \ref{alg:B-spline_TO}.

%Since this problem needs repeatedly stopping, generating new costs for obstacles, and restarting, a fast restarting solver is required. According to the experiment results from \cite{zhou2020ego-planner}, \acrfull{l-bfgs} method, which is able to restart fast, outperforms other quasi-Newton methods such as Barzilai-Borwein \cite{quasi-Newton_barzilai1988two} and truncated Newton \cite{quasi-Newton_dembo1983truncated} method. Therefore, the \acrfull{l-bfgs} solver is used in this paper with Lewis-Overton line search \cite{lewis2013nonsmooth} for non-convex optimization.

%%%%%%%%%%%%%%%%%%%%%%%%%%%%%%%%%%%%%%%%%%%%%%%%%%%%%%%%%%%%%%%%%%%%%%%%%%%%%%%%
\section{Evaluations}
\label{section:evaluation}
\begin{table}[t]
%\caption{SIMULATION PARAMETER SETTINGS}
\caption{Simulation parameter settings}
\label{table:parameter_settings}
\begin{center}
\begin{tabular}{c | c | c} 
     \Xhline{2\arrayrulewidth}
     \textbf{Parameter} & \textbf{Description and unit} & \textbf{Value} \\ %[0.5ex] 
     \Xhline{2\arrayrulewidth}
     $p_b$ & Degree of B-spline curve & 3 \\
     \hline
     $\textsc{N}_\textsc{DISC, O}$ & Number of discs for \acrshort{d-sv} & 5 \\
     \hline
     $\textsc{L}_\textsc{F}$ & Front hang length of the ego vehicle ($m$)  & 1.015 \\
     \hline
     $\textsc{L}_\textsc{R}$ & Rear hang length of the ego vehicle ($m$)  & 1.015 \\
     \hline
     $\textsc{L}_\textsc{W}$ & Wheelbase of the ego vehicle ($m$) & 2.87 \\
     \hline
     $\textsc{L}_\textsc{B}$ & Width of the ego vehicle ($m$) & 1.86 \\
     \hline
     $\textsc{L}_\text{min}$ & Minimum length $l_i$ ($m$) & 0.1 \\
     \hline
     $v^s_\text{max}, v^s_\text{min}$ & Max \& Min longitudinal velocity ($m/s$) & 5.55, 0.00 \\
     \hline
     $a^s_\text{max}, a^s_\text{min}$ & Max \& Min longitudinal acceleration ($m/s^2$) & 4.0, -4.0 \\
     \hline
     $a^d_\text{max}, a^d_\text{min}$ & Max \& Min lateral acceleration ($m/s^2$) & 2.0, -2.0 \\
     \hline
     $\kappa_\text{max}, \kappa_\text{min}$ & Max \& Min curvature ($m^{-1}$) & 0.2, -0.2 \\
     \hline
     $\Delta t$ & Time interval ($s$) & $\text{L}_\text{W} / 2 v^s_\text{max}$ \\
     \hline
     $\textsc{R}_\textsc{CIRCLE}$ & Circle disc radius ($m$) & $\textsc{R}_\textsc{DISC,O}$ \\
     \hline
     $s_f$ & Safe clearance ($m$) & $2\textsc{R}_\textsc{DISC,O}$ \\
     \hline
     $\lambda$ & Elastic coefficient & 0.8 \\
     \hline
     $\text{iter}_\text{max}$ & Max rebound \& refinement iteration & 10 \\
     \hline
     $\epsilon_g, \epsilon_c$ & Gradient \& Cost epsilon & 1e-2, 1e-5 \\
     \hline
     $\alpha$ & Early exit iteration number & 100 \\
     \hline
     $s_\text{A}, s_\text{J}$ & Scale parameter for acceleration and jerk & 3.0, 5.0 \\
     \hline
     $\lambda_{sm}$ & Weight for smoothness penalty & 1.0 \\
     \hline
     $\lambda_{cl}$ & Weight for collision penalty & 1.0 \\
     \hline
     $\lambda_{fl}$ & Weight for flattening penalty & 1.0 \\
     \hline
     $\lambda_{ft}$ & Weight for fitness penalty & 2.0 \\
     \hline
     $\lambda_{fs}$ & Weight for feasibility penalty & 5.0 \\
     \hline
     $w_i^{fl}$ & Local flattening weight & 1.0 \\
     \hline
     $\gamma^{fl}$ & Local flattening weight ratio & 10.0 \\ %[1ex] 
     \hline
     $\gamma^{ft}$ & Fitness weight ratio & 2.0 \\ %[1ex] 
     \Xhline{2\arrayrulewidth}
\end{tabular}
\end{center}
\end{table}

The effectiveness of the \acrshort{sv} estimation method, \acrshort{ipf}, and kinodynamic feasibility penalties and the tracking performance, and time efficiency of the proposed algorithm were evaluated. The results demonstrate that our proposal outperformed two \acrlong{sota} trajectory optimization algorithms \cite{DL-IAPS_zhou2020autonomous, curvy_li2022autonomous} in various environments. Path-velocity decoupled and coupled trajectory optimization algorithms, which are designed for car-like vehicles, were utilized as baselines to show the superiority of the proposed method over both algorithm types. A video of the experiments is available at \url{https://youtu.be/iRCl1vtn5dk}.

\subsection{Simulation Setup}
All simulation experiments were conducted on a desktop with a Ryzen 7 3800x CPU, which runs at 3.9GHz by default. We simulated experiments using the open-source simulator \carla \cite{dosovitskiy2017carla}. All parameter settings are listed in Table \ref{table:parameter_settings}. As mentioned in Section \ref{subsubsection:C-sv estimation method}, since $\textsc{N}_\textsc{DISC,O}$ is not large, $\textsc{R}_\textsc{DISC,O}$ is sufficiently large to encompass the second and third zones. Therefore, for time efficiency, we excluded these zones from the \acrshort{c-sv} and demonstrated in our experiments that these zones are adequately covered. A solver, which enables fast restarting, was required because our problem needs repeated stopping, generation of new costs for obstacles, and restarting. According to the experiment results of \cite{zhou2020ego-planner}, the \acrshort{l-bfgs} method, which is capable of fast restarting, outperforms other quasi-Newton methods, such as the Barzilai\textendash Borwein \cite{quasi-Newton_barzilai1988two} and truncated Newton \cite{quasi-Newton_dembo1983truncated} method. Therefore, the \acrshort{l-bfgs} solver was used in this experiment with Lewis\textendash Overton line search \cite{lewis2013nonsmooth} for non-convex optimization. The maximum iterations for the \acrshort{l-bfgs} solver we used were set to 200.

For a reference trajectory initially satisfying terminal constraints mentioned in Section \ref{subsection:B-spline_Primitives}, the hybrid $\text{A}^*$ \cite{hybrid_a_star_dolgov2010path} algorithm was used to generate a kinematically feasible reference path and a trapezoidal velocity profile \cite{B-spline_ding2019efficient, sfc_liu2017planning} for an initial time allocation. The first baseline, called DL-IAPS \cite{DL-IAPS_zhou2020autonomous}, is a path-velocity decoupled planner solved using the QP solver \osqp \cite{osqp_stellato2020osqp}. The second baseline, called C-Planner \cite{curvy_li2022autonomous}, is a path-velocity coupled planner based on \acrshort{ocp} that is solved using the \acrshort{nlp} solver \ipopt \cite{ipopt_wachter2006implementation} with MA27. Additionally, to compare the computational times of state-of-the-art OCP solvers, we used the \acados \cite{acados_verschueren2022acados} with \hpipm \cite{frison2020hpipm} in C-Planner during random tests, creating a new baseline referred to as $\text{C-Planner}^\text{ACD}$. This baseline utilized AS-RTI-D-2 \cite{acados_rti_frey2024advanced} iterations and a partial condensing horizon of $\frac{N}{2}$. The optimization time resolution $\delta t$ was set to $0.1 s$ for DL-IAPS and $0.05s$ for C-Planner.

% A simulated environment that includes large left and right turns, and a narrow corridor is used to analyze the effectiveness of the proposed method introduced in the previous Fig. \ref{section:B-spline_trajectory_optimization}. 

% A reference trajectory depicted in Fig. \ref{fig:analysis_init_opt_compare} is derived from an $\textsc{A}^*$ based path and a trapezoidal velocity with a constant acceleration. Aside from the initial-/goal-heading collision-free condition, the initial trajectory does not need a vehicle kinematic collision-free path but a disc collision-free path. Optimizing the initial trajectory by the \acrshort{ipf} algorithm consumes $61ms$ faster than other \acrshort{sota} algorithms compared in the run time analysis subsection.

\begin{figure}[t]
    \centering
    \includegraphics[scale=1, width=\linewidth]{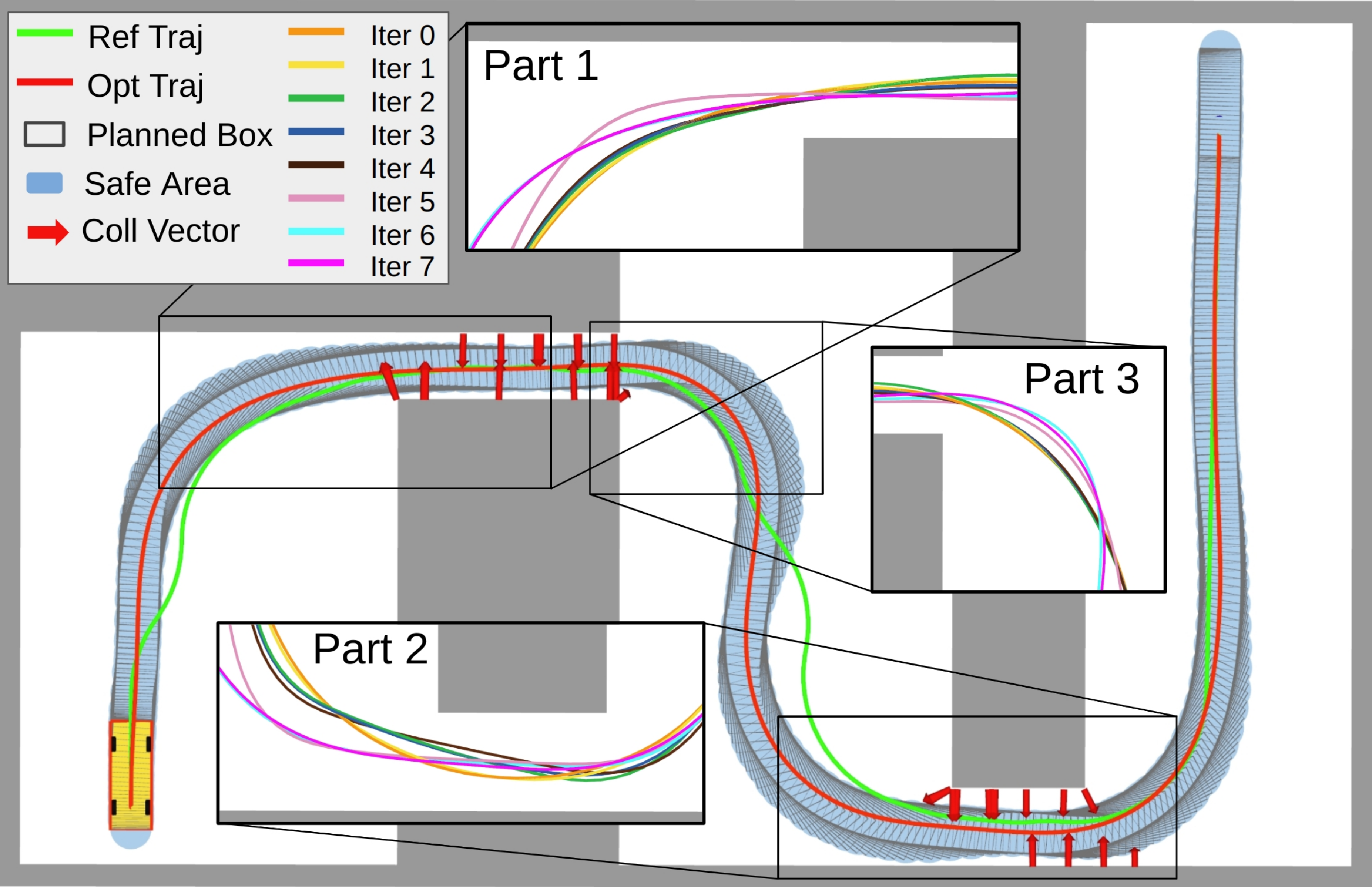}
    \caption{Trajectory optimization result with double narrow corridors and multiple curves. A reference trajectory is generated using the hybrid $\text{A}^*$ algorithm. The zoomed-in images show the optimizing path in each iteration. The planned boxes are generated at every $0.1s$.}
    %IPF:174ms, ref: 47ms
    \label{fig:analysis_init_opt_compare}
\end{figure}

\begin{figure}[t]
    \centering
    \begin{subfigure}[b]{0.38\linewidth}        %% or \columnwidth
        \centering
        \includegraphics[height=7cm, width=\linewidth]{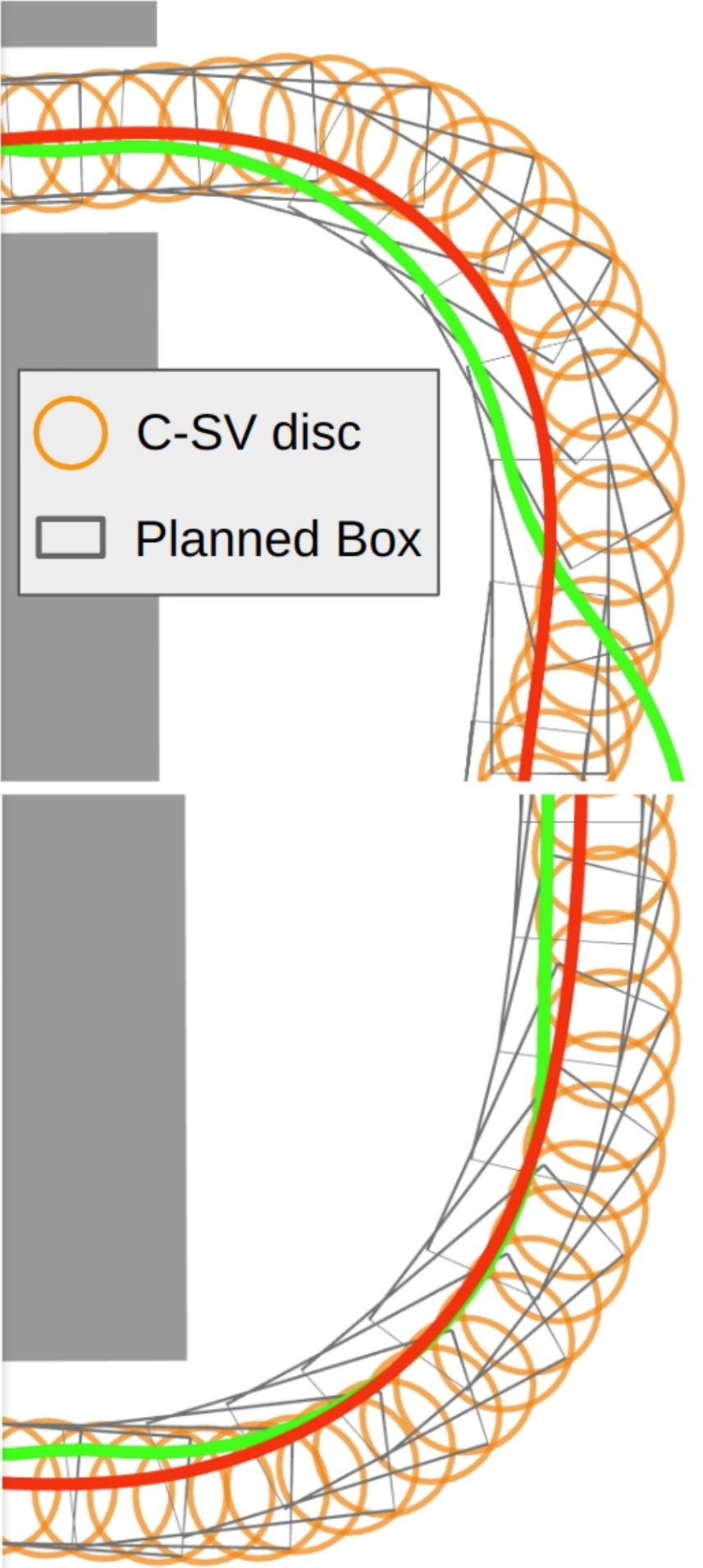}
        \caption{\acrshort{c-sv} discs}
        \label{fig:SV_test1}
    \end{subfigure}
    %\\[1ex]
    \begin{subfigure}[b]{0.38\linewidth}        %% or \columnwidth
        \centering
        \includegraphics[height=7cm, width=\linewidth]{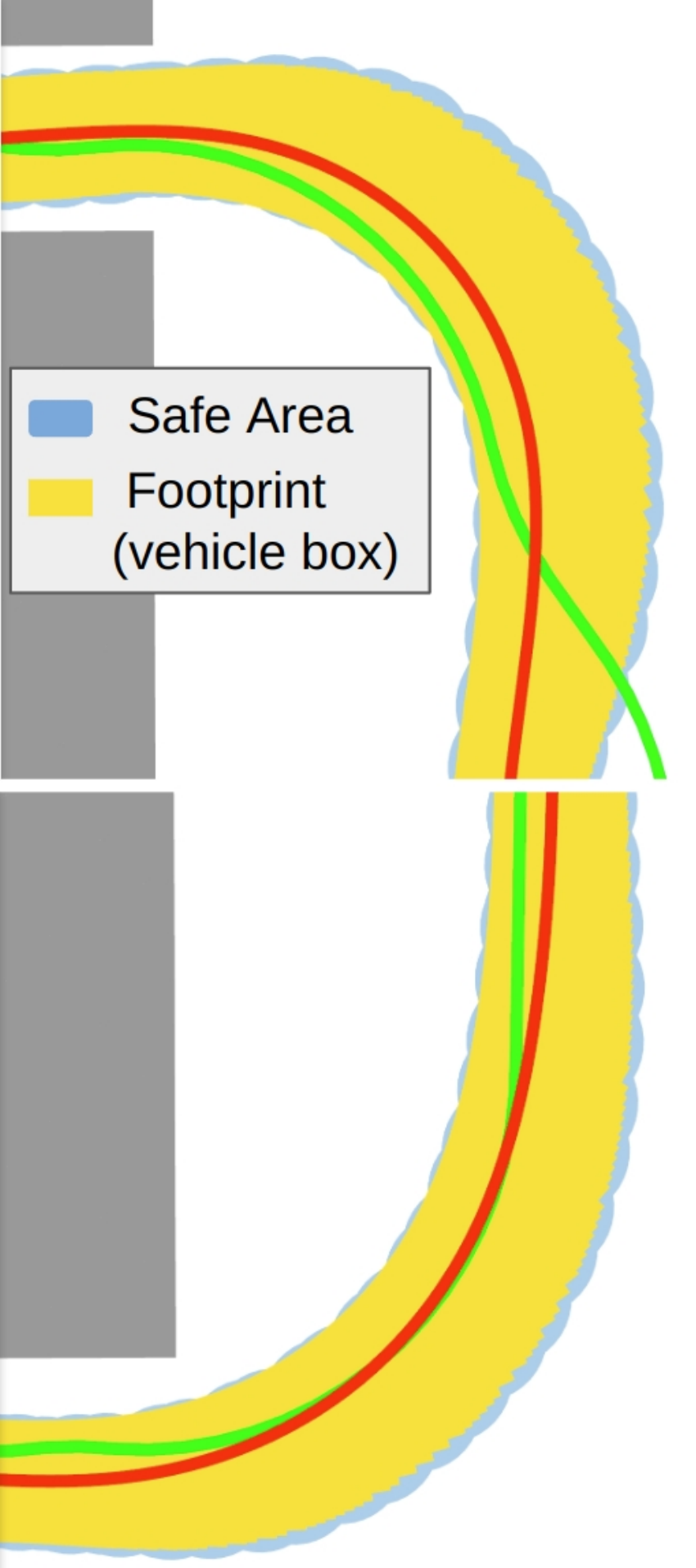}
        \caption{Actual Vehicle \acrshort{sv}}
        \label{fig:SV_test2}
    \end{subfigure}
    \caption{Evaluation of \acrshort{sv} estimation: (a) C-\acrshort{sv} discs and planned vehicle boxes, (b) safe area from \acrshort{sv}s (\acrshort{d-sv}, \acrshort{c-sv}) and actual vehicle's footprint. In (a), the \acrshort{c-sv} discs (orange) cover the outermost edge of the planned vehicle boxes (gray) at every time interval $\Delta t$ around the maximum curvature ($\approx \kappa_\text{max}$) of the path. 
    % The outside discs and inside discs are generated using the outside pairs $\{F_L, F^{'}_L\}$ and $\{R_L, R^{'}_L\}$, respectively. Some inside discs are not generated because the vertices of the pair are not outside of the vehicle box at different time steps. 
    In (b), the safe area (blue) generated using the proposed \acrshort{sv} estimation method can cover the footprint (yellow) of the vehicle box, which is the actual \acrshort{sv} of the vehicle.}
     % The outside discs and inside discs are generated using the outside pairs $\{F_L, F^{'}_L\}$ and $\{R_L, R^{'}_L\}$, respectively. Some inside discs are not generated because the vertices of the pair are not outside of the vehicle box at different time steps.
    \label{fig:SV_test}
\end{figure}

\subsection{\acrshort{sv} Estimation Method}
We evaluated whether the proposed \acrshort{sv} estimation method can cover the actual vehicle \acrshort{sv} between time steps. The evaluation environment (Fig. \ref{fig:analysis_init_opt_compare}) had double narrow corridors and multiple curves, which required a sophisticated trajectory optimization algorithm with tight \acrshort{sv} collision checking. Fig. \ref{fig:SV_test1} shows that the \acrshort{c-sv} discs fully cover the outermost edge, referred to as the first zone in Section \ref{subsubsection:C-sv estimation method}, in the region of maximum curvature. The number of \acrshort{c-sv} discs appropriately increases as the area of the \acrshort{sv} to be covered grows, thereby reducing over-estimation. The safe area in Fig. \ref{fig:SV_test2} combines the areas of \acrshort{d-sv} and \acrshort{c-sv}, ensuring that the actual vehicle's footprint is safely encompassed.

\begin{figure}[t]
    \centering
    \begin{subfigure}[b]{0.48\linewidth}        %% or \columnwidth
        \centering
        \includegraphics[width=\linewidth]{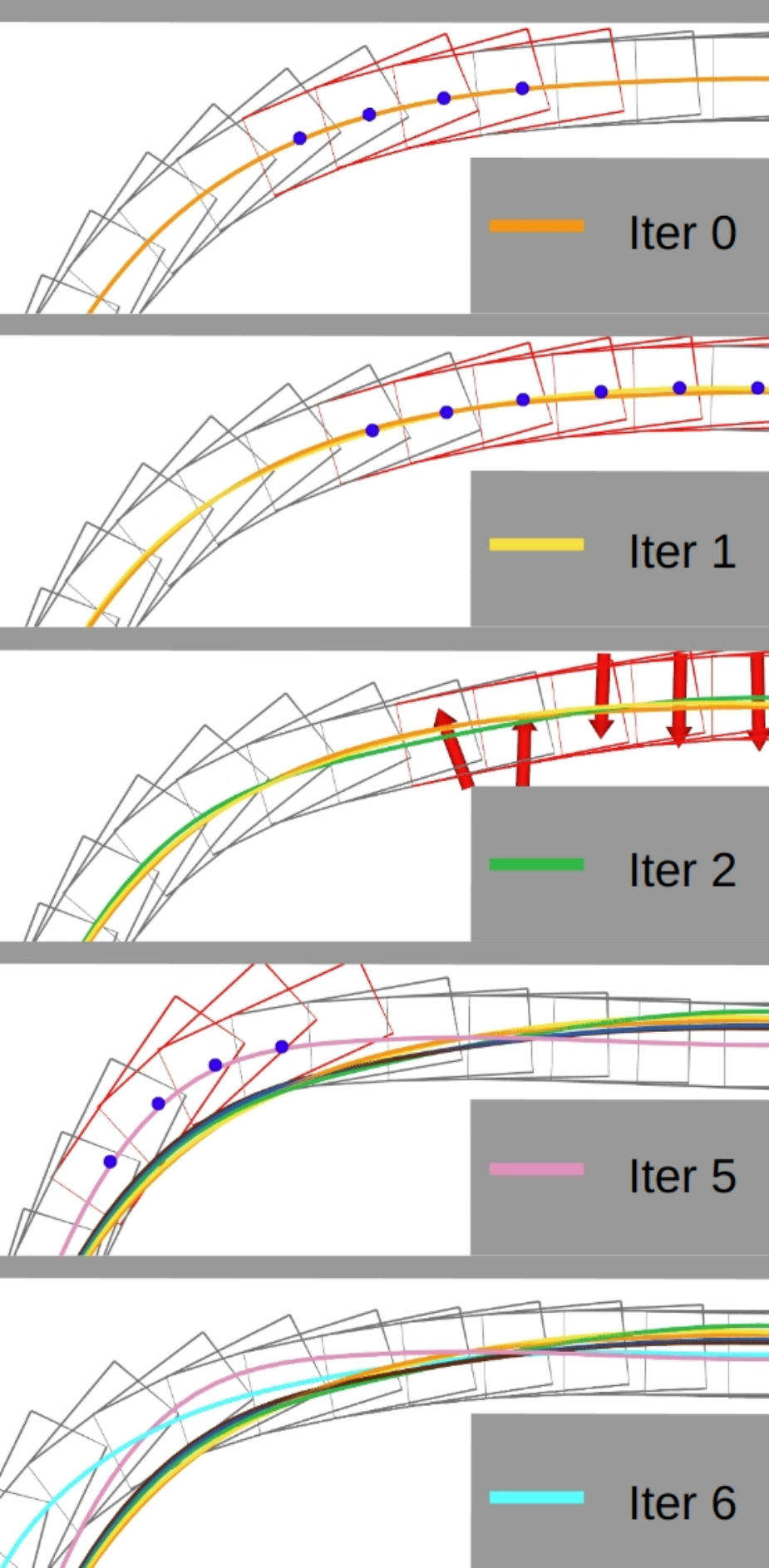}
        \caption{Part 1}
        \label{fig:opt_analysis_IPF1}
    \end{subfigure}
    \begin{subfigure}[b]{0.48\linewidth}        %% or \columnwidth
        \centering
        \includegraphics[width=\linewidth]{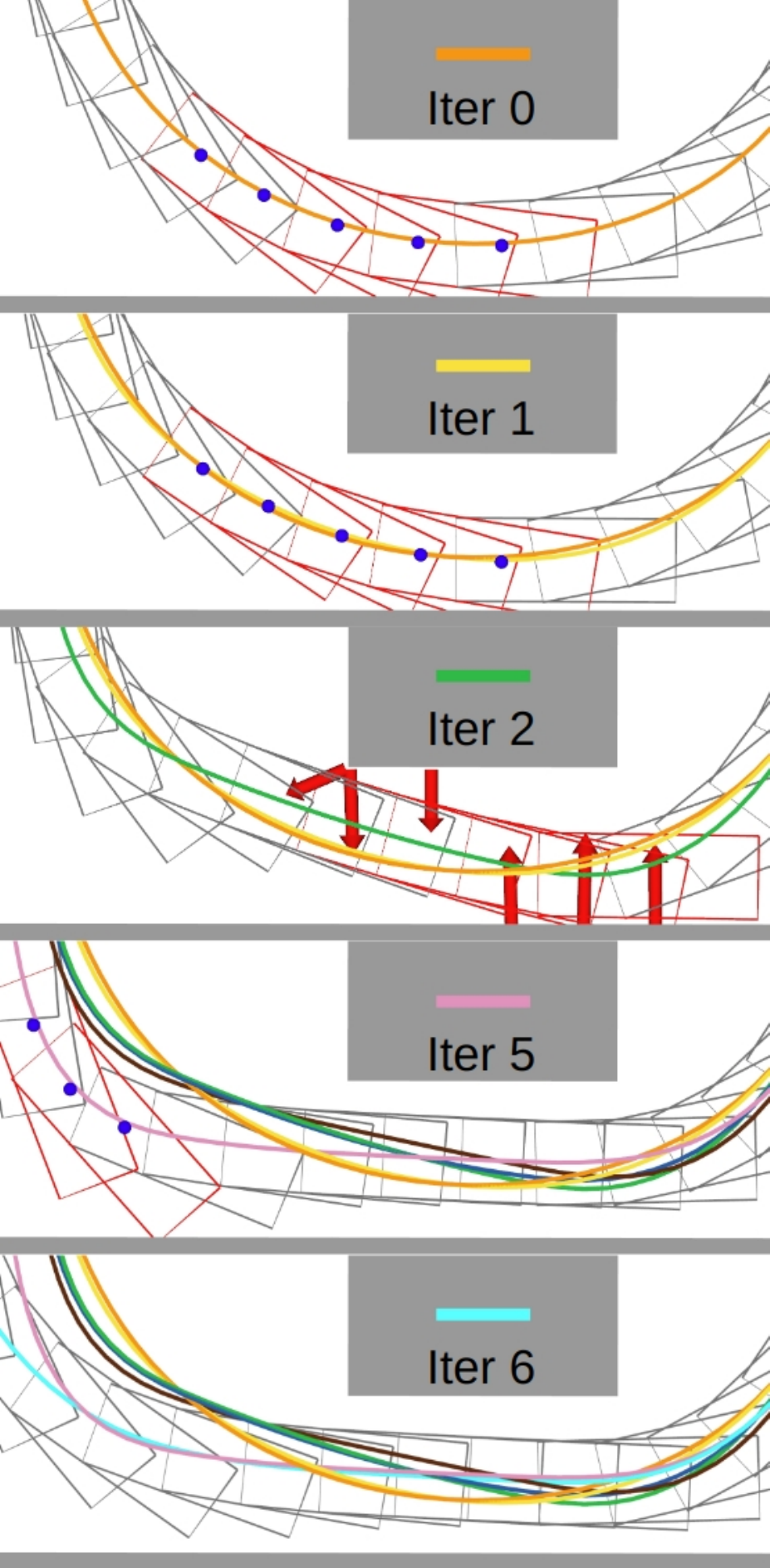}
        \caption{Part 2}
        \label{fig:opt_analysis_IPF2}
    \end{subfigure}
    \caption{Iterations of rebound optimization with \acrshort{ipf}: (a) part 1 in Fig. \ref{fig:analysis_init_opt_compare}, (b) part 2 in Fig. \ref{fig:analysis_init_opt_compare}. Planned collision-free vehicle boxes (gray) and planned collision vehicle boxes (red) are generated at every time interval $\Delta t$. The blue dots represent the flattening points $\mathbf{\Omega}^{flat}$, and the red arrows show the collision penalties from the nearby obstacles.}
    \label{fig:opt_analysis_for_IPF}
\end{figure}

% \begin{figure}[t]
%     \centering
%     \begin{subfigure}[b]{0.48\linewidth}        %% or \columnwidth
%         \centering
%         \includegraphics[width=\linewidth]{pic/5.IPF_analysis_wo_IPF.pdf}
%         \caption{without \acrshort{ipf}}
%         \label{fig:opt_analysis_wo_IPF}
%     \end{subfigure}
%     \begin{subfigure}[b]{0.48\linewidth}        %% or \columnwidth
%         \centering
%         \includegraphics[width=\linewidth]{pic/5.IPF_analysis_with_IPF.pdf}
%         \caption{with \acrshort{ipf}}
%         \label{fig:opt_analysis_with_IPF}
%     \end{subfigure}
%     \caption{Trajectory optimization with each iteration is shown. Gray boxes are planned collision-free vehicle boxes and red boxes are planned collision vehicle boxes. (a) The iter 0 path is too far from the obstacle to generate a collision penalty ($d(p_t) > s_f$). (b) The flattening points $\mathbf{\Omega}^{flat}$ are shown and the path is getting flattened through iterations.}
%     \label{fig:opt_analysis_for_IPF}
% \end{figure}

\subsection{\acrshort{ipf}}
The \acrshort{ipf} method was evaluated in the same environment as the \acrshort{sv} estimation method because the obstacle walls at the entrances of the narrow corridors made it difficult to generate a feasible path. Fine changes had to be made in the path before these entrances; otherwise, trajectory optimization would be rendered infeasible by the \acrshort{sv} collision checking. Therefore, we evaluated the \acrshort{ipf} method by analyzing all iterations of the trajectory optimization algorithm, as shown in Fig. \ref{fig:analysis_init_opt_compare}. The iterations at the entrances of the narrow corridors are enlarged in the figure (parts 1 and 2). Iterations 0-6 were from rebound optimization with \acrshort{ipf}, and iteration 7 was from refinement optimization. 

Each iteration of rebound optimization with \acrshort{ipf} is illustrated in Fig. \ref{fig:opt_analysis_for_IPF}. Iterations 0, 1, and 5 in both parts generated flattening points due to the \acrshort{sv} collision, and no circle collision occurred. Then, the path of the next iterations 1, 2, and 6 was flattened near the vehicle collision points by increases in the flattening weights at the flattening points. However, this flattened path could collide with nearby obstacles, as shown in iteration 2. The collision penalty from the close obstacles pushed the path away from nearby obstacles. Consequently, \acrshort{ipf} works in conjunction with the close obstacle collision penalty to find a collision-free trajectory.

\begin{figure}[t]
    \centering
    %\includegraphics[scale=1, height=5cm, width=\linewidth]{pic/5.refinement_analysis.pdf}
    % height=4.5cm, width=0.8\linewidth
    \includegraphics[width=\linewidth]{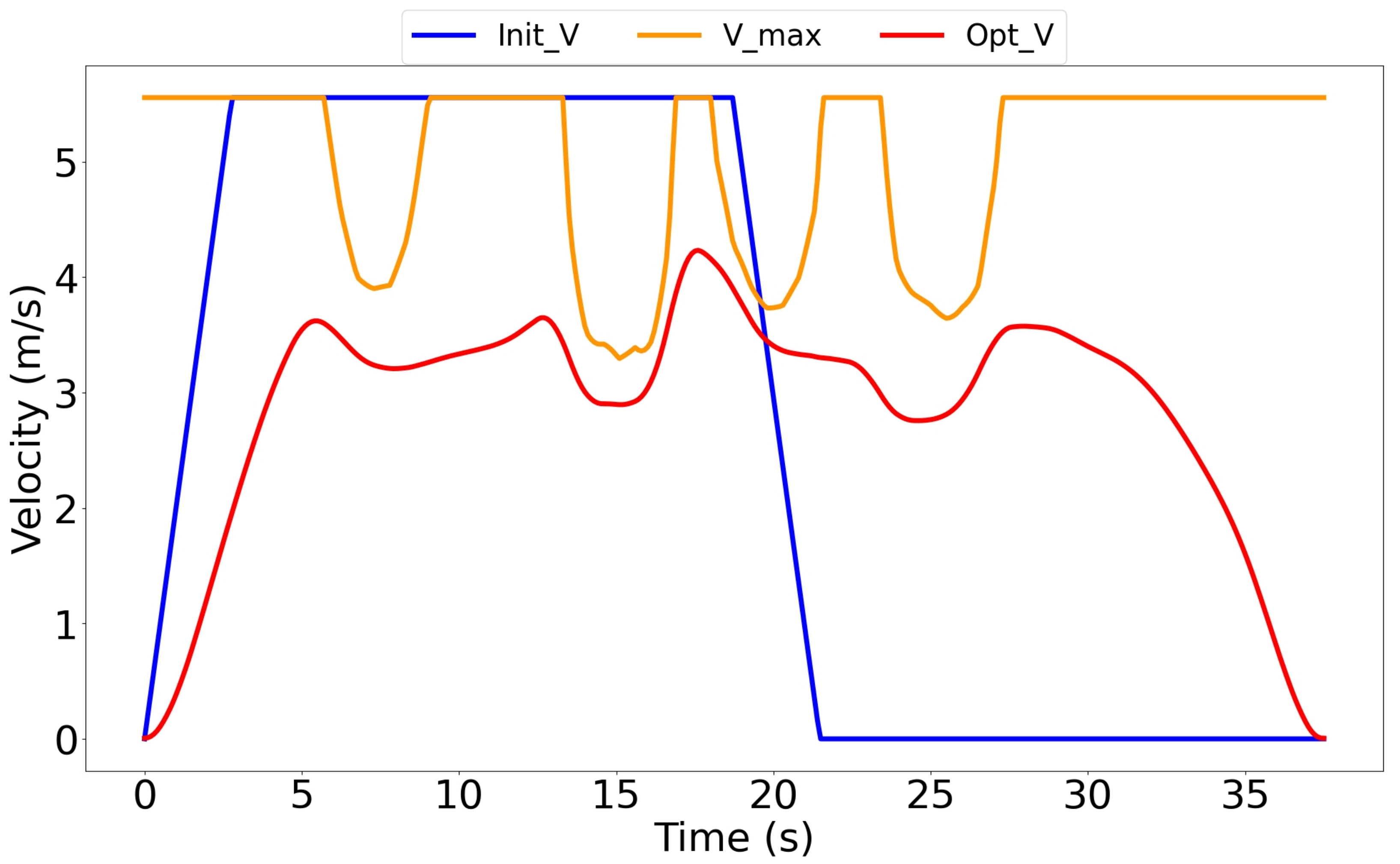}
    \caption{Velocity profile (red) of optimized trajectory with a maximum curvature velocity (yellow) obtained by $\sqrt{a^d_\text{max} / \kappa_t}$.}
    \label{fig:5.curvature_maximum_velocity_constraint}
\end{figure}

\begin{figure}[t]
    \centering
    \begin{subfigure}[b]{0.48\linewidth}        %% or \columnwidth
        \centering
        \includegraphics[width=\linewidth]{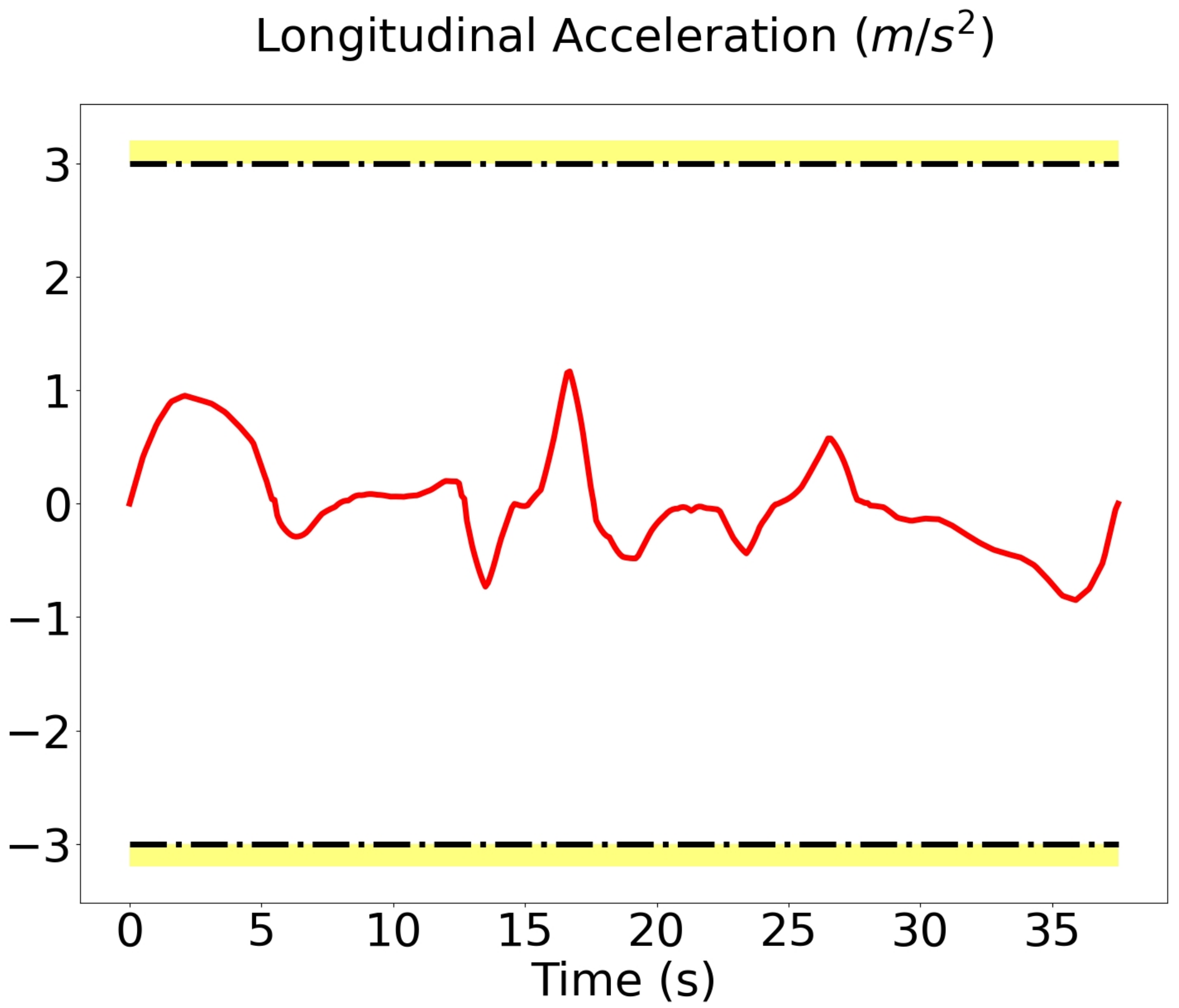}
        %\caption{}
        \label{fig:analysis_feasibility_acc_long}
    \end{subfigure}
    \begin{subfigure}[b]{0.48\linewidth}        %% or \columnwidth
        \centering
        \includegraphics[width=\linewidth]{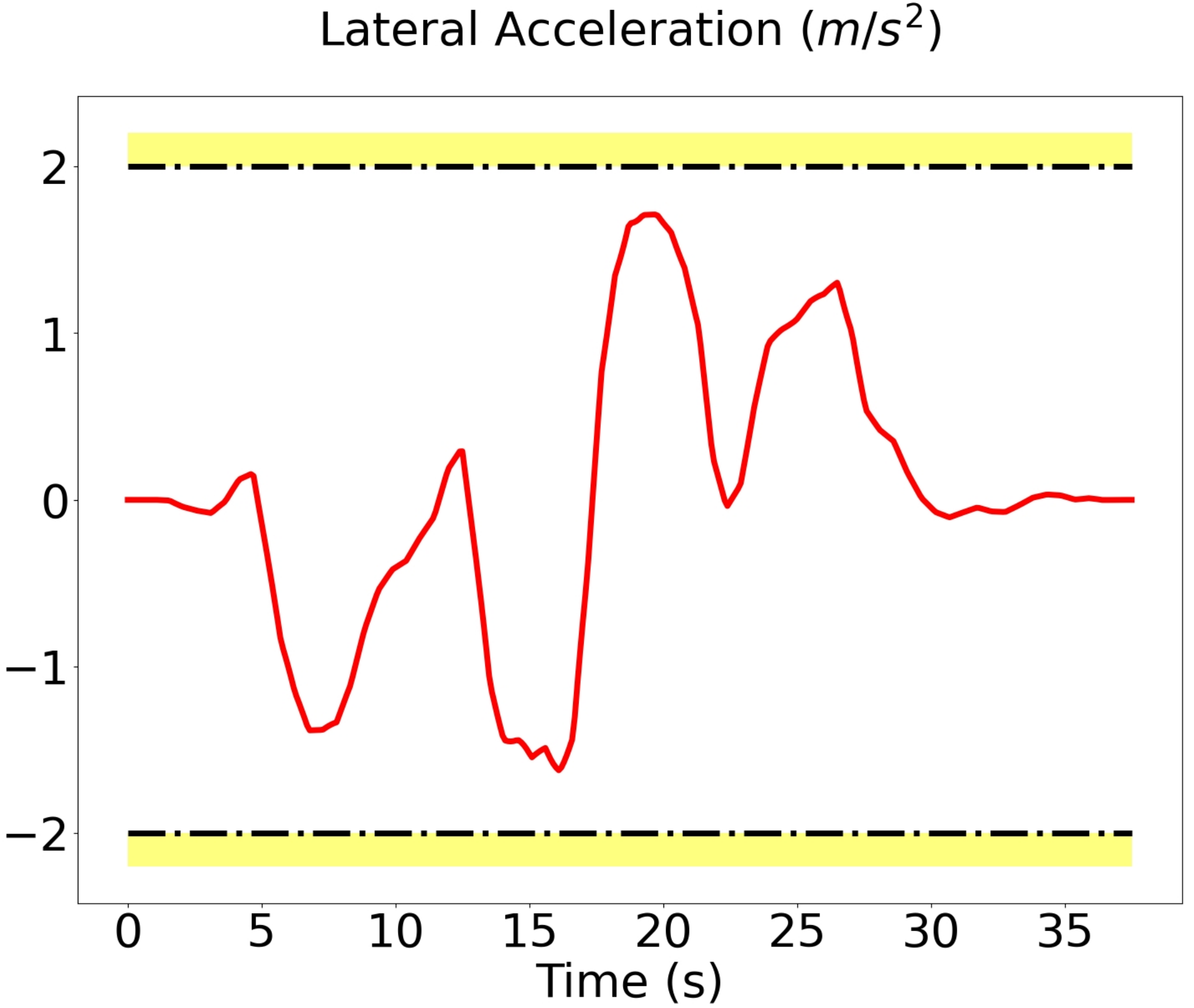}
        %\caption{}
        \label{fig:analysis_feasibility_acc_lat}
    \end{subfigure}
    \caption{Longitudinal and lateral acceleration}
    \label{fig:5.analysis_feasibility_acc}
\end{figure}

\subsection{Trajectory Refinement}
If the trajectory violates the kinodynamic feasibility constraints (velocity, acceleration, and curvature) after rebound optimization, then the trajectory refinement process ensures it meets the constraints through trajectory refinement. The maximum curvature of iteration 7, as shown in parts 1, 2, and 3 in Fig. \ref{fig:analysis_init_opt_compare}, was decreased from $0.2595m^{-1}$ to $0.1844m^{-1}$ by the curvature penalties, and the other feasibility constraints were fulfilled. Fig. \ref{fig:5.curvature_maximum_velocity_constraint} shows that the proposed method met velocity and lateral acceleration constraints simultaneously. Additionally, the longitudinal and lateral acceleration were bounded by the constraints, as shown in Fig. \ref{fig:5.analysis_feasibility_acc}. Despite the challenging environment, which required a long time horizon of $37.5s$ and multiple iterations, the computation times of rebound optimization with \acrshort{ipf} and refinement optimization were $159ms$ and $78ms$, respectively.

\begin{figure}[t]
    \centering
    \includegraphics[width=\linewidth]{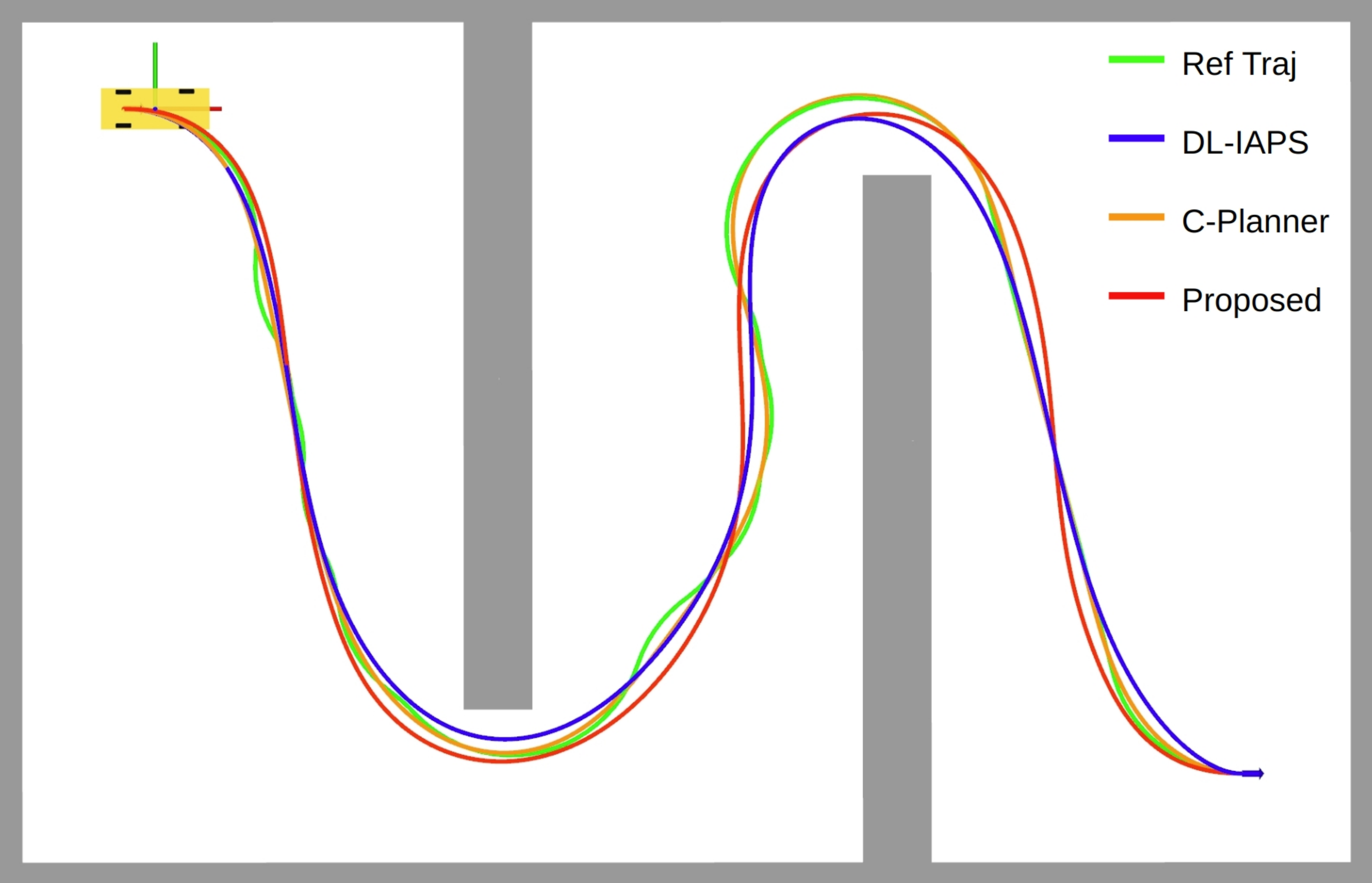}
    \caption{Evaluation of curvature penalty in the double-U-turn environment consisting of both a wide U-turn and a sharp U-turn. The reference trajectory is generated using the hybrid $\text{A}^*$ algorithm.}
    \label{fig:5.SOTA_path_compare}
\end{figure}

\begin{figure}[t]
    \centering
    \includegraphics[width=\linewidth]{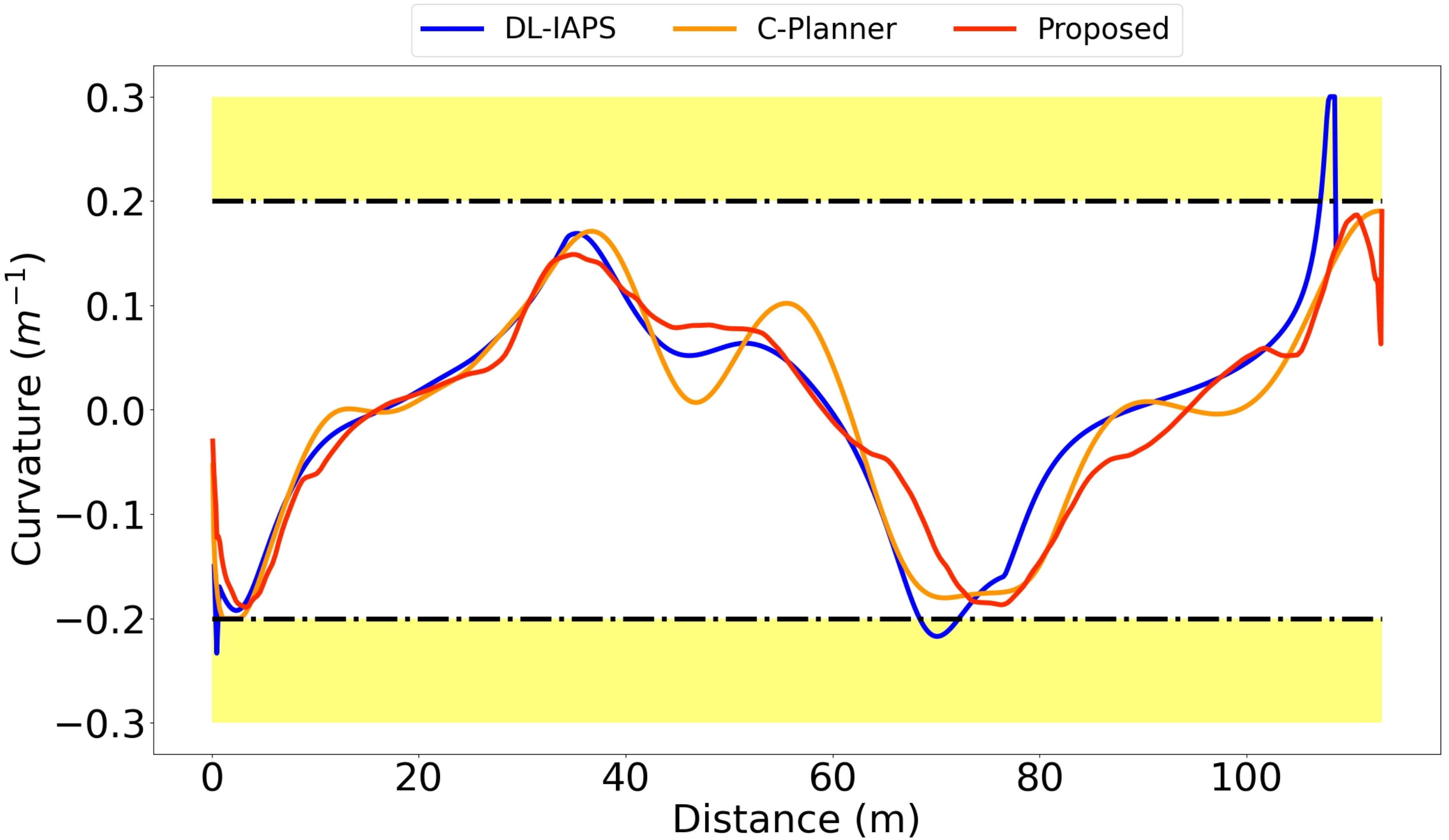}
    \caption{Comparison of curvature of optimized trajectories.}
    \label{fig:5.SOTA_curvature_compare}
\end{figure}

\subsection{Curvature Penalty}
% The curvature constraint of the proposed method was evaluated using the baselines in the double-U-turn environment shown in Fig. \ref{fig:5.SOTA_path_compare}. We evaluated the curvature smoothness in the first (wide) U-turn part and the curvature constraint in the second (sharp) U-turn part.

% The curvatures of the optimized trajectory from the proposed method and the baselines are compared in Fig. \ref{fig:5.SOTA_curvature_compare}. C-Planner reached the maximum curvature in the first U-turn part, whereas DL-IAPS and the proposed method had smaller curvatures under the maximum curvature. Therefore, DL-IAPS and the proposed method generated smoother paths than C-Planner during the wide turn. These smoother paths could increase the maximum curvature velocity, as shown in Fig. \ref{fig:5.curvature_maximum_velocity_constraint}, or lower the tracking error. Fig. \ref{fig:5.SOTA_curvature_compare} also shows that DL-IAPS violated the maximum curvature in the second U-turn part, whereas C-Planner and the proposed method did not exceed the curvature constraint. Therefore, the path from DL-IAPS could increase the tracking error. The curvatures in both U-turn parts show that the path from the proposed method was smoother, more tractable than the paths generated by the baselines. In addition, the computation times of the proposed method, DL-IAPS, and C-Planner were $106$, $1043$, and $499ms$, respectively. The tracking performance of the compared
% methods was then analyzed in detail, as explained in the following Section \ref{subsection: tracking performance}.
To evaluate the curvature penalty, we designed a path, as shown in Fig. \ref{fig:5.SOTA_path_compare}, that includes curves at the start and goal points, as well as a first wide U-turn and a second sharp U-turn. This design allows us to compare the curvature changes in each segment and demonstrate how well the proposed method satisfies the curvature constraint compared to other baselines, and how it further reduces the curvature.

The C-Planner and DL-IAPS both reached the maximum curvature at the start point. In the first wide U-turn part, both methods satisfied the constraint and reduced the curvature. However, in the second sharp U-turn part, while the C-Planner satisfied the curvature constraint, DL-IAPS violated it. At the goal point, the C-Planner again met the curvature constraint, whereas DL-IAPS significantly violated it. In contrast, the proposed method not only satisfies the curvature constraint across all segments but also generally reduces the overall curvature. The momentary increase in curvature at the goal point can be explained by the $\textsc{L}_\text{min}$ for numerical stability, as mentioned in \ref{subsection:Curvature}. This occurs over a very short segment and has negligible impact on actual tracking performance, as shown in the following Section \ref{subsection: tracking performance}. The computation times of DL-IAPS, C-Planner, and the proposed method were $1043ms$, $499ms$, and $106ms$, respectively.
%TO: 93 (55, 38), DL-IAPS: 3987 (3976, 11), CP: 3181 (20435, 2006, 17254)
%TO: 106 (63, 43), DL-IAPS: 1049 (1043, 6), CP: 499 (Opt time: 7443 ms, pnpoly time: 771 ms, coll time: 6944)

\begin{figure} [t]
    \centering
    \begin{subfigure}[b]{\linewidth}        %% or \columnwidth
        \centering
        \includegraphics[width=0.8\linewidth]{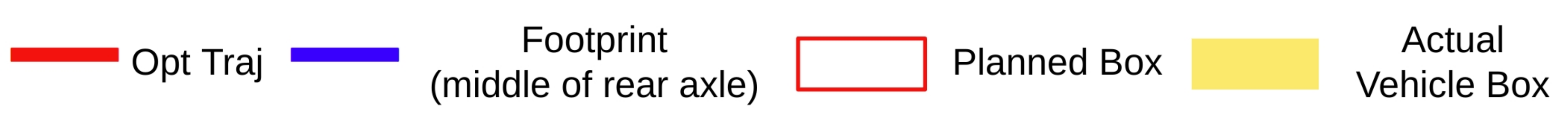}
        %\caption{DL-IAPS}
        \label{fig:ref_opt_legend}
    \end{subfigure}
    \\[1ex]
    \begin{subfigure}[b]{0.3\linewidth}        %% or \columnwidth
        \centering
        \includegraphics[width=\linewidth]{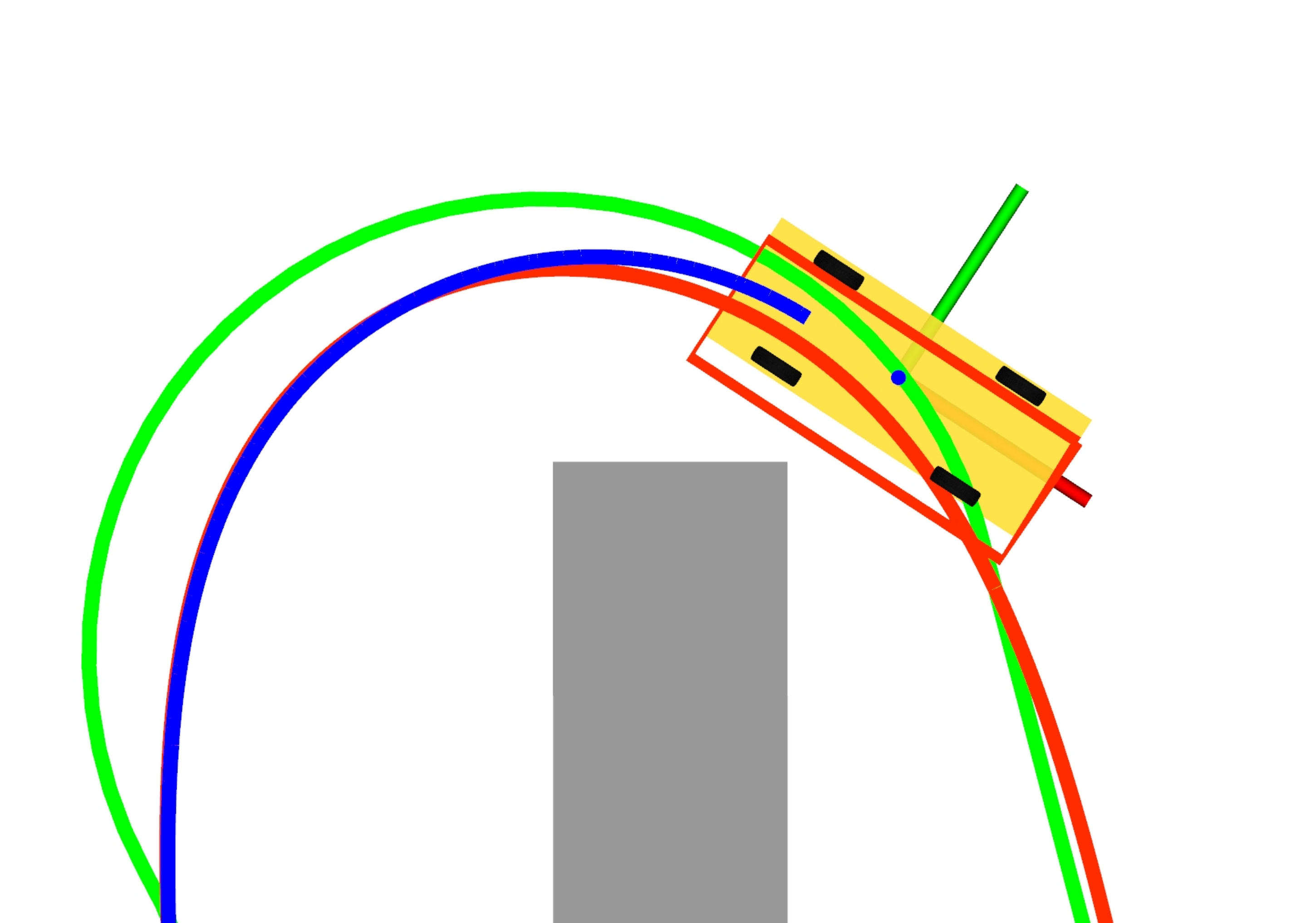}
        \caption{}
        \label{fig:ctrl_error_DL_IAPS}
    \end{subfigure}
    \begin{subfigure}[b]{0.3\linewidth}        %% or \columnwidth
        \centering
        \includegraphics[width=\linewidth]{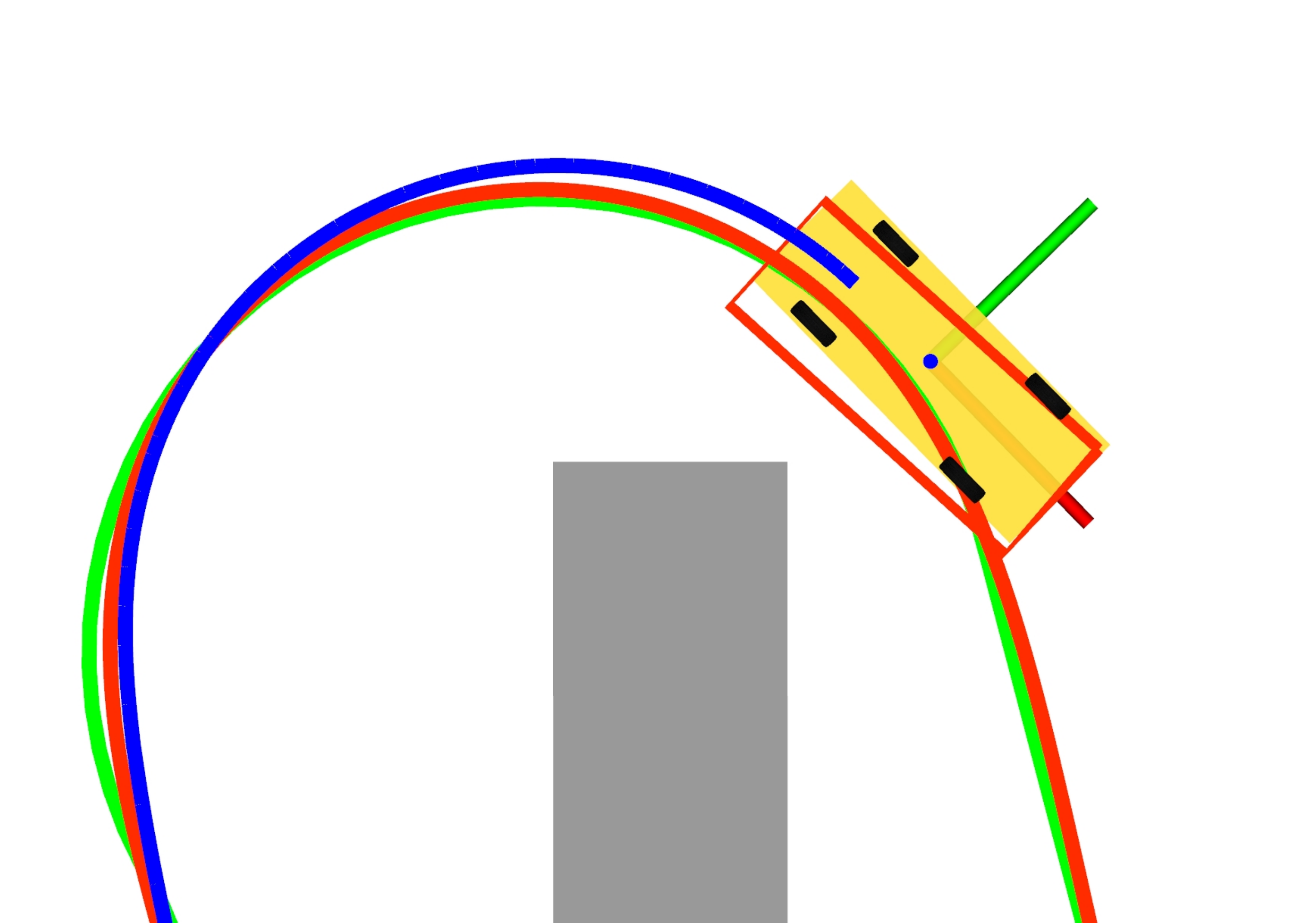}
        \caption{}
        \label{fig:ctrl_error_CP}
    \end{subfigure}
    \begin{subfigure}[b]{0.3\linewidth}        %% or \columnwidth
        \centering
        \includegraphics[width=\linewidth]{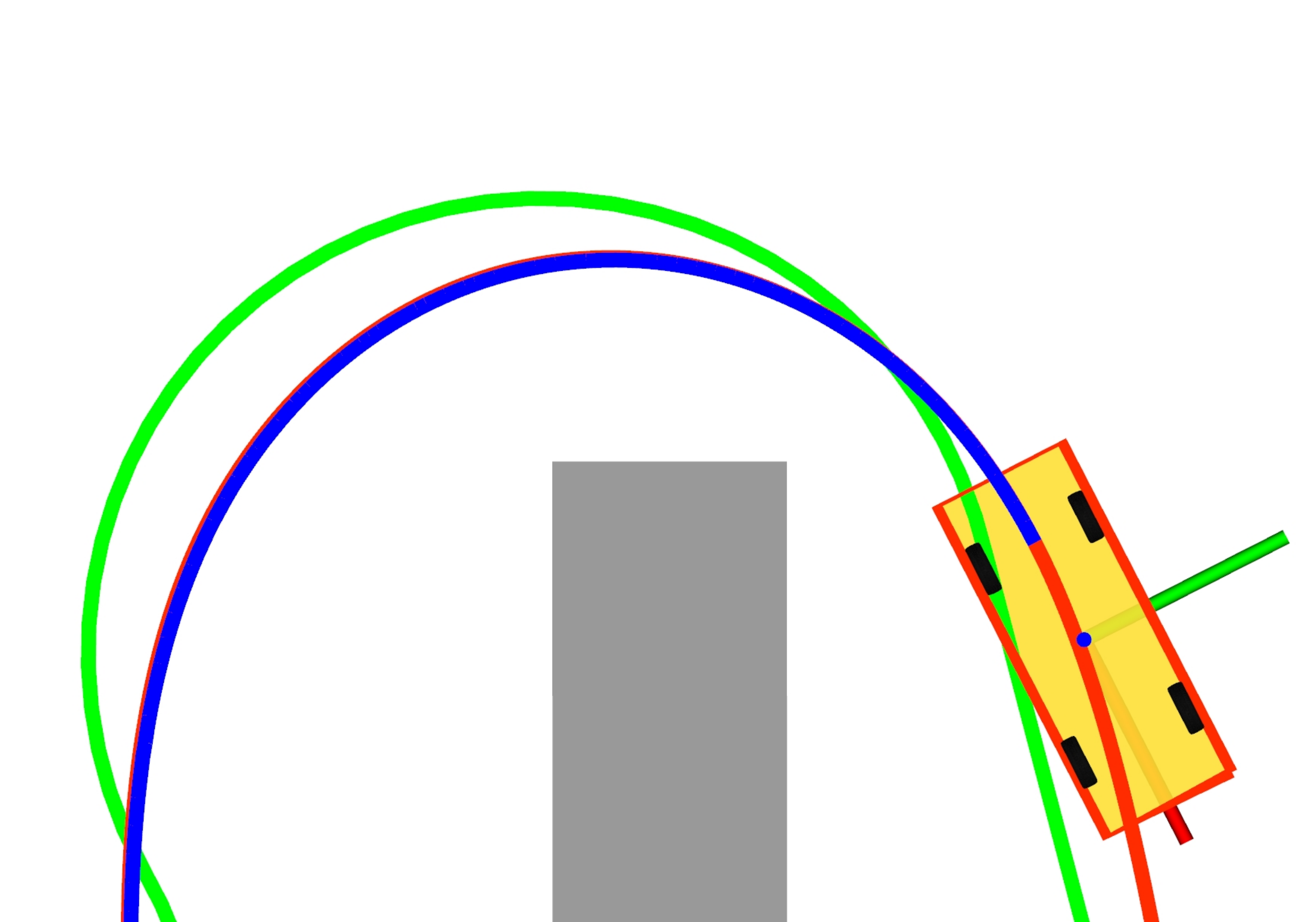}
        \caption{}
        \label{fig:ctrl_error_TO}
    \end{subfigure}
    \caption{Comparison of tracking errors: (a) DL-IAPS, (b) C-Planner, (c) proposed method.}
    \label{fig:tracking_error_comparision}
\end{figure}

% \begin{figure} [h]
%     \centering
%     \begin{subfigure}[b]{0.48\linewidth}        %% or \columnwidth
%         \centering
%         \includegraphics[width=\linewidth]{pic/5.tracking_error_lateral.pdf}
%         \caption{Lateral Error}
%         \label{fig:tracking_error_lateral}
%     \end{subfigure}
%     \begin{subfigure}[b]{0.48\linewidth}        %% or \columnwidth
%         \centering
%         \includegraphics[width=\linewidth]{pic/5.tracking_error_heading.pdf}
%         \caption{Heading Error}
%         \label{fig:tracking_error_heading}
%     \end{subfigure}
%     \caption{}
%     \label{fig:tracking_error_comparision}
% \end{figure}

\begin{figure}[t]
    \centering
    \includegraphics[width=\linewidth]{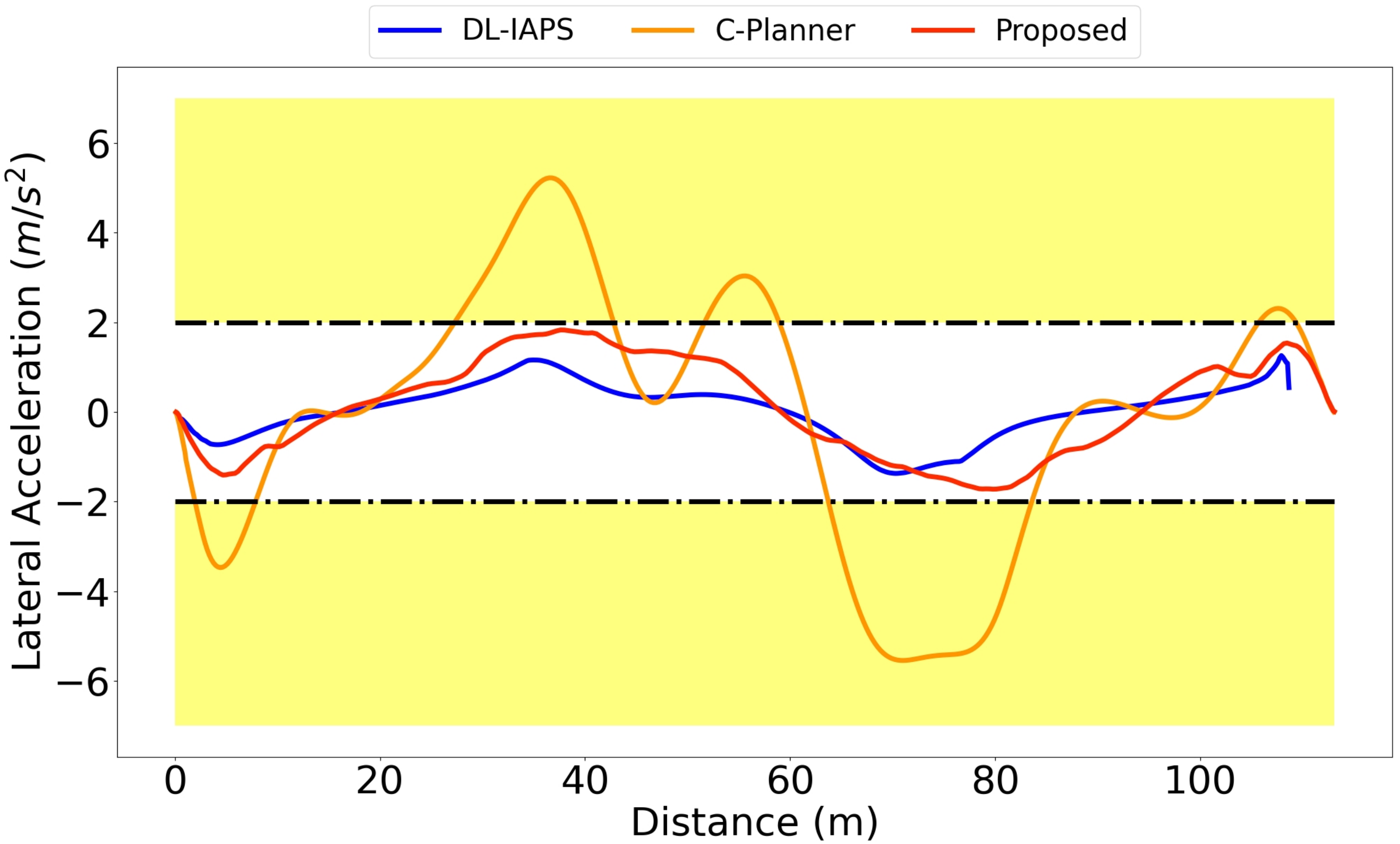}
    \caption{Comparison of lateral acceleration.}
    \label{fig:5.SOTA_lateral_acc_compare}
\end{figure}

\begin{figure} [t]
    \centering
    \begin{subfigure}[b]{0.9\linewidth}        %% or \columnwidth
        \centering
        \includegraphics[width=\linewidth]{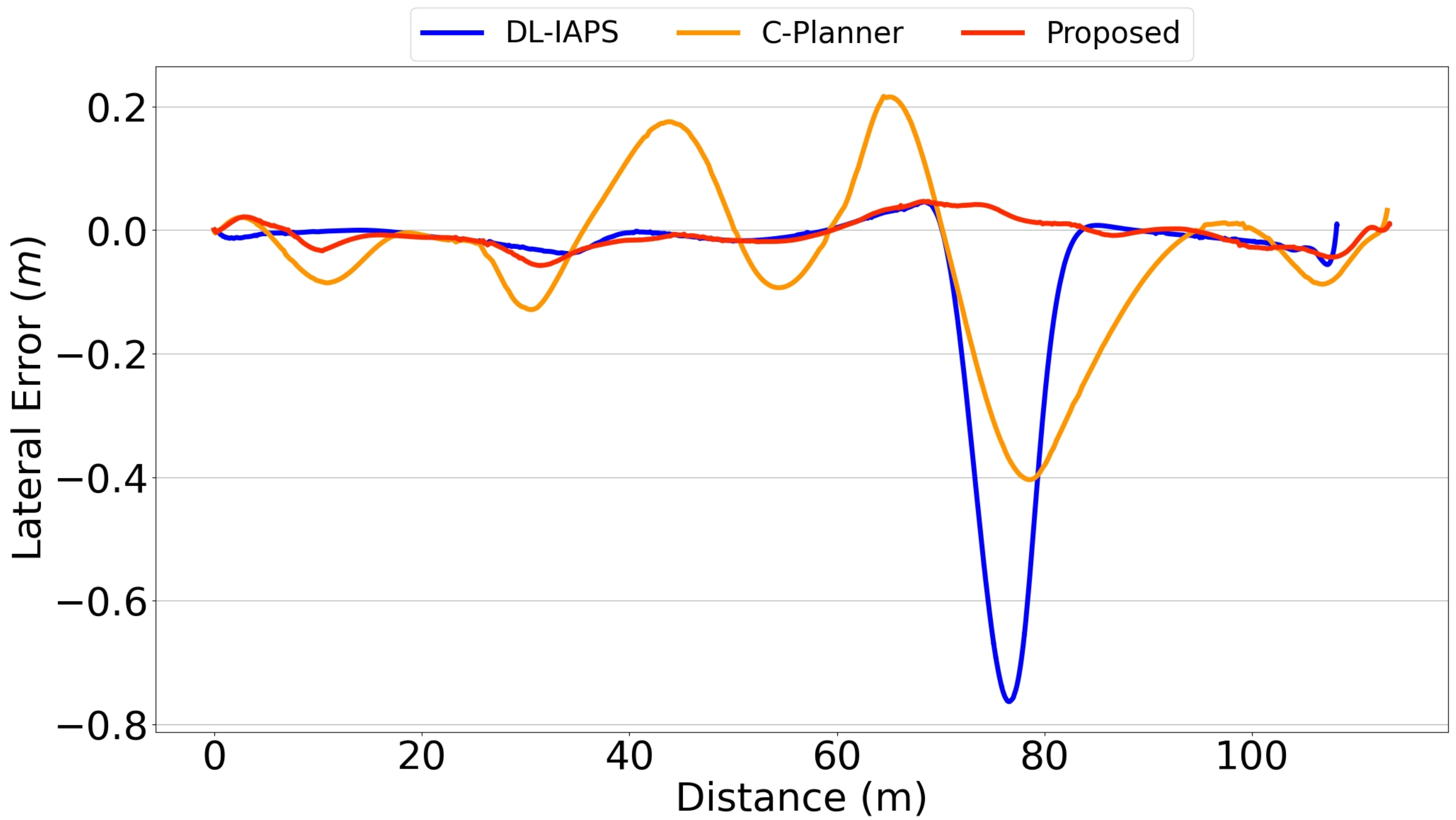}
        %\includegraphics[height=5cm, width=\linewidth]{pic/5.Evaluation/5.tracking_error_lateral.pdf}
        %\caption{Lateral Error}
        \label{fig:tracking_error_lateral}
    \end{subfigure}
    \\[1ex]
    \begin{subfigure}[b]{0.9\linewidth}        %% or \columnwidth
        \centering
        \includegraphics[width=\linewidth]{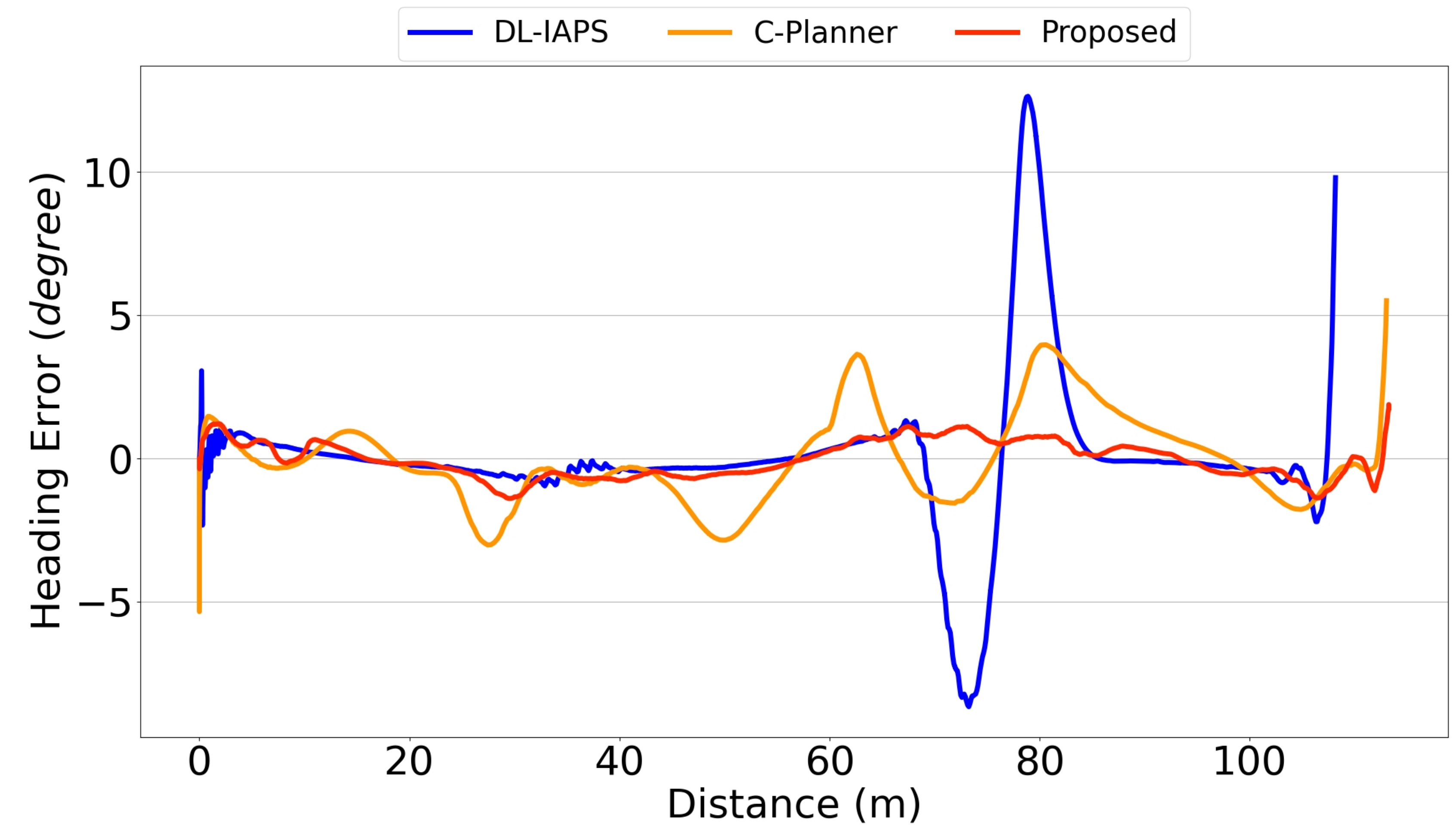}
        %\includegraphics[height=5cm, width=\linewidth]{pic/5.Evaluation/5.tracking_error_heading.pdf}
        %\caption{Heading Error}
        \label{fig:tracking_error_heading}
    \end{subfigure}
    \caption{Comparison of lateral and heading errors.}
    \label{fig:tracking_error_graph_comparision}
\end{figure}

\subsection{Tracking Performance} \label{subsection: tracking performance}
The tracking performance of the compared approaches was evaluated using their generated trajectories in the same environment (Fig. \ref{fig:5.SOTA_path_compare}). We used a \acrfull{mpc}-based controller to track the optimized trajectories. Fig. \ref{fig:tracking_error_comparision} demonstrates the displacement between the actual vehicle box and the planned vehicle box of the optimized trajectory in the second U-turn part. DL-IAPS and C-Planner had large displacements between the boxes and between the planned optimal trajectory and the footprint of the middle of the rear axle. The displacement of DL-IAPS was due to the violation of the maximum curvature, whereas that of C-Planner was due to the violation of the lateral acceleration constraint, as shown in Fig. \ref{fig:5.SOTA_lateral_acc_compare}, since lateral acceleration is not included in its algorithm. Conversely, the trajectory from the proposed method was tracked by the vehicle with little displacement.

For more details about the tracking error, Fig. \ref{fig:tracking_error_graph_comparision} demonstrates the lateral and heading errors of the three methods throughout their trajectories. The baselines had absolute lateral errors of more than $0.3m$ and absolute heading errors of more than $5.0^\circ$, whereas the proposed algorithm had absolute lateral errors of less than $0.1m$ and absolute heading errors of less than $2.0^\circ$ at most. Thus, the proposed method was quantitatively superior in tracking performance.

\begin{figure}[t]
    \centering
    \includegraphics[scale=0.9, width=\linewidth]{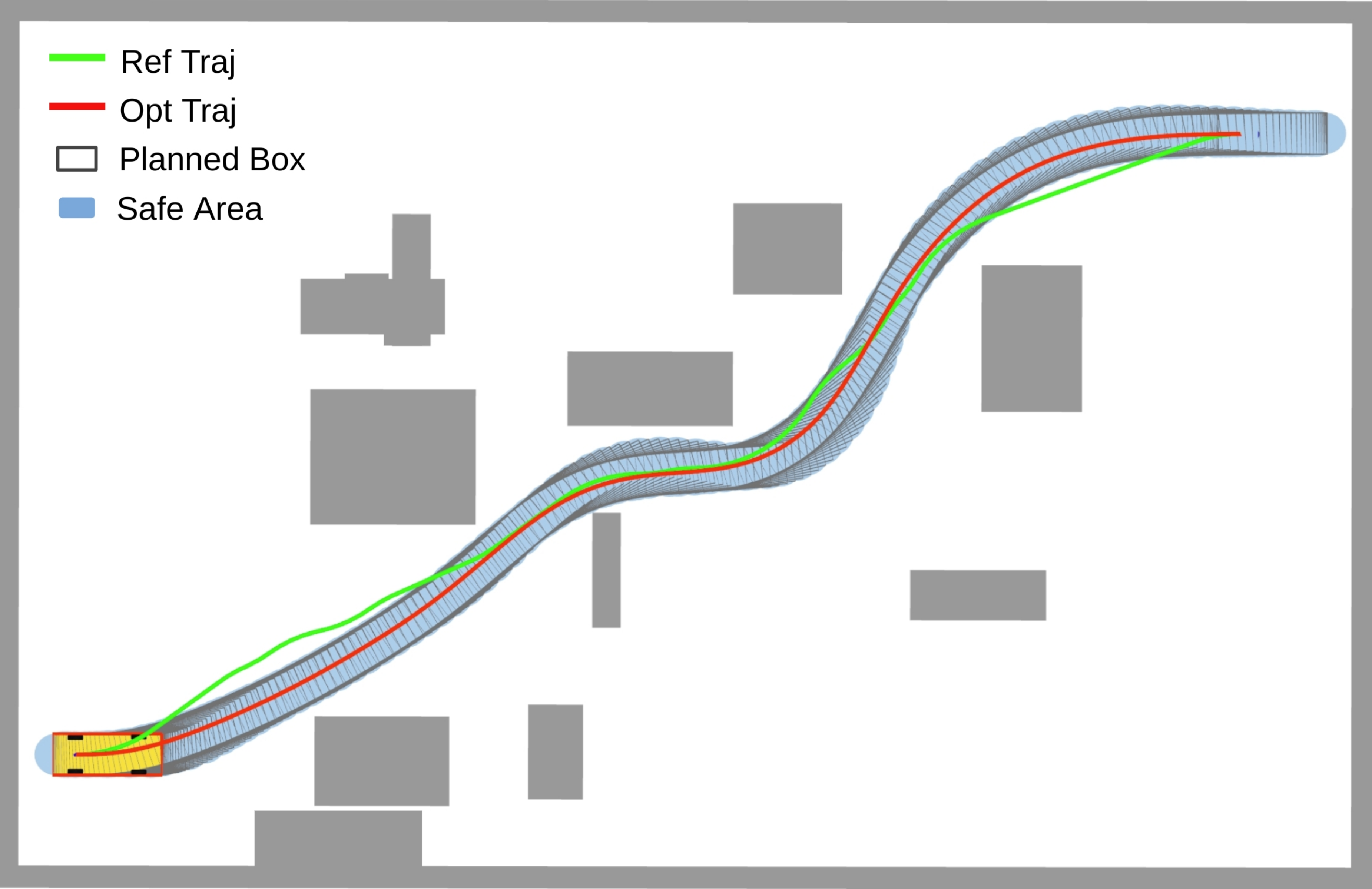}
    \caption{Random-obstacle test.}
    \label{fig:5.random_obstacle_test}
\end{figure}

\begin{table}[t]
\caption{Computation times of proposed method}
\label{table:random_task_computation_time_of_our_method}
%\begin{center}
\begin{adjustbox}{width=\columnwidth,center}
\begin{tabular}{c | c | c  c  c} 
     % \Xhline{2\arrayrulewidth}
     % \multirow{2}{*}{\shortstack{Time\\(ms)}} & \multirow{2}{*}{\shortstack{Reference\\($\text{A}^*$)}} & \multicolumn{2}{c}{Optimization Steps} & \multirow{2}{*}{\shortstack{Optimization\\Total Time}}\\
     % & & \acrshort{ipf} & KW-RF &  \\
     % \Xhline{2\arrayrulewidth}
     % Min & 53.15 & 31.10 & 8.97 & 76.83 \\
     % \hline
     % Avg & 89.31 & 88.74 & 253.88 & 340.63 \\
     % \hline
     % Max & 183.28 & 281.90 & 993.43 & 1215.84  \\
     \Xhline{3\arrayrulewidth}
     \multirow{2}{*}{\shortstack{Time\\(ms)}} & \multirow{2}{*}{\shortstack{Reference\\(Hybrid $\text{A}^*$)}} & \multicolumn{2}{c}{Optimization Steps} & \multirow{2}{*}{\shortstack{Optimization\\Total Time}}\\
     & & Rebound & Refinement &  \\
     \Xhline{2\arrayrulewidth}
     Min & 102.57 & 13.53 & 16.78 & 34.13 \\
     \hline
     Avg & 304.92 & 25.40 & 32.49 & 57.89 \\
     \hline
     Max & 952.42 & 121.92 & 231.67 & 305.63  \\
     \Xhline{2\arrayrulewidth}
\end{tabular}
\end{adjustbox}
%\end{center}
\end{table}

\begin{table}[t]
\begin{adjustbox}{width=\columnwidth,center}
\begin{threeparttable}
\caption{Ablation Study}
\label{table:random_task_ablation_study}
%\begin{center}
\begin{tabular}{c | c | c | c | c | c} 
     \Xhline{2\arrayrulewidth}
     \multirow{2}{*}{\shortstack{Method}} & \multirow{2}{*}{\shortstack{Success\tnote{a} \\Rate (\%)}} & \multicolumn{4}{c}{Violation Count} \\
     & & vel & long acc & lat acc & curvature\\
     \Xhline{2\arrayrulewidth}
     \multirow{2}{*}{\shortstack{Proposed w/o \acrshort{sv}}} & \multirow{2}{*}{88.00} & \multirow{2}{*}{0} & \multirow{2}{*}{0} & \multirow{2}{*}{0} & \multirow{2}{*}{80} \\
     & & & & &\\
     \hline
     \multirow{2}{*}{\shortstack{Proposed w/o \acrshort{ipf}\tnote{b}}} & \multirow{2}{*}{34.60} & \multirow{2}{*}{0} & \multirow{2}{*}{0} & \multirow{2}{*}{0} & \multirow{2}{*}{4} \\
     & & & & &\\
     \hline
     \multirow{2}{*}{\shortstack{Proposed w/o \acrshort{ipf}\tnote{c}}} & \multirow{2}{*}{95.10} & \multirow{2}{*}{0} & \multirow{2}{*}{0} & \multirow{2}{*}{0} & \multirow{2}{*}{12} \\
     & & & & &\\
     \hline
     \multirow{2}{*}{\shortstack{Proposed w/o \acrshort{sv}\&\acrshort{ipf}\tnote{b}}} & \multirow{2}{*}{23.00} & \multirow{2}{*}{0} & \multirow{2}{*}{0} & \multirow{2}{*}{0} & \multirow{2}{*}{\bf{3}} \\
     & & & & &\\
     \hline
     \multirow{2}{*}{\shortstack{Proposed w/o \acrshort{sv}\&\acrshort{ipf}\tnote{c}}} & \multirow{2}{*}{84.30} & \multirow{2}{*}{0} & \multirow{2}{*}{0} & \multirow{2}{*}{1} & \multirow{2}{*}{26} \\
     & & & & &\\
     \hline
     \multirow{2}{*}{\shortstack{Proposed w/o refinement}} & \multirow{2}{*}{13.60} & \multirow{2}{*}{714} & \multirow{2}{*}{145}  & \multirow{2}{*}{737} & \multirow{2}{*}{213} \\
     & & & & &\\
     \hline
     \multirow{2}{*}{\shortstack{Proposed w/o Kinodynamic \\ Feasibility penalty}} & \multirow{2}{*}{90.70} & \multirow{2}{*}{0} & \multirow{2}{*}{0} & \multirow{2}{*}{6} & \multirow{2}{*}{84} \\
     & & & & &\\
     \hline
     \multirow{2}{*}{\shortstack{Proposed}} & \multirow{2}{*}{\bf{97.70}} & \multirow{2}{*}{\bf{0}} & \multirow{2}{*}{\bf{0}} & \multirow{2}{*}{\bf{0}} & \multirow{2}{*}{18} \\
     & & & & &\\
     \Xhline{2\arrayrulewidth}
\end{tabular}
%\end{center}
\begin{tablenotes}\footnotesize
\item [a] The path is free of collision, and the violation of the kinodynamic feasibility constraints (velocity, acceleration, and curvature) is within 5\%. Lateral acceleration is excluded from the success rate of C-Planner.
\item [b] without close obstacle collision penalty.
\item [c] with close obstacle collision penalty.
\end{tablenotes}
\end{threeparttable}
\end{adjustbox}
\end{table}

\begin{table}[t]
\begin{adjustbox}{width=\columnwidth,center}
\begin{threeparttable}
\caption{Comparison with baselines}
\label{table:random_task_comparison_with_sota_algorithms}
%\begin{center}
\begin{tabular}{c | c | c | c | c} 
     \Xhline{2\arrayrulewidth}
     %   \multirow{2}{*}{Method}&  \multirow{2}{*}{DL-IAPS \cite{DL-IAPS_zhou2020autonomous}} &  \multirow{2}{*}{C-Planner \cite{curvy_li2022autonomous}} & \multicolumn{2}{c}{ \multirow{2}{*}{Proposed}} \\ & & & \\
     % %\Xhline{2\arrayrulewidth}
     % Reference & Hybrid $\text{A}^*$ & Hybrid $\text{A}^*$ & Hybrid $\text{A}^*$ & $\text{A}^*$ \\
     Method &  DL-IAPS &  C-Planner  &  $\text{C-Planner}^\text{ACD}$ & Proposed \\
     \Xhline{2\arrayrulewidth}
     \multirow{2}{*}{\shortstack{Success\tnote{a}\\Rate (\%)}} & \multirow{2}{*}{73.30} & \multirow{2}{*}{95.60}  & \multirow{2}{*}{92.80} & \multirow{2}{*}{\bf{97.70}} \\
     % & & & &\\
     % \hline
     % \multirow{2}{*}{\shortstack{Success\tnote{b}\\Rate (\%)}} & \multirow{2}{*}{99.30} & \multirow{2}{*}{95.60}  & \multirow{2}{*}{\bf{100.00}} & \multirow{2}{*}{99.50} \\
     & & & &\\
     \hline
     Violation Count & & & &\\
     \multirow{4}{*}{\shortstack{Long. Vel\\Long. Acc\\Lat. Acc\\Curvature}} & \multirow{4}{*}{\shortstack{0\\0\\1\\260}} & \multirow{4}{*}{\shortstack{0\\0\\(761)\\\bf{0}}} & \multirow{4}{*}{\shortstack{13\\14\\(736)\\72}} & \multirow{4}{*}{\shortstack{\bf{0}\\\bf{0}\\\bf{0}\\18}} \\
     & & & & \\ & & & & \\ & & & & \\
     \hline
     \multirow{2}{*}{\shortstack{Mean. Max. Abs\\Curvature ($m^{-1}$)}} & \multirow{2}{*}{0.1954} & \multirow{2}{*}{0.1784} & \multirow{2}{*}{0.1710} & \multirow{2}{*}{\bf{0.1556}} \\
     & & & &\\
     \hline
     \multirow{2}{*}{\shortstack{Mean. Abs. Acc\\($m/s^{2}$)}} & \multirow{2}{*}{\bf{0.2804}} & \multirow{2}{*}{1.1250} & \multirow{2}{*}{1.1372} & \multirow{2}{*}{0.8629} \\
     & & & &\\
     \hline
     \multirow{2}{*}{\shortstack{Mean. Abs. Jerk\\($m/s^{3}$)}} & \multirow{2}{*}{\bf{0.2019}} & \multirow{2}{*}{1.6270} & \multirow{2}{*}{1.7224} & \multirow{2}{*}{1.0066} \\
     & & & &\\
     \hline
     \multirow{2}{*}{\shortstack{Mean. Planning\\Horizon (s)}} & \multirow{2}{*}{24.96} & \multirow{2}{*}{14.55} & \multirow{2}{*}{\bf{14.39}} & \multirow{2}{*}{16.88} \\
     & & & &\\
     % \hline
     % \multirow{2}{*}{\shortstack{Min. Comp\\Time (ms)}} & \multirow{2}{*}{120.88} & \multirow{2}{*}{1385.50} & \multirow{2}{*}{1385.50} & \multirow{2}{*}{\bf{28.65}} \\
     % & & & &\\
     % \hline
     % \multirow{2}{*}{\shortstack{Avg. Comp\\Time (ms)}} & \multirow{2}{*}{1753.71} & \multirow{2}{*}{5988.27} & \multirow{2}{*}{5988.27} & \multirow{2}{*}{\bf{59.84}} \\
     % & & & &\\
     % \hline
     % \multirow{2}{*}{\shortstack{Max. Comp\\Time(ms)}} & \multirow{2}{*}{6256.03} & \multirow{2}{*}{30483.24} & \multirow{2}{*}{30483.24} & \multirow{2}{*}{\bf{297.66}} \\
     % & & & &\\
     \Xhline{2\arrayrulewidth}
\end{tabular}
%\end{center}
% \begin{tablenotes}\footnotesize
% \item [b] The condition for successful optimization in each solver.
% \end{tablenotes}
\end{threeparttable}
\end{adjustbox}
\end{table}

\begin{figure}[t]
    \centering
    \includegraphics[scale=0.9, width=\linewidth]{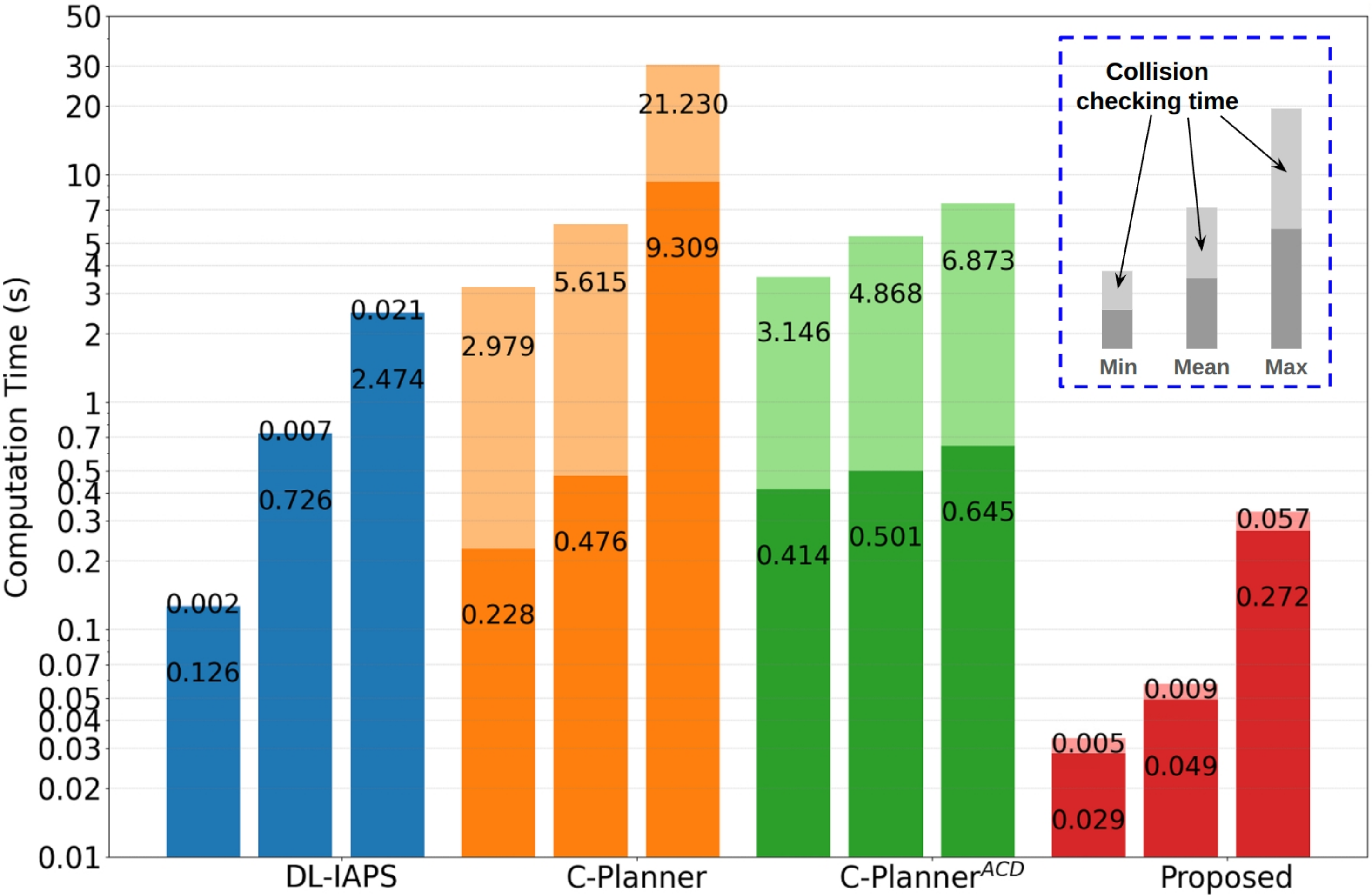}
    \caption{Computation time comparison with baselines.}
    \label{fig:5.bar_plot}
\end{figure}

\subsection{Random Obstacles}
We evaluated the performance of the proposed method compared with the baselines in a random-obstacle environment. A maximum of 10 random-sized rectangular obstacles were randomly positioned between the fixed initial and final poses (position and heading angle), as shown in Fig. \ref{fig:5.random_obstacle_test}. The random test was conducted 1000 times for each algorithm, with all environments consisting only of those where a feasible trajectory exists.

% The computation time of the proposed method is detailed in Table \ref{table:random_task_computation_time_of_our_method}. Aside from the reference computation time (that of hybrid $\text{A}^*$), the computation time of trajectory optimization is broken down into two stages (rebound and refinement) in the table. The average computation time in the refinement stage was $7.36ms$ faster than that in the rebound stage. However, the maximum time of the refinement stage was $79.56ms$ longer than the maximum time of the rebound stage. This was because the fitness penalty in the refinement stage made it difficult to lower curvature to meet the curvature constraint.
The computation time of the proposed method is detailed in Table \ref{table:random_task_computation_time_of_our_method}. Aside from the reference computation time (that of hybrid $\text{A}^*$), the computation time of trajectory optimization is broken down into two stages (rebound and refinement) in the table. The table shows that the refinement stage generally takes longer than the rebound stage, which can be interpreted as the difficulty in simultaneously satisfying both the fitness penalty and the kinodynamic feasibility penalty in the refinement stage. The length of the path is approximately 60 to 80 meters, and the time horizon ranges from about 14 to 30 seconds.

\subsubsection{Ablation Study}

Table \ref{table:random_task_ablation_study} aims to validate the impact of each methodology we propose, namely \acrshort{sv}, \acrshort{ipf} with and without close obstacle collision penalty, refinement, and kinodynamic feasibility penalty, by conducting ablation study through random testing. First, in the methodology excluding \acrshort{sv}, only circle collision checks were performed, resulting in approximately $10\%$ lower success rate due to frequent curvature constraint violations. This supports our claim that both \acrshort{ipf} and \acrshort{sv} need to be used together to achieve the intended trajectory. Otherwise, it can be interpreted as generating a path with difficult-to-relieve curvature.

Next, we conducted ablation study by excluding \acrshort{ipf}, using two methods: one that simultaneously removes both \acrshort{ipf} and the close obstacle collision penalty (\acrshort{ipf}$^\text{b}$), and another that removes \acrshort{ipf} but retains the close obstacle collision penalty (\acrshort{ipf}$^\text{c}$). The results demonstrated that the close obstacle collision penalty significantly impacts the success rate, and the lower feasibility violation count was due to the inability to find a collision-free trajectory during the rebound optimization stage. Even with the close obstacle collision penalty, the absence of \acrshort{ipf} still affected the success rate. Additionally, excluding both \acrshort{sv} and \acrshort{ipf} resulted in a lower success rate compared to excluding only \acrshort{ipf}, indicating that using both \acrshort{sv} and \acrshort{ipf} together improves success rates.

Excluding refinement resulted in a significantly lower success rate, suggesting that the initial time allocation was inaccurate and the refinement process effectively performed time relaxation. The high violation count indicated that a collision-free trajectory was found, but the refinement process failed due to kinodynamic feasibility constraint violations. Finally, when applying time relaxation from refinement but excluding the kinodynamic feasibility penalty, it was shown that longitudinal velocity and acceleration constraints could be satisfied with time relaxation alone, but lateral acceleration and curvature constraints required the kinodynamic feasibility penalty. These ablation study results confirm the effectiveness of all proposed methods (\acrshort{sv}, \acrshort{ipf} with and without close obstacle collision penalty, refinement, and kinodynamic feasibility penalty) and demonstrate that our randomly set test environment was appropriately challenging and capable of yielding meaningful results.

\subsubsection{Comparison with Baselines}
Table \ref{table:random_task_comparison_with_sota_algorithms} presents an experiment to demonstrate how the proposed method outperforms other baselines in various metrics, using the same random test environment as in the previous ablation study. In this experiment, we consistently used DL-IAPS and C-Planner, and additionally employed $\text{C-Planner}^\text{ACD}$, which utilizes \acrlong{sota} \acrshort{ocp} solver \acados, for diverse comparisons.

% First, when comparing success rates, proposed method가 다른 baseline들 보다 더 좋은 성능을 보여주고 있습니다. 다만 우리는 1000번의 test중 곡률 위반에 의한 실패 (18건)이 아닌 rebound optimization cases (5건)을 분석한 결과 tight L shaped turns과 같은 어려운 case에서의 실패 case들도 확인 할 수 있었는데 이는 우리가 제한한 방법론이 이러한 어려운 상황에서의 성공을 보장하지 못한다는 limitation을 보여줍니다. 이러한 이유중에 하나는 우리가 기반한 EGO-Planner의 경우를 보더라고 완전한 completeness를 보장하지 못한다는 점입니다. 다만 다른 baseline들에 비하여 높은 성능을 보여주고 있다는 점에서 우리의 방법론의 경쟁력을 보여줄 수 있습니다.
% However, upon analyzing 1000 test cases, we observed that failures due to curvature violations occurred in 18 instances, while rebound optimization cases accounted for 5 instances. These failures often occurred in challenging scenarios, such as tight L-shaped turns, highlighting a limitation of our method in guaranteeing success under such difficult conditions. One reason for this is that the \acrshort{ego-planner} \cite{zhou2020ego-planner}, which our method is based on, does not guarantee complete completeness. Nevertheless, the superior performance of our method compared to other baselines indicates its competitive advantage.

First, when comparing success rates, the proposed method demonstrates superior performance compared to other baselines. The proposed method also showed superior performance in terms of kinodynamic feasibility constraint violations. The lateral acceleration of C-Planner and $\text{C-Planner}^\text{ACD}$ was excluded from the evaluation metrics as it was not originally formulated.

In terms of mean max absolute curvature, which indicates the overall smoothness of the trajectory, the proposed method significantly mitigated the trajectory curvature. Although DL-IAPS exhibited the lowest mean absolute acceleration and jerk, this is due to the decoupled method's drawback, as mentioned in Section \ref{section:Introduction}, of accurately calculating speed limits caused by lateral acceleration. This results in the vehicle traveling excessively slowly, leading to the longest mean planning horizon (the planned travel time to reach the goal). C-Planner and $\text{C-Planner}^\text{ACD}$ had shorter mean planning horizons due to the lack of lateral acceleration constraints, but their mean absolute acceleration and jerk were higher than those of the proposed method. The proposed method demonstrated better metrics in curvature, acceleration, and jerk compared to other baselines while maintaining a short mean planning horizon, indicating superior overall trajectory quality.

However, upon analyzing 1000 test cases, we observed 18 cases of failures due to curvature violations and 5 cases of rebound optimization. These failures often occurred in challenging scenarios, such as tight L-shaped turns, highlighting a limitation of our method in guaranteeing success under such difficult conditions. One reason for this is that the \acrshort{ego-planner} \cite{zhou2020ego-planner}, which our method is based on, does not guarantee completeness. Nevertheless, the superior performance of our method compared to other baselines indicates its competitive advantage.

Fig. \ref{fig:5.bar_plot} shows a bar plot comparing the computation times of the proposed method with the baselines. We distinguished collision checking time separately to exclude its influence, as our method using \textit{KD-tree} \cite{kd_tree_rusu20113d} may differ from the collision check methods of the baselines. While DL-IAPS has the shortest minimum computation time among the baselines, its average computation time is longer due to re-optimization steps when collisions are detected, which involve reducing the radius from the anchoring point.

C-Planner and $\text{C-Planner}^\text{ACD}$ exhibit similar mean computation times. C-Planner formulates the \acrshort{ocp} problem as a non-linear problem and solves it iteratively, leading to a mean computation time comparable to $\text{C-Planner}^\text{ACD}$, which uses a state-of-the-art solver. However, C-Planner's maximum computation time is significantly longer than that of $\text{C-Planner}^\text{ACD}$. The latter demonstrates better performance due to solving the problem with less variance. The proposed method, however, achieves an average computation time that is approximately ten times faster than $\text{C-Planner}^\text{ACD}$ and more than twice as fast in terms of maximum computation time. Therefore, the proposed method is superior to the baselines in terms of computation time.

\section{EXPERIMENTAL RESULTS}
\label{section:experimental_result}
\begin{figure}[t]
    \centering
    \includegraphics[scale=1, width=0.9\linewidth]{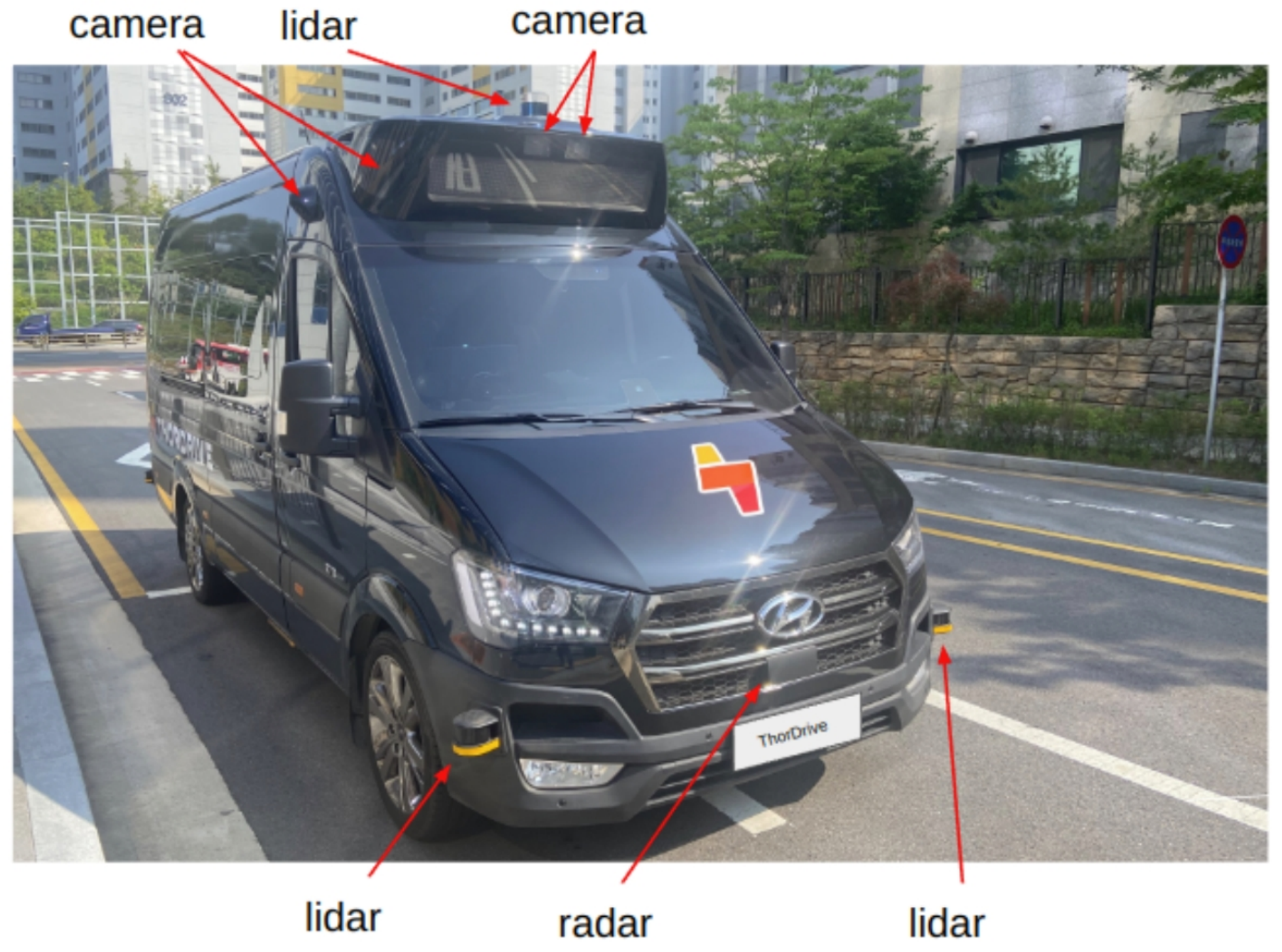}
    \caption{Test vehicle for the real-world experiment. The test vehicle is equipped with several lidars, radars, and cameras. The wheelbase is $3.72 m$, and the minimum turning radius is $6 m$.}
    \label{fig:6.test_vehicle}
\end{figure}

\begin{figure}[t]
    \centering
    \begin{subfigure}[b]{\linewidth}        %% or \columnwidth
        \centering
        \includegraphics[width=0.8\linewidth]{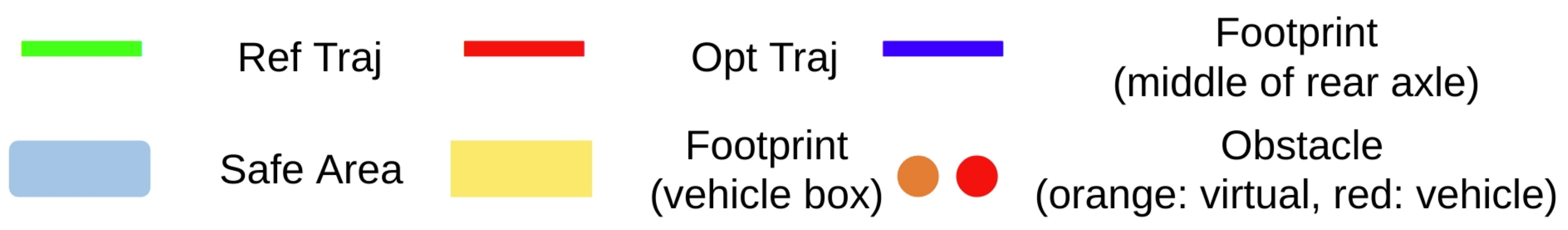}
        %\caption{}
        \label{fig:ref_rviz_legend}
    \end{subfigure}
    \\[1ex]
    \begin{subfigure}[b]{0.48\linewidth}        %% or \columnwidth
        \centering
        \includegraphics[width=\linewidth]{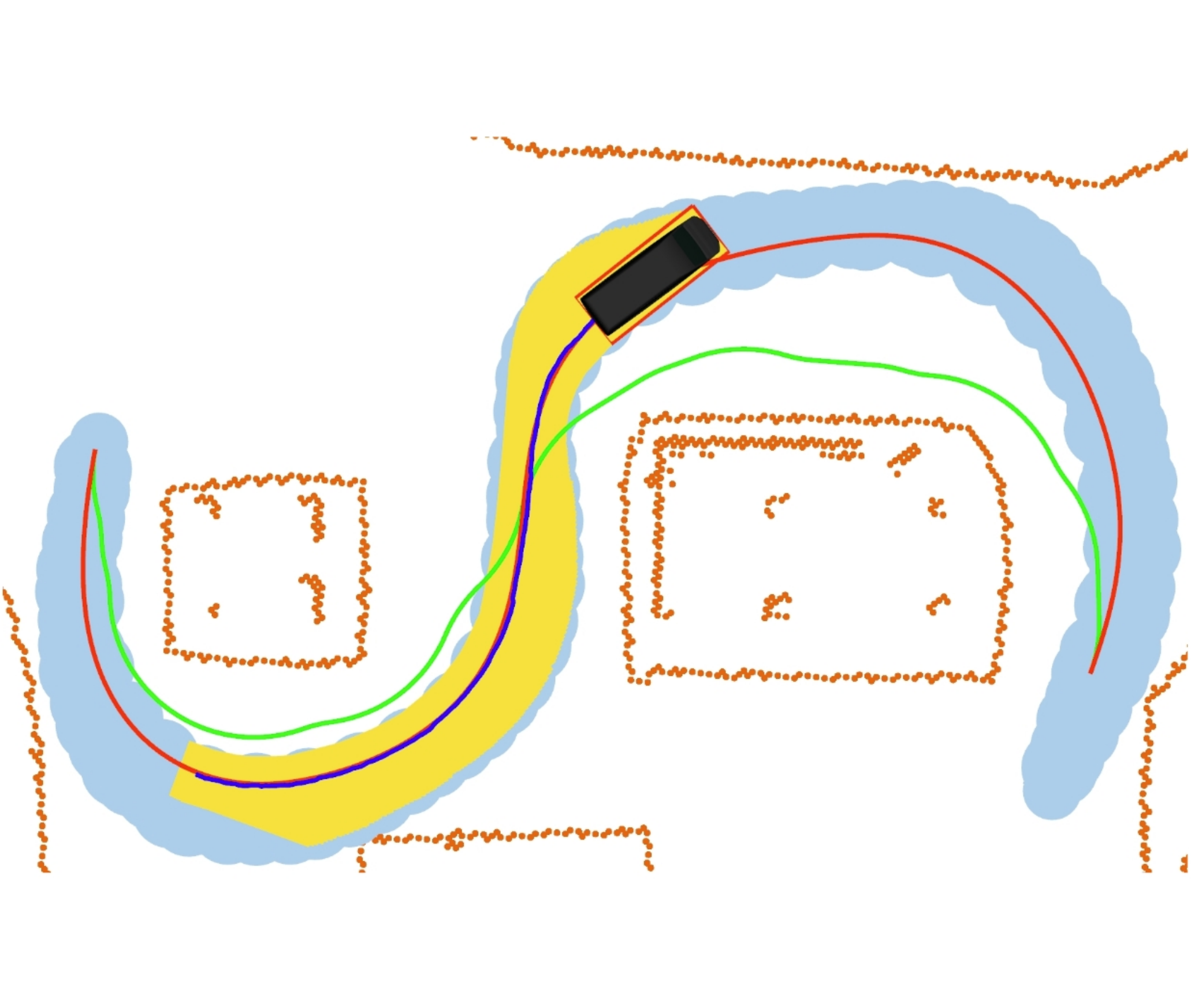}
        \caption{}
        \label{fig:vehicle_test_1}
    \end{subfigure}
    \begin{subfigure}[b]{0.48\linewidth}        %% or \columnwidth
        \centering
        \includegraphics[width=\linewidth]{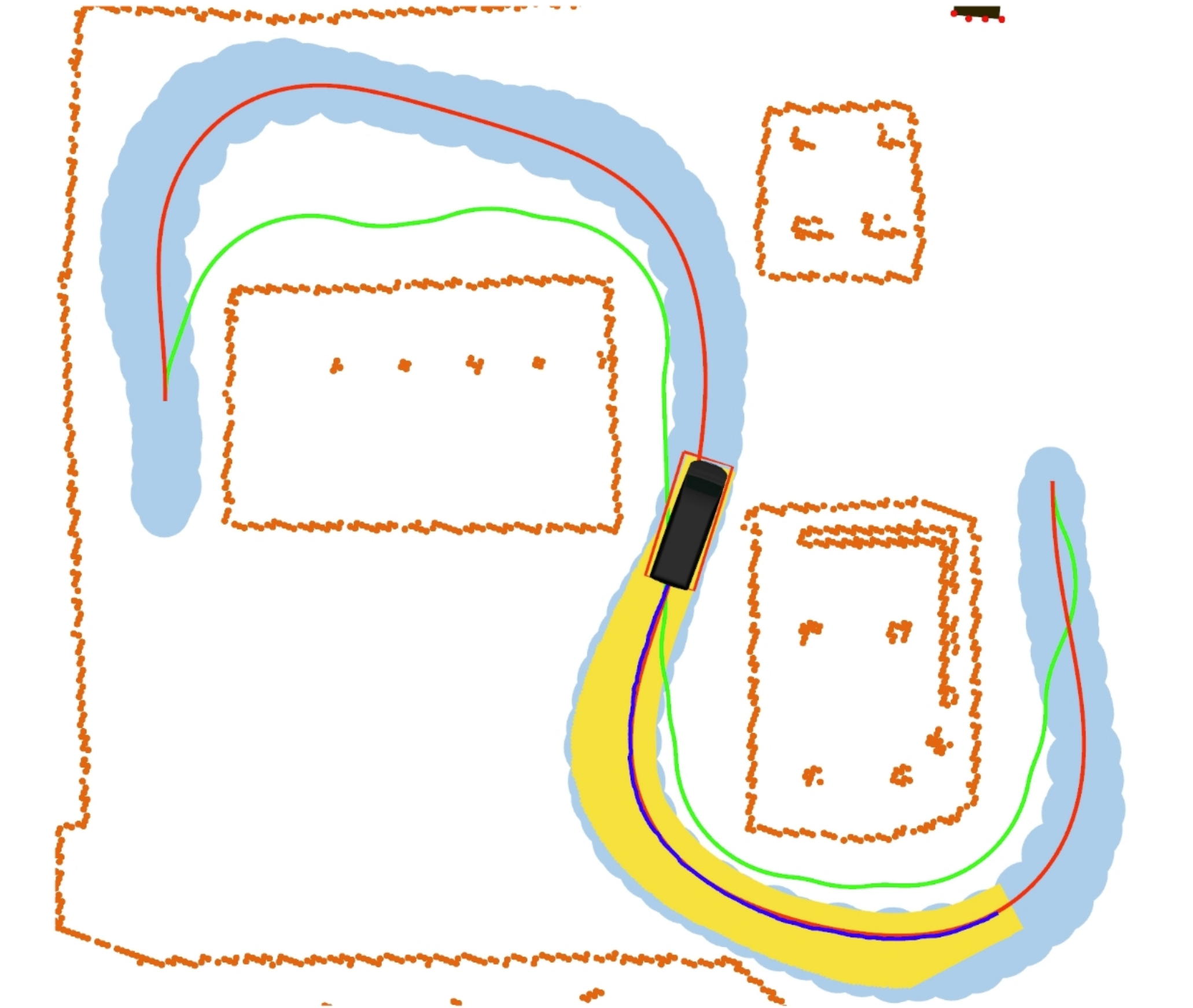}
        \caption{}
        \label{fig:vehicle_test_2}
    \end{subfigure}
    %\\[1ex]
    \begin{subfigure}[b]{0.48\linewidth}        %% or \columnwidth
        \centering
        \includegraphics[width=\linewidth]{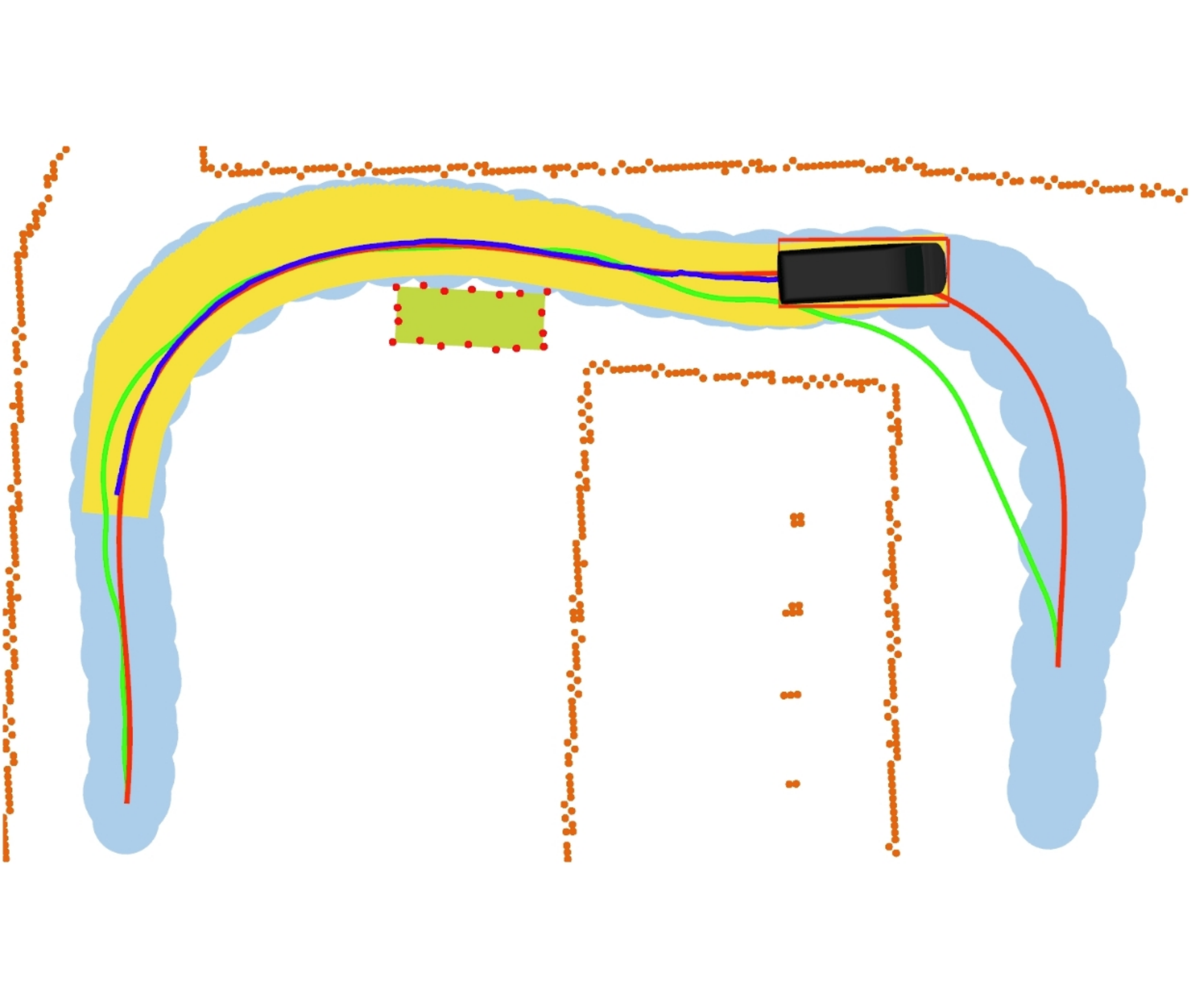}
        \caption{}
        \label{fig:vehicle_test_3}
    \end{subfigure}
    \begin{subfigure}[b]{0.48\linewidth}        %% or \columnwidth
        \centering
        \includegraphics[width=\linewidth]{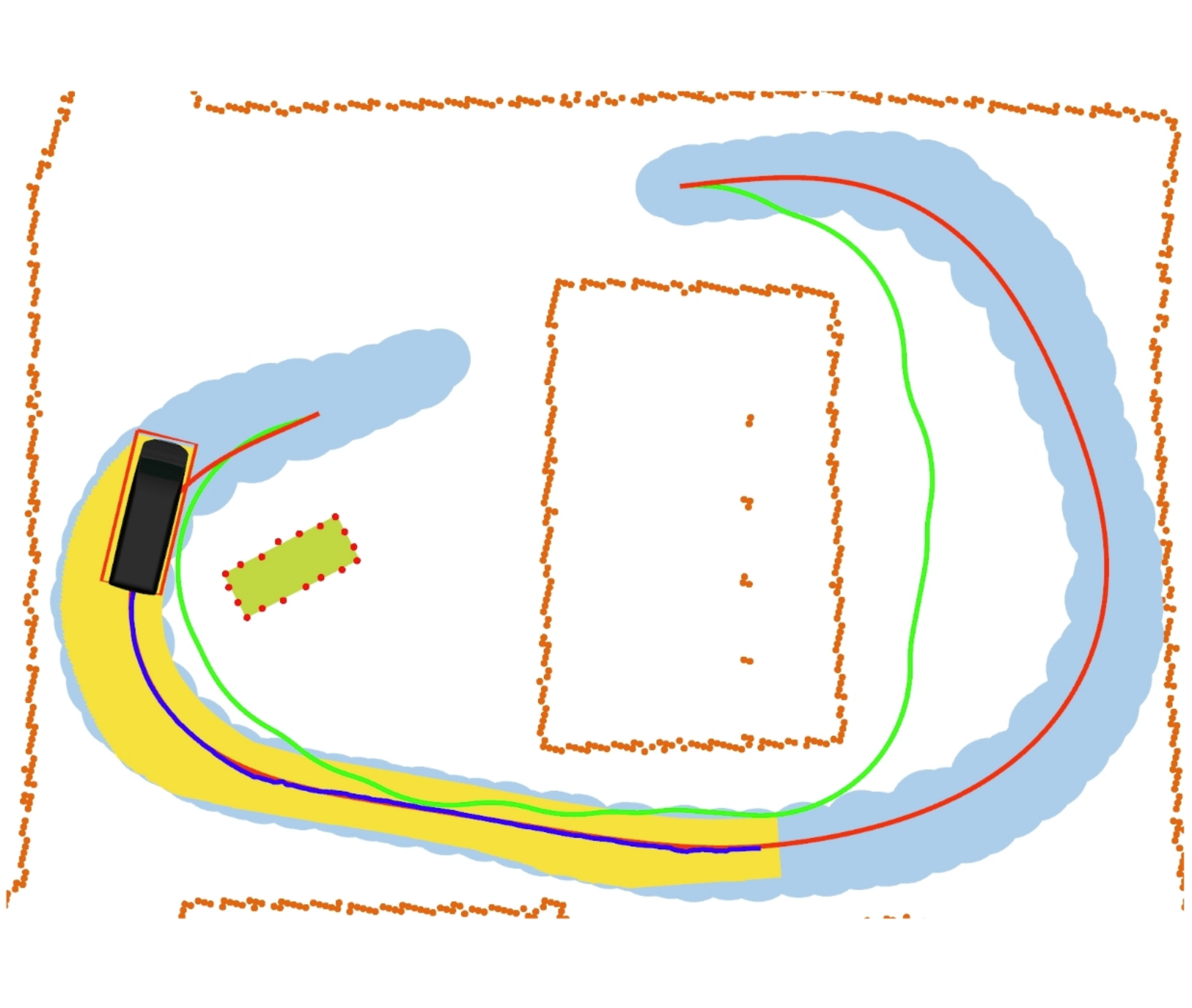}
        \caption{}
        \label{fig:vehicle_test_4}
    \end{subfigure}
    \caption{Four scenarios for the real-world experiments: (a) s-curve 1, (b) s-curve 2, (c) narrow corridor, (d) sharp turn.}
    \label{fig:vehicle_test_rviz}
\end{figure}

\begin{figure} [t]
    \centering
    \begin{subfigure}[b]{0.9\linewidth}        %% or \columnwidth
        \centering
        \includegraphics[height=3.5cm, width=\linewidth]{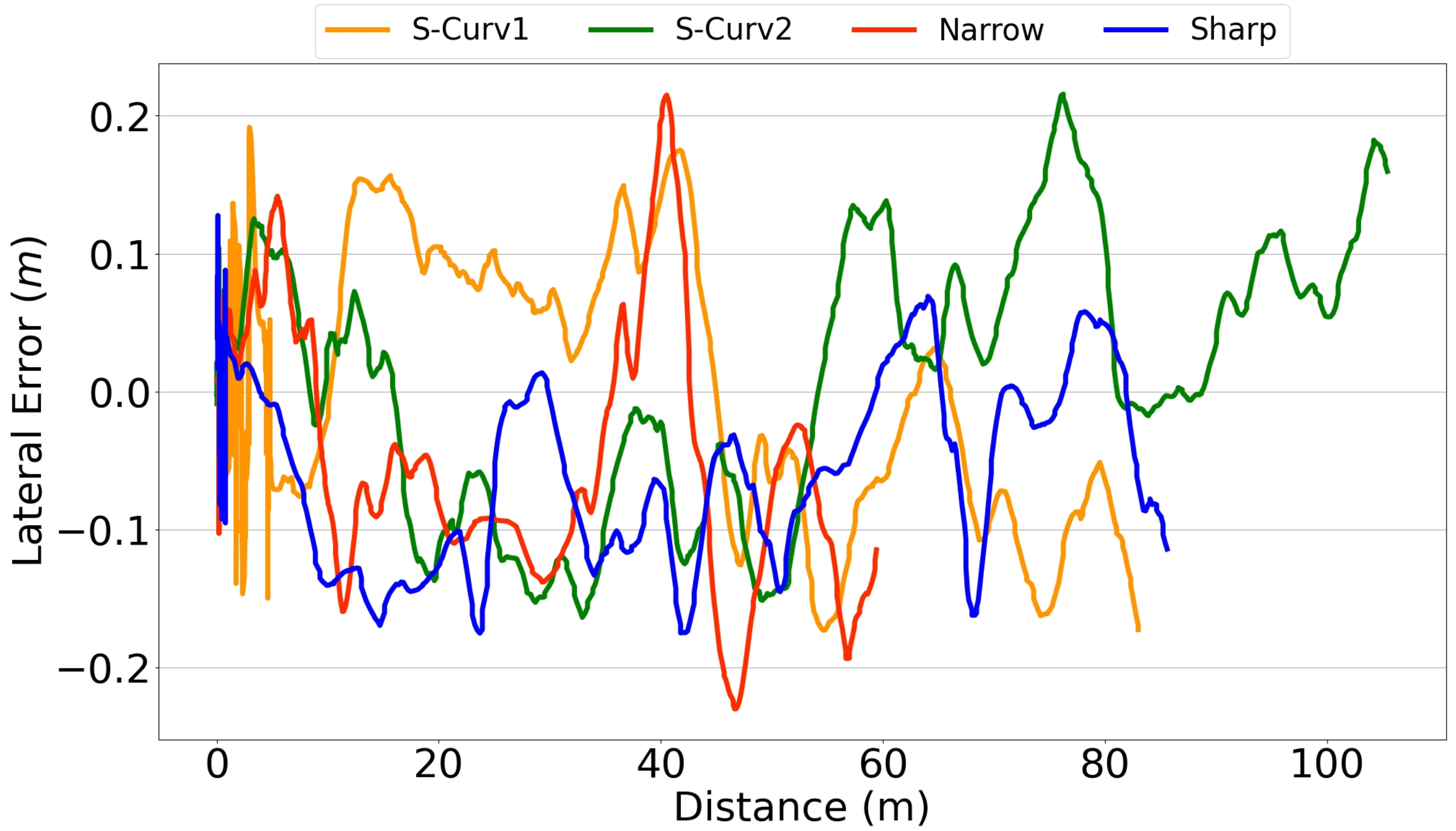}
        %\caption{Lateral Error}
        \label{fig:tracking_error_lateral_real_world}
    \end{subfigure}
    \\[1ex]
    \begin{subfigure}[b]{0.9\linewidth}        %% or \columnwidth
        \centering
        \includegraphics[height=3.5cm, width=\linewidth]{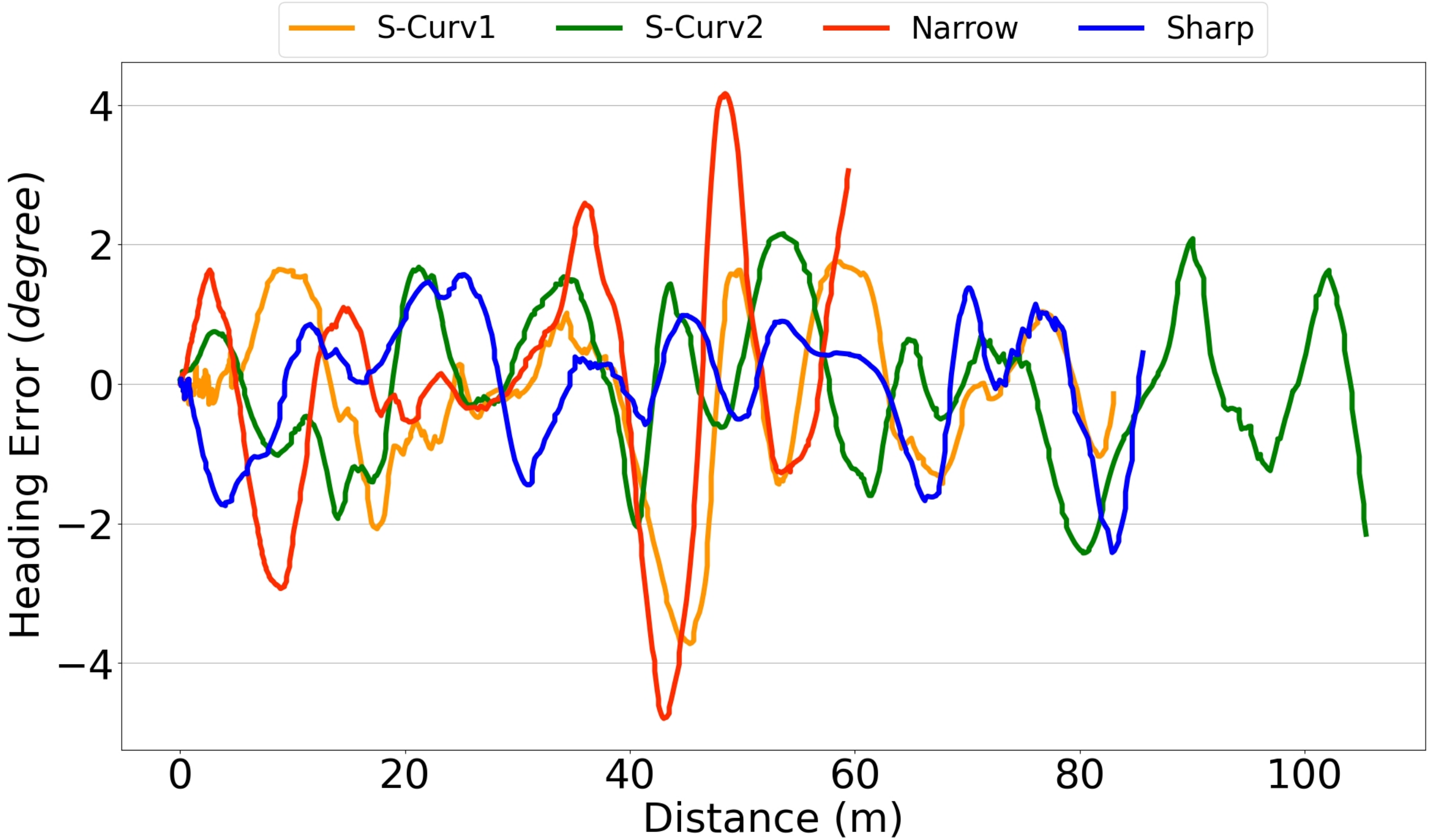}
        %\caption{Heading Error}
        \label{fig:tracking_error_heading_real_world}
    \end{subfigure}
    \caption{Lateral and heading errors in real world.}
    \label{fig:tracking_error_graph_real_world}
\end{figure}

\begin{table}[t]
\begin{adjustbox}{width=\columnwidth,center}
\begin{threeparttable}
\caption{Actual vehicle status in real-world}
\label{table:vehicle_test_result}
%\begin{center}
\begin{tabular}{c | c | c | c | c} 
     \Xhline{2\arrayrulewidth}
     %   \multirow{2}{*}{Method}&  \multirow{2}{*}{DL-IAPS \cite{DL-IAPS_zhou2020autonomous}} &  \multirow{2}{*}{C-Planner \cite{curvy_li2022autonomous}} & \multicolumn{2}{c}{ \multirow{2}{*}{Proposed}} \\ & & & \\
     % %\Xhline{2\arrayrulewidth}
     % Reference & Hybrid $\text{A}^*$ & Hybrid $\text{A}^*$ & Hybrid $\text{A}^*$ & $\text{A}^*$ \\
     \multirow{2}{*}{Scenarios} &  \multirow{2}{*}{S-curve 1} &  \multirow{2}{*}{S-curve 2} & \multirow{2}{*}{\shortstack{Narrow\\corridor}} & \multirow{2}{*}{\shortstack{Sharp\\turn}} \\ & & & & \\
     \Xhline{2\arrayrulewidth}
     \hline
     \multirow{2}{*}{\shortstack{Max. Abs\\Long. Vel ($m/s)$}} & \multirow{2}{*}{1.9718} & \multirow{2}{*}{2.0524} & \multirow{2}{*}{2.4233} & \multirow{2}{*}{2.5633} \\
     & & & & \\
     \hline
     \multirow{2}{*}{\shortstack{Max. Abs\\Long. Acc ($m/s^{2}$)}} & \multirow{2}{*}{1.4554} & \multirow{2}{*}{0.8802} & \multirow{2}{*}{1.1868} & \multirow{2}{*}{1.2832} \\
     & & & & \\
     \hline
     \multirow{2}{*}{\shortstack{Max. Abs\\Lat. Acc ($m/s^{2}$)}} & \multirow{2}{*}{0.4169} & \multirow{2}{*}{0.4276} & \multirow{2}{*}{0.3784} & \multirow{2}{*}{0.4448} \\
     & & & & \\
     \hline
     \multirow{2}{*}{\shortstack{Max. Abs\\Curvature ($m^{-1})$}} & \multirow{2}{*}{0.1433} & \multirow{2}{*}{0.1478} & \multirow{2}{*}{0.1565} & \multirow{2}{*}{0.1426} \\
     & & & & \\
     \hline
     \multirow{2}{*}{\shortstack{Comp. Time\\ ($ms$)}} & \multirow{2}{*}{37.12} & \multirow{2}{*}{73.42} & \multirow{2}{*}{91.35} & \multirow{2}{*}{81.82} \\ %22+15, 53+20, 79+12, 61+20
     & & & & \\
     \Xhline{2\arrayrulewidth}
\end{tabular}
%\end{center}
%\begin{tablenotes}\footnotesize
%\item [] means.
%\end{tablenotes}
\end{threeparttable}
\end{adjustbox}
\end{table}

We applied the proposed algorithm on a test vehicle, as shown in Fig. \ref{fig:6.test_vehicle}. It was a Hyundai Solati model with several sensors and an onboard computer (AMD Ryzen 7 series clocked at 2.2 GHz). We used the \acrshort{mpc} controller from the simulation test and set the number of discs $\textsc{N}_\textsc{DISC}$ to $3$ to increase the safety buffer. The test site, as shown in Fig. \ref{fig:vehicle_test_rviz}, was an open outdoor space with static obstacles (virtual and vehicle), which were all detected before trajectory optimization. The virtual obstacles included predefined obstacles and nonmoving obstacles, which were detected by sensors. The vehicle obstacles, also detected by the sensors, were surrounded by multiple static obstacles. The objective of this experiment was to evaluate the tracking performance of the proposed method using a real vehicle. The vehicle could reach a given goal without collision.

Fig. \ref{fig:vehicle_test_rviz} shows that the vehicle followed the optimized trajectory in four different scenarios: two s-curves, a narrow corridor, and a sharp turn. The lateral and heading errors in the four scenarios are plotted in Fig. \ref{fig:tracking_error_graph_real_world}. The errors in the last $2m$ were excluded from the plots because the controller had limitations in tracking short paths during the stopping phase, which is beyond the scope of this paper. These plots show that the lateral errors were between $-0.24m$ and $+0.24m$ and the heading errors were between $-5.0^\circ$ and $4.5^\circ$, which were lower than the simulation results of the baselines in Fig. \ref{fig:tracking_error_graph_comparision}. Thus, the tracking performance of the proposed method was valid in the real world despite some errors caused by perception and localization in this setting. Furthermore, Table \ref{table:vehicle_test_result} shows the actual vehicle status by following the optimized trajectory. The vehicle sped up to $2.56m/s$, and the velocity, acceleration, and curvature were bounded within the kinodynamic feasibility constraints, namely, longitudinal velocity ($m/s$): $[0, 2.78]$; longitudinal acceleration ($m/s^{2}$): $[-2.0, 2.0]$; lateral acceleration ($m/s^{2}$): $[-1.0, 1.0]$; curvature constraint ($m^{-1}$): $[-1.67, 1.67]$.

%%%%%%%%%%%%%%%%%%%%%%%%%%%%%%%%%%%%%%%%%%%%%%%%%%%%%%%%%%%%%%%%%%%%%%%%%%%%%%%%
\section{CONCLUSIONS}
\label{section:conclusion}
In this paper, we presented novel methodologies, namely disc-type \acrshort{sv}, \acrshort{ipf}, and kinodynamic feasibility penalty, to enable path-velocity coupled trajectory planning for \acrshort{av}s using B-spline curves. Unlike traditional approaches that either suffer from over-approximation issues or use decoupled methods, our proposed approach effectively reduces the over-approximation of the vehicle shape and integrates path and velocity planning to satisfy all kinodynamic constraints of \acrshort{avs}.

The disc-type \acrshort{sv} estimation method reduces over-approximation, thus improving trajectory feasibility and safety. \acrshort{ipf} is a novel method that can identify a collision-free path for \acrshort{avs} by using the collision points obtained from the disc-type \acrshort{sv} to push the path away from obstacles with close obstacle collision penalties and gradually increasing the curvature to flatten the path. Furthermore, we applied a clamped B-spline curvature penalty along with longitudinal and lateral velocity and acceleration penalties to generate kinodynamically feasible trajectories suitable for \acrshort{avs}.

Our extensive experimental results demonstrate that the proposed methods outperform \acrlong{sota} baselines in various simulated environments. Furthermore, real-world experiments validate the simulated tracking performance of our approach, proving its practical applicability. In future work, we will use the proposed method with dynamic obstacles. As this approach is more time efficient than traditional trajectory optimization algorithms, trajectory planning algorithms can be further improved using this method.

\bibliography{references}
\bibliographystyle{IEEEtran}

\printglossary

%\newpage

\begin{IEEEbiography}[{\includegraphics[width=1in,height=1.25in,clip,keepaspectratio]{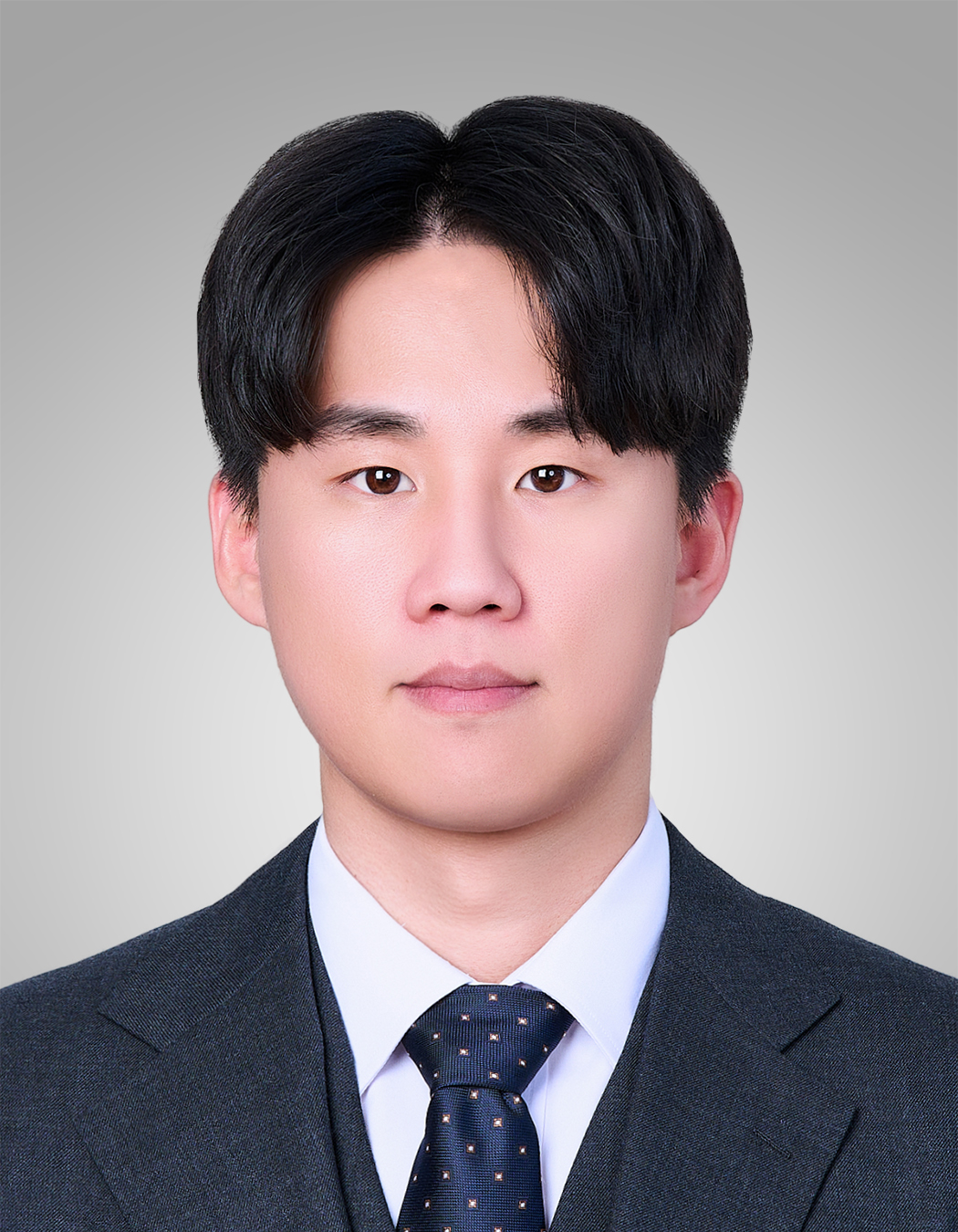}}]{Jongseo Choi} %was born in Gunpo, South Korea, in 1989.
received the B.S. degree in Electronic Engineering from Chungbuk National University, South Korea, in 2015. He received the M.S. degree in Automotive Software Engineering from Chemnitz University of Technology, Germany, in 2020. He was a quality engineer in the mechatronics quality department of Hyundai Mobis Co., Ltd., South Korea, from 2015 to 2018. He worked with IAV Automotive Engineering, Inc., Germany, as an intern researcher working on motion prediction for autonomous vehicles, from 2019 to 2020. Since 2020, he has been with ThorDrive Co., Ltd, Seoul, South Korea, as a senior researcher working on motion planning and multi-agent planning for autonomous driving, which are also his research interests.
\end{IEEEbiography}

\begin{IEEEbiography}[{\includegraphics[width=1in,height=1.25in,clip,keepaspectratio]{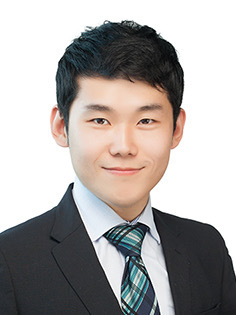}}]{Hyuntai Chin} %was born in Suwon, South Korea, in 1991.
received the B.E. and M.E. degrees in Mechanical Engineering from Nagoya University, Nagoya, Japan, in 2015 and 2017, respectively. In 2018, he was a visiting student researcher of Mechanical Engineering Department of U.C.Berkeley, United States working on cooperative adaptive cruise control.
Since 2020, he has been with ThorDrive Co., Ltd, Seoul, South Korea, working on vehicle control for autonomous driving. His research interests include robust path and speed tracking, vehicle platooning, and optimal control.
\end{IEEEbiography}

\begin{IEEEbiography}[{\includegraphics[width=1in,height=1.25in,clip,keepaspectratio]{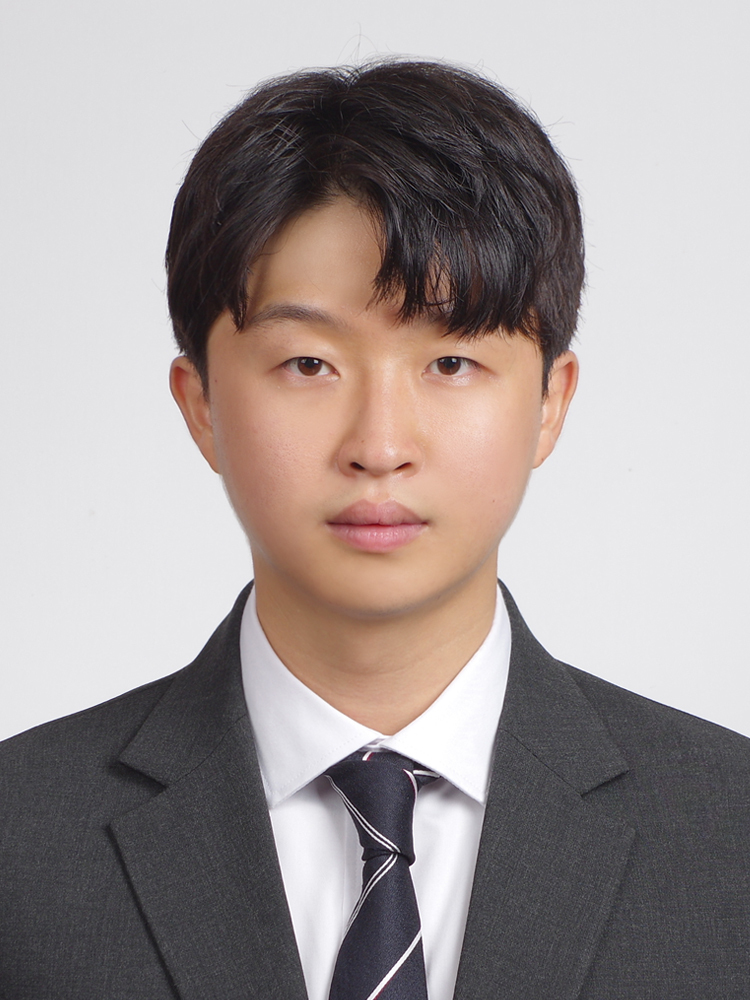}}]{Hyunwoo Park}
received the B.S. degree in mechanical engineering from Yonsei University, Seoul, South Korea, in 2022. He is currently working as a researcher in the Planning team at ThorDrive Co., Ltd, Seoul, South Korea. His research interests include motion planning for autonomous vehicles and reinforcement learning for robots.
\end{IEEEbiography}

\begin{IEEEbiography}[{\includegraphics[width=1in,height=1.25in,clip,keepaspectratio]{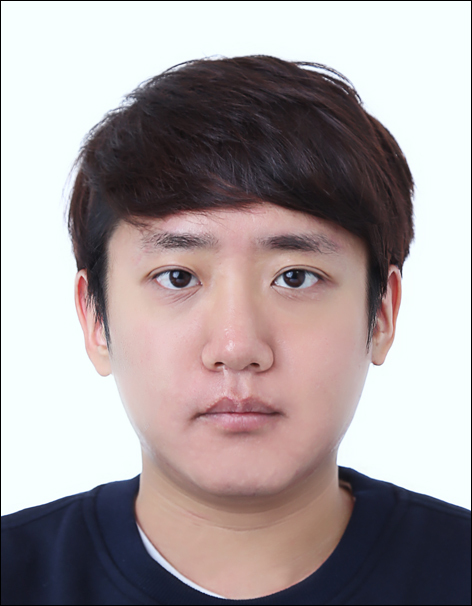}}]{Daehyeok Kwon}
received the B.S. degree in Electronic Engineering from Hanyang University, Seoul, South Korea, in 2018. He is currently pursuing the Ph.D. degree with the Department of Electrical Engineering and Computer Science, Seoul National University, Seoul, South Korea. He is also a Researcher with ThorDrive Co., Ltd, Seoul, South Korea. His current research areas include machine learning and autonomous driving.
\end{IEEEbiography}

\begin{IEEEbiography}[{\includegraphics[width=1in,height=1.25in,clip,keepaspectratio]{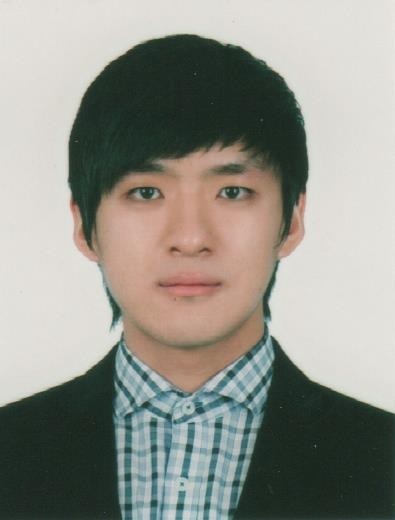}}]{Doosan Baek}
received the B.S. degree in the Department of Electrical Engineering from Seoul National University, Seoul, South Korea, in 2012. He is currently pursuing the Ph.D. degree in the Department of Electrical Engineering and Computer Science from Seoul National University, South Korea. He was a CTO (Chief Technical Officer) at ThorDrive Co., Ltd, South Korea. He is also a CEO at whereable.ai Inc., South Korea. His current research areas include the system architecture of autonomous driving, sensor fusion-based perception, localization, and calibration for autonomous vehicles.
\end{IEEEbiography}

\begin{IEEEbiography}[{\includegraphics[width=1in,height=1.25in,clip,keepaspectratio]{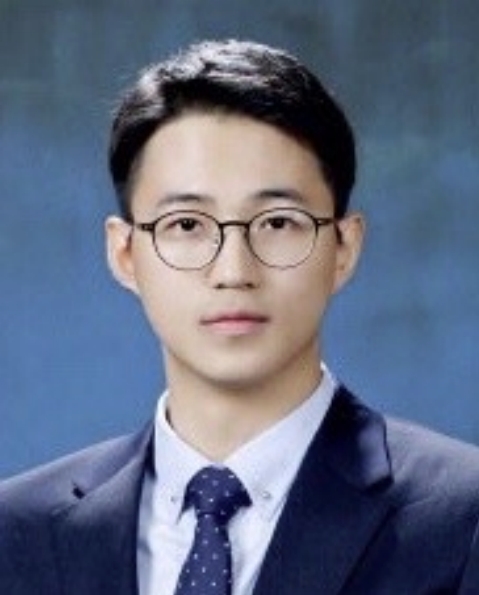}}]{Sang-Hyun Lee}
received the B.S. degree in the Department of Automotive Engineering from Hanyang University, Seoul, South Korea, in 2015, and the Ph.D. degree in the Department of Electrical and Computer Engineering from Seoul National University, Seoul, South Korea, in 2024. He previously worked as a Senior Researcher at ThorDrive, Seoul, South Korea. He is currently an Assistant Professor in the Department of Mobility Engineering at Ajou University, Suwon, South Korea. His research interests include reinforcement learning, imitation learning, and autonomous driving.
\end{IEEEbiography}

\vspace{11pt}


\vfill

\end{document}